%% file: arXiv_version_230608.tex
\documentclass[mnsc,nonblindrev]{informs3} 

\OneAndAHalfSpacedXI 
\usepackage{hyperref}
\usepackage{bbm}
\usepackage{xfakebold}

\usepackage{endnotes}
\let\footnote=\endnote

 \newcommand{\E}{\mathbb E}
\usepackage{natbib}
 \bibpunct[, ]{(}{)}{,}{a}{}{,}%
 \def\bibfont{\small}%

\usepackage{algorithm}
\usepackage{algorithmic}
\usepackage{subfigure}
\usepackage{graphicx}
\usepackage{makecell}

\usepackage{ulem}
\usepackage{booktabs}
\usepackage{multirow}
\usepackage{comment}
\usepackage{tikz}
\usepackage{amsmath}
\newcommand\numberthis{\addtocounter{equation}{1}\tag{\theequation}}

\usepackage{hyperref}

\usepackage{xcolor}
\hypersetup{
    colorlinks,
    linkcolor={red!50!black},
    citecolor={blue!50!black},
    urlcolor={blue!80!black}
}
\TheoremsNumberedThrough     
\ECRepeatTheorems

\EquationsNumberedThrough    

\newcommand{\fbseries}{\unskip\setBold\aftergroup\unsetBold\aftergroup\ignorespaces}
\usepackage{amsmath}
\usepackage{amsfonts}
\usepackage{amssymb}
\usepackage{dsfont}

\begin{document}




\TITLE{A Doubly Stochastic Simulator with Applications in Arrivals Modeling and Simulation}




\ARTICLEAUTHORS{%
\AUTHOR{Yufeng Zheng}
\AFF{Microsoft Research Asia }
\AUTHOR{Zeyu Zheng\footnote{Correspondence to: zyzheng@berkeley.edu. Authors are alphabetically ordered. }}
\AFF{Department of Industrial Engineering and Operations Research, University of California}
\AUTHOR{Tingyu Zhu}
\AFF{ Department of Industrial Engineering and Operations Research, University of California}
 }

\ABSTRACT{%
We propose a framework that integrates classical Monte Carlo simulators and Wasserstein generative adversarial networks to model, estimate, and simulate a broad class of arrival processes with general  non-stationary and multi-dimensional random arrival rates.  Classical Monte Carlo simulators have advantages at capturing the interpretable ``physics" of a stochastic object, whereas neural-network-based simulators have advantages at capturing less-interpretable complicated dependence within a high-dimensional distribution. We propose a doubly stochastic simulator that integrates a stochastic generative neural network and a classical Monte Carlo Poisson simulator, to utilize both advantages. Such integration brings challenges to both theoretical reliability and computational tractability for the estimation of the simulator given real data, where the estimation is done through minimizing the Wasserstein distance between the distribution of the simulation output and the distribution of real data. Regarding theoretical properties, we prove consistency and convergence rate for the estimated simulator under a non-parametric smoothness assumption. Regarding computational efficiency and tractability for the estimation procedure, we address a challenge in gradient evaluation that arise from the discontinuity in the Monte Carlo Poisson simulator. Numerical experiments with synthetic and real data sets are implemented to illustrate the performance of the proposed framework. 
}%

%

\maketitle

%

\input{section/full}

\bibliographystyle{informs2014}

\bibliography{arXiv_version_230608}

\end{document}

%% file: section/full.tex
\section{Introduction}

For several areas in the domain of operations research and management science, such as service systems, logistics and supply chain systems, and financial systems, the randomness of arrivals is one primary source of uncertainty. The arrivals can be customer flows, demand orders, traffic, among others. To put the discussion in context, in this paper, we consider {demand arrivals} to a system that have cycles of operations. A cycle can be a day or a week. Appropriately modeling and statistically characterizing the arrival processes is critical for policy evaluation and performance evaluation in the related systems. Further, efficiently simulating new arrival data that contains certain statistical features is often needed for what-if simulation analysis and decision-making. 
After a statistical model is characterized, efficiently simulating new arrival data that contains certain statistical features is often needed for what-if simulation analysis and decision-making. However, the non-stationarities such as time-of-day effects and complicated correlation structures presented in many arrival data give rise to challenges in statistically reliable modeling and estimation.  To address these needs and challenges, this work provides new and effective approaches for the modeling, estimation, and simulation of arrival processes that have potentially complicated non-stationary and stochastic structures. 

A natural and powerful model for such arrival processes is a non-stationary \textit{Doubly Stochastic Poisson Process} (DSPP), also known as a Cox Process (\cite{cox1954superposition}). It is a Poisson process with a random and time-varying rate (or, intensity) process.
The Poisson structure is a natural assumption for arrival processes in a large number of applications, in particular when the arrivals come from people making their own decisions on whether and when to join the system. 
The Poisson superposition theorem is a mathematical support for the Poisson structure assumption. More theory and empirical justifications can be found in \cite{whitt2002stochastic}, \cite{brown2005statistical}, \cite{koole2013call}, \cite{kim2014choosing}, \cite{ibrahim2016modeling}, among others. In addition to the Poisson structure, empirical evidence suggests that the arrival rate process is not only time-varying but follows a stochastic process by itself. Standard Non-Homogeneous Poisson process (NHPP), on the other hand, does not have  randomness in the arrival rate process. To this regard, empirical evidence has shown that NHPP is not a sufficient model in many cases as it fails to capture the complicated variance and correlation behaviors that arise in real arrival data. \cite{avramidis2004modeling}, \cite{brown2005statistical},  \cite{steckley2005performance}, \cite{steckley2009forecast}, \cite{channouf2012normal}, \cite{ibrahim2016modeling} and \cite{oreshkin2016rate} provide theory and empirical support for modeling the arrival rate with an underlying stochastic process.

In particular, in the seminal work by \cite{oreshkin2016rate}, the authors demonstrate strong empirical evidence for the DSPP model and propose a DSPP model that uses Gamma-compounded Gaussian copula to model the random arrival rates. The model estimation in \cite{oreshkin2016rate} is done through likelihood-based method. One major difficulty in estimation for any DSPP model is that the random arrival rate process is not directly observed, and can only be inferred from the observed arrivals. Moreover, the unobserved random arrival rate process itself may have complicated stochastic structure. This can sometimes pose serious challenges in not only model mis-specification but also computational intractability for model estimation. Consider that a 24-hour day is partitioned into disjoint time intervals at any user-specified desired size. Take half-hour time intervals for example. Under the DSPP model, the arrival count in each half-hour time interval is assumed to have a doubly stochastic Poisson distribution with a random rate. These random rates from all half-hour intervals in a day form a 48-dimensional random vector, whose joint distribution can be complicated to model and estimate.  

To adequately model and efficiently estimate the aforementioned complicated structure in arrival rates, we employ a generative neural network that takes a vector of uniform random variables as input, and maps it to a random vector. With different parameter and structure settings of the generative neural network, the simulated random vector of arrival rates follows different joint distributions. After a random vector of rates is simulated from the stochastic generative neural network, they are sent as the input to standard Monte Carlo Poisson simulators to generate arrival counts. The stochastic generative neural network and the Monte Carlo Poisson simulator  together form the \textit{doubly stochastic simulator} as the focus of this work. To estimate this simulator from data, we fit the parameters of the generative neural networks with the objective of matching the generated distribution of arrival counts with the empirical distribution constructed from data. Overall, we integrate the doubly stochastic simulator and the associated the estimation procedure through Wasserstein distance and adversarial training. We refer to the simulator and estimation procedure as the \textit{Doubly Stochastic Wasserstein Generative Adversarial Networks} (DS-WGAN) framework for the convenience of description. 



 The advantages of the DS-WGAN framework are three-fold. First, the use of generative neural networks in modeling the joint distribution of unobserved arrival rates assume no parametric or specific distribution family, which provides adequate modeling capacity and reduces the risk of mis-specification. Second, the simulator exploits the known doubly stochastic Poisson structure in arrival processes, and assigns the generative neural network to only focus on modeling the unobserved random arrival rate process. This is in contrast to the pure generative neural networks used in the literature where the networks are assigned to learn everything, which requires excessive representative data. This advantage may be particularly
attractive in any operations research applications for which massive datasets are
unavailable. Third, exploiting the known data structure enables what-if simulation
analysis (\cite{nelson2016some}), which can support performance evaluation and decision making for systems that do
not yet exist. Consider a what-if planning decision in the service system for the future three months. What is an appropriate workers scheduling design if the expected number of daily user arrivals in the future three months is 23\% larger, compared to that in the historical data? To answer such what-if questions, there is a need to simulate new arrival processes that have 23\% of increase in the average daily demand but preserve the time-of-day pattern and correlation structures within a day. The DS-WGAN framework is a natural fit for this task, which cannot be handled by the plain generative neural networks framework.  


The challenges are mainly from two aspects, the statistical aspect and the computational aspect. The statistical aspect considers what types of doubly stochastic Poisson processes can be consistently learned by DS-WGAN, as well as the rate of convergence. In fact, even for the standard Wasserstein Generative Adversarial Networks (WGAN) framework (\cite{villani2008optimal} and \cite{arjovsky2017wasserstein}) without the doubly stochastic structure, limited theory studies exist on the statistical guarantees. \cite{bai2018approximability} and \cite{chen2020statistical} are the first to provide statistical theory for WGAN,  {but require the target distribution to have bounded support, and the sample data to be independent and identically distributed.} In our work, we exploit the knowledge of doubly stochastic Poisson structure to provide statistical guarantees for DS-WGAN,  {which generates unbounded distributions. We also allow weak correlation to exist among the sample data.} The computational aspect meets a challenge in the evaluation of stochastic gradient. Nominally, the model estimation (training) of neural-network assisted framework is carried out through stochastic gradient descent algorithms. Gradients of the discriminator output with respect to the generator network parameters are computed via backpropagation. However, as  backpropagation goes through the Poisson counts simulator, the sample-path gradient vanishes almost everywhere, because the simulator output is discrete. To circumvent this challenge, we construct an approximation that provides a non-vanishing and meaningful gradient, and can be conveniently implemented in the backpropagation routine. Numerical experiments suggest that this approximation gives a stable and successful model estimation (training) of DS-WGAN.


The use of neural networks among other machine learning tools to model arrival processes is receiving rising interests.  \cite{du2016recurrent} and \cite{mei2017neural} are two of the pioneering works in connecting Hawkes processes with Recurrent Neural Networks (RNN). They model and learn the conditional intensity function and use it to generate the next inter-arrival time. \cite{wang2020estimating} consider a DSPP model in which the intensity process is driven by a general Stochastic Differential Equation (SDE). They discuss the use of neural networks in modeling the unknown drift and diffusion functions of SDE, as well as the associated model training using variational inference. \cite{thiongane2022learning} use neural networks to estimate conditional waiting time distributions that aims to provide distributional wait forecasts to customers. 
\cite{herbert2020nim} use neural networks to model and simulate general probability distributions, including that of inter-arrival times for a non-homogeneous Poisson process. Our work differentiates from theirs by focusing on the doubly stochastic Poisson processes with general arbitrary random intensity processes. Theory-wise, to the best of our knowledge, we are the first to provide statistical guarantees for the estimation procedure of neural-network assisted doubly stochastic simulators. From an operational point of view, instead of modeling and simulation of the next inter-arrival time (microscopic structure), our work focuses on learning the dependence structures at a longer time scale (e.g., half-hour counts) that is relevant to service operations (mesoscopic structure). Our numerical experiments in Section \ref{sec:exper_dspp_cir} suggest that learning this mesoscopic structure can be critical and sufficient to support service operations performance evaluation. The idea of studying mesoscopic structure and running relevant run-through-queue experiments was first introduced by \cite{glynn2014}. 


\section{Model Setup}
\label{sec:model_setup}

Suppose that $n$ independent days of arrival data are observed. Let $N_i$ denote the arrival process on the $i$-th day, for $i=1,2,\ldots,n$. We assume that the $N_i$'s are $n$ replications of a doubly stochastic Poisson process (DSPP). A DSPP is a point process $N=(N(t):t\in[0,T])$ with a random intensity function $\lambda=(\lambda(t):t\in[0,T])$. For a point process, with any given $t$, the random variable $N(t)$ denotes the number of arrivals in a day over the time interval $[0,t]$. The DSPP point process is \textit{simple} in the sense that $N$ increases exactly by one at each arrival epoch and no batch arrivals are possible. (The feature of batch arrivals can be separately modeled and then compounded with an underlying DSPP, but is not the focus of this work.) The random intensity function $\lambda=(\lambda(t):t\in[0,T])$ is a random object defined on a functional space on $[0,T]$ with proper topology. For example, the function space can be $C[0,T]$ endowed with the Skorokhod $J_1$ topology, which includes all the continuous functions on $[0,T]$. Conditional on a realization of the random intensity function $\lambda=(\lambda(t):t\in[0,T])$, $N=(N(t):t\in[0,T])$ is a nonhomogeneous Poisson process. That is, conditional on $\lambda$, for any set of disjoint time intervals $(t_1,t_1']$, $(t_2,t_2']$, ... , $(t_m,t_m']$ on $[0,T]$, the arrival counts on each time interval, $N(t_1') - N(t_1)$, $N(t_2') - N(t_2)$, ... , $N(t_m') - N(t_m)$ are independent Poisson random variables. Conditional on $\lambda$, the expectation of $N(t_k') - N(t_k)$ is given by 
\[
\int_{t_k}^{t_k'}\lambda(s)\,ds
\]
for $k=1,2,\ldots,m$. Unconditionally, the marginal distributions of $N(t_1') - N(t_1)$, $N(t_2') - N(t_2)$, ... , $N(t_m') - N(t_m)$ are not necessarily Poisson nor independent, because of the stochastic structure of the underlying random intensity process. 

Because of the continuous-time nature, the random intensity function is an infinite-dimensional random object. Various statistical features such as variance and correlation can arise with different magnitude at different time scales, such as $1$-minute, $5$-minutes, half-hour, etc. It is therefore impossible for a fixed model to capture the statistical features at every time scale for a general DSPP, using only a finite size of data. {Therefore, our model aims to learn not the continuous-time random process, but the integration of it over discretized time intervals.}  In our modeling framework, the simulated arrival processes are targeted to \textbf{match the statistical features of the given arrival data at all the time scales that are equal to or larger than a given resolution}. We next elaborate on this goal. Consider a resolution $\Delta>0$. Set $p = T/\Delta$ and we presume that $p$ is an integer. The time horizon $[0,T]$ is partitioned into $p$ time intervals of equal length. The equal length notion is for presentation simplicity and not necessarily required. 
Let $\mathbf{X} = (X_{1},X_{2},\ldots,X_{p})$ be a random vector that records the interval counts in each time interval. Specifically, for $i=1,2,\ldots,p$, 
\[
X_i = N\left(i\Delta\right)-N\left((i-1)\Delta\right).
\]
Because $N = (N(t): t\in [0,T])$ is a doubly stochastic Poisson process, conditional on the random intensity function $\lambda = (\lambda(t): 0\le t\le T)$, the marginal distribution of $X_i$ is Poisson with mean 
\[
\Lambda_i \triangleq \int_{(i-1)\Delta}^{i\Delta} \lambda(s) \, ds.
\]
The unconditional joint distribution of $\mathbf{X}$ is therefore determined by the joint distribution of $\mathbf{\Lambda} = (\Lambda_1,\Lambda_2,\ldots,\Lambda_p)$. Because the underlying random intensity function $\lambda$ can have arbitrary stochastic structure, we do not impose any distributional assumption on $\mathbf{\Lambda}$ when building our model and simulator. 

Denote the $\mathbf{X}_1,\mathbf{X}_2,\ldots,\mathbf{X}_n$ as $n$ independent and identically distributed copies (iid) of $\mathbf{X}$ observed in the $n$ days of arrival data. We call the data set $\mathbf{X}_1,\mathbf{X}_2,\ldots,\mathbf{X}_n$ as \textit{interval count data}. Based on such data, we build a doubly stochastic simulator that, once trained using the data, can generate independent copies of arrival counts vector that is targeted to have the same joint distribution as $\mathbf{X}$. The specific structures of the doubly stochastic simulator is discussed in section  \ref{sec:ds_wgan_simulator}, and in later sections \ref{sec:ds_wgan_discriminator}, \ref{sec:ds_wgan_statistical} and \ref{sec:model_estimation_ds_wgan} we discuss how the doubly stochastic simulator is estimated from observed interval count data, as well as the statistical properties. Additionally, one can choose to add an \textit{arrival epochs simulator}, which takes as inputs a vector of arrival counts $\mathbf{X}=(x_1,x_2,\ldots,x_p)$ simulated by the doubly stochastic simulator, and simulate arrival epochs in continuous time. {Within each of the $p$ intervals, the simulated arrival process is assumed to have piecewise-linear arrival rate $\lambda = (\lambda(t): 0\le t\le T)$.} We give the details of the arrival epochs simulator in Section \ref{ec:arrival_epochs_simulator}.


\section{Doubly Stochastic Simulator and Wasserstein Training}
\label{sec:ds_neural_network}
In this section, we describe a framework called Doubly Stochastic Wasserstein Generative Adversarial Networks (DS-WGAN). In Section \ref{sec:ds_wgan_simulator}, we discuss the doubly stochastic simulator in the DS-WGAN framework. In Section \ref{sec:ds_wgan_discriminator}, we discuss the model estimation and training for the proposed doubly stochastic simulator of the DS-WGAN framework. 



\subsection{Modeling Framework of DS-WGAN: A Doubly Stochastic Simulator}
\label{sec:ds_wgan_simulator}
Note that $\mathbf{X} = (X_{1},X_{2},\ldots,X_{p})$ represents the random vector in $\mathbb{Z}^p$ that records the arrival counts in each consecutive time interval within a day. Let $\mu$ denote the true joint probability distribution of $\mathbf{X}$. The training data is $n$ iid copies of $\mathbf{X}$, given by $\mathbf{X}_1,\mathbf{X}_2,\ldots,\mathbf{X}_n$. Let $\hat{\mu}_n$ denote the empirical distribution of the data. Different from a classical routine of first estimating (parameters of) a probability distribution $\mu$ from data and then simulating new samples according to the estimated distribution, DS-WGAN contains a routine that implicitly learns  the distribution from data and obtains a simulator that can generate new samples from the desired distribution $\mu$. The simulator aims at taking a vector of iid simple random variables (such as Uniform(0,1)) as inputs and mapping this vector into one that has the same distribution as $\mu$. In the framework of DS-WGAN, the simulator is composed of two parts - a neural-network based generator and a Poisson counts simulator. Figure \ref{fig:fram_ds_wgan1} illustrates the doubly stochastic simulator, with the details and notation to be specified right afterwards. 

\begin{figure}[ht!]
\centering

\tikzset{every picture/.style={line width=0.75pt}} 

\tikzset{every picture/.style={line width=0.75pt}} 

\tikzset{every picture/.style={line width=0.75pt}} 

\begin{tikzpicture}[x=0.75pt,y=0.75pt,yscale=-1,xscale=1]

\draw  [color={rgb, 255:red, 153; green, 0; blue, 1 }  ,draw opacity=1 ][fill={rgb, 255:red, 244; green, 204; blue, 205 }  ,fill opacity=1 ][line width=0.75]  (35.07,30.12) .. controls (35.07,30.12) and (35.07,30.12) .. (35.07,30.12) -- (45.25,30.12) .. controls (45.25,30.12) and (45.25,30.12) .. (45.25,30.12) -- (45.25,130.62) .. controls (45.25,130.62) and (45.25,130.62) .. (45.25,130.62) -- (35.07,130.62) .. controls (35.07,130.62) and (35.07,130.62) .. (35.07,130.62) -- cycle ;
\draw  [color={rgb, 255:red, 56; green, 118; blue, 30 }  ,draw opacity=1 ][fill={rgb, 255:red, 217; green, 234; blue, 211 }  ,fill opacity=1 ][line width=0.75]  (206.9,30.12) .. controls (206.9,30.12) and (206.9,30.12) .. (206.9,30.12) -- (217.08,30.12) .. controls (217.08,30.12) and (217.08,30.12) .. (217.08,30.12) -- (217.08,130.62) .. controls (217.08,130.62) and (217.08,130.62) .. (217.08,130.62) -- (206.9,130.62) .. controls (206.9,130.62) and (206.9,130.62) .. (206.9,130.62) -- cycle ;
\draw [color={rgb, 255:red, 118; green, 118; blue, 118 }  ,draw opacity=1 ][line width=0.75]    (166.99,80.37) -- (204.2,80.37) ;
\draw [shift={(206.2,80.37)}, rotate = 180] [color={rgb, 255:red, 118; green, 118; blue, 118 }  ,draw opacity=1 ][line width=0.75]    (10.93,-3.29) .. controls (6.95,-1.4) and (3.31,-0.3) .. (0,0) .. controls (3.31,0.3) and (6.95,1.4) .. (10.93,3.29)   ;
\draw  [color={rgb, 255:red, 118; green, 118; blue, 118 }  ,draw opacity=1 ][line width=0.75]  (85.46,38.67) .. controls (85.46,33.95) and (89.29,30.12) .. (94.01,30.12) -- (158.36,30.12) .. controls (163.08,30.12) and (166.9,33.95) .. (166.9,38.67) -- (166.9,122.08) .. controls (166.9,126.8) and (163.08,130.62) .. (158.36,130.62) -- (94.01,130.62) .. controls (89.29,130.62) and (85.46,126.8) .. (85.46,122.08) -- cycle ;
\draw  [color={rgb, 255:red, 118; green, 118; blue, 118 }  ,draw opacity=1 ][line width=0.75]  (257.27,38.67) .. controls (257.27,33.95) and (261.1,30.12) .. (265.82,30.12) -- (330.17,30.12) .. controls (334.89,30.12) and (338.72,33.95) .. (338.72,38.67) -- (338.72,122.08) .. controls (338.72,126.8) and (334.89,130.62) .. (330.17,130.62) -- (265.82,130.62) .. controls (261.1,130.62) and (257.27,126.8) .. (257.27,122.08) -- cycle ;
\draw [color={rgb, 255:red, 118; green, 118; blue, 118 }  ,draw opacity=1 ][line width=0.75]    (217.82,80.37) -- (255.03,80.37) ;
\draw [shift={(257.03,80.37)}, rotate = 180] [color={rgb, 255:red, 118; green, 118; blue, 118 }  ,draw opacity=1 ][line width=0.75]    (10.93,-3.29) .. controls (6.95,-1.4) and (3.31,-0.3) .. (0,0) .. controls (3.31,0.3) and (6.95,1.4) .. (10.93,3.29)   ;
\draw [color={rgb, 255:red, 118; green, 118; blue, 118 }  ,draw opacity=1 ][line width=0.75]    (338.55,80.37) -- (375.76,80.37) ;
\draw [shift={(377.76,80.37)}, rotate = 180] [color={rgb, 255:red, 118; green, 118; blue, 118 }  ,draw opacity=1 ][line width=0.75]    (10.93,-3.29) .. controls (6.95,-1.4) and (3.31,-0.3) .. (0,0) .. controls (3.31,0.3) and (6.95,1.4) .. (10.93,3.29)   ;
\draw  [color={rgb, 255:red, 10; green, 83; blue, 148 }  ,draw opacity=1 ][fill={rgb, 255:red, 201; green, 218; blue, 248 }  ,fill opacity=1 ][line width=0.75]  (378.51,30.12) .. controls (378.51,30.12) and (378.51,30.12) .. (378.51,30.12) -- (388.69,30.12) .. controls (388.69,30.12) and (388.69,30.12) .. (388.69,30.12) -- (388.69,130.62) .. controls (388.69,130.62) and (388.69,130.62) .. (388.69,130.62) -- (378.51,130.62) .. controls (378.51,130.62) and (378.51,130.62) .. (378.51,130.62) -- cycle ;
\draw [color={rgb, 255:red, 118; green, 118; blue, 118 }  ,draw opacity=1 ][line width=0.75]    (46,80.37) -- (83.21,80.37) ;
\draw [shift={(85.21,80.37)}, rotate = 180] [color={rgb, 255:red, 118; green, 118; blue, 118 }  ,draw opacity=1 ][line width=0.75]    (10.93,-3.29) .. controls (6.95,-1.4) and (3.31,-0.3) .. (0,0) .. controls (3.31,0.3) and (6.95,1.4) .. (10.93,3.29)   ;
\draw  [draw opacity=0] (52.51,145) .. controls (48.72,144.92) and (45.66,141.82) .. (45.66,138) .. controls (45.66,137.97) and (45.66,137.94) .. (45.66,137.92) -- (52.66,138) -- cycle ; \draw  [color={rgb, 255:red, 118; green, 118; blue, 118 }  ,draw opacity=1 ] (52.51,145) .. controls (48.72,144.92) and (45.66,141.82) .. (45.66,138) .. controls (45.66,137.97) and (45.66,137.94) .. (45.66,137.92) ;
\draw [color={rgb, 255:red, 118; green, 118; blue, 118 }  ,draw opacity=1 ]   (52.51,145) -- (205.33,145) ;
\draw  [draw opacity=0] (205.33,145) .. controls (209.13,145.08) and (212.19,148.18) .. (212.19,152) .. controls (212.19,152.03) and (212.19,152.06) .. (212.18,152.09) -- (205.19,152) -- cycle ; \draw  [color={rgb, 255:red, 118; green, 118; blue, 118 }  ,draw opacity=1 ] (205.33,145) .. controls (209.13,145.08) and (212.19,148.18) .. (212.19,152) .. controls (212.19,152.03) and (212.19,152.06) .. (212.18,152.09) ;
\draw  [draw opacity=0] (371.86,145) .. controls (375.65,144.92) and (378.71,141.82) .. (378.71,138) .. controls (378.71,137.97) and (378.71,137.94) .. (378.71,137.92) -- (371.71,138) -- cycle ; \draw  [color={rgb, 255:red, 118; green, 118; blue, 118 }  ,draw opacity=1 ] (371.86,145) .. controls (375.65,144.92) and (378.71,141.82) .. (378.71,138) .. controls (378.71,137.97) and (378.71,137.94) .. (378.71,137.92) ;
\draw [color={rgb, 255:red, 118; green, 118; blue, 118 }  ,draw opacity=1 ]   (371.86,145) -- (219.04,145) ;
\draw  [draw opacity=0] (219.04,145) .. controls (215.24,145.08) and (212.18,148.18) .. (212.18,152) .. controls (212.18,152.03) and (212.18,152.06) .. (212.18,152.09) -- (219.18,152) -- cycle ; \draw  [color={rgb, 255:red, 118; green, 118; blue, 118 }  ,draw opacity=1 ] (219.04,145) .. controls (215.24,145.08) and (212.18,148.18) .. (212.18,152) .. controls (212.18,152.03) and (212.18,152.06) .. (212.18,152.09) ;

\draw (88.18,69.87) node [anchor=north west][inner sep=0.75pt]  [font=\fontsize{0.6em}{0.72em}\selectfont] [align=left] {\begin{minipage}[lt]{56.229375000000005pt}\setlength\topsep{0pt}
\begin{center}
{\fontfamily{ptm}\selectfont Neural-network}\\{\fontfamily{ptm}\selectfont based generator }$\displaystyle g$
\end{center}

\end{minipage}};
\draw (269.99,70.37) node [anchor=north west][inner sep=0.75pt]  [font=\fontsize{0.6em}{0.72em}\selectfont] [align=left] {\begin{minipage}[lt]{39.280625pt}\setlength\topsep{0pt}
\begin{center}
{\fontfamily{ptm}\selectfont Poisson counts}\\{\fontfamily{ptm}\selectfont simulator $\mathbf{h}$}
\end{center}

\end{minipage}};
\draw (1.66,14.58) node [anchor=north west][inner sep=0.75pt]  [font=\fontsize{0.6em}{0.72em}\selectfont] [align=left] {\begin{minipage}[lt]{53.86000000000001pt}\setlength\topsep{0pt}
\begin{center}
{\fontfamily{ptm}\selectfont Random noise $\mathbf{Y}$}
\end{center}

\end{minipage}};
\draw (173.99,3.58) node [anchor=north west][inner sep=0.75pt]  [font=\fontsize{0.6em}{0.72em}\selectfont] [align=left] {\begin{minipage}[lt]{58.120625000000004pt}\setlength\topsep{0pt}
\begin{center}
{\fontfamily{ptm}\selectfont Simulated Poisson}\\{\fontfamily{ptm}\selectfont rates }$\displaystyle g( \mathbf{Y};\theta )$
\end{center}

\end{minipage}};
\draw (335.1,3.58) node [anchor=north west][inner sep=0.75pt]  [font=\fontsize{0.6em}{0.72em}\selectfont] [align=left] {\begin{minipage}[lt]{72.549375000000005pt}\setlength\topsep{0pt}
\begin{center}
{\fontfamily{ptm}\selectfont Simulated arrival counts}\\{\fontfamily{ptm}\selectfont  vector }$\displaystyle \mathbf{h}( g( \mathbf{Y};\theta ))$
\end{center}

\end{minipage}};
\draw (129.68,153.62) node [anchor=north west][inner sep=0.75pt]  [font=\fontsize{0.6em}{0.72em}\selectfont] [align=left] {\begin{minipage}[lt]{117.089375pt}\setlength\topsep{0pt}
\begin{center}
{\fontfamily{ptm}\selectfont The simulator of DS-WGAN (Section 3.1)}
\end{center}

\end{minipage}};

\end{tikzpicture}

\caption{The simulator of DS-WGAN and the arrival epochs simulator.} 
\label{fig:fram_ds_wgan1}
\end{figure}

The neural-network based generator is a mapping $g: \mathbb{R}^p\rightarrow \mathbb{R}^p$ that adopts a neural network architecture with neural network parameters. Specifically, given the number of layers $L\in \mathbb{Z}^+$ in the neural network and $n_l\in \mathbb{Z}^+$ as the width of the $l$-th layer for $l=1,2,\cdots,L$, for an input variable $\mathbf{y}\in\mathbb{R}^p$, the functional form of the generator is given by   
\begin{equation}
\label{eq:generator_NN_definition}
g(\mathbf{y};\mathbf{W},\mathbf{b})=\mathbf{W}_L \cdot \sigma\left(\mathbf{W}_{L-1} \cdots \sigma\left(\mathbf{W}_{1} \mathbf{y}+\mathbf{b}_{1}\right) \cdots+\mathbf{b}_{L-1}\right)+\mathbf{b}_{L}    
\end{equation}
in which $\mathbf{W} = (\mathbf{W}_1,\mathbf{W}_2,\cdots,\mathbf{W}_L)$ and $\mathbf{b} = (\mathbf{b}_1,\mathbf{b}_2,\cdots,\mathbf{b}_L)$ represent all the parameters in the neural network, with $\mathbf{W}_l\in \mathbb{R}^{n_l\times n_{l-1}}$, $\mathbf{b}_l\in \mathbb{R}^{n_l\times 1}$ for $l=1,2,\ldots,L$, where $n_0 = p$ is set as the dimension of the input variable. The operator $\sigma(\cdot)$, known as the activation function, takes a vector of any dimension as input and is a component-wise operator. As an example, for any $q\in\mathbb{Z}^+$ and any vector $\mathbf{x} = (x_1,x_2,\cdots,x_q)^\top \in \mathbb{R}^{q\times 1}$, we can have
\[
\sigma(\mathbf{x}) \triangleq (\max(x_1,0),\max(x_2,0),\cdots,\max(x_q,0))^\top.
\]

Next we introduce the \textit{Poisson counts simulator} that takes the output from the generator as input, and simulates a vector of Poisson counts. Denote $\mathbf{Y}$ as a random vector in which each dimension is an independent Uniform(0,1) random variable. (Other possible choices of distribution may be exponential or standard normal.) Taking $\mathbf{Y}$ as inputs, the generator $g$ outputs a $p$-dimensional vector $g(\mathbf{Y}): = (g_1(\mathbf{Y}),g_2(\mathbf{Y}),\ldots,g_p(\mathbf{Y}))$ that is intended to be a realization of the vector of random Poisson rates for $\mathbf{X}$. The Poisson counts simulator, denoted as $\mathbf{h}$, is a random mapping from $\mathbb{R}_+^p$ to $\mathbb{Z}^p$. Mathematically, the Poisson counts simulator involves $p$ mutually independent Poisson random variables with $g_i(\mathbf{Y})$, denoted as $M_i(g_i(\mathbf{Y})),i=1,2,\ldots,p$, which are also independent of the randomness in $\mathbf{Y}$. The Poisson counts simulator $\mathbf{h}$ maps any realization of the random intensities $g(\mathbf{Y})$ into
\begin{equation}
    \label{eq:poisson_count_simulator}
    \mathbf{h}(g(\mathbf{Y})) = (M_1(g_1(\mathbf{Y})), M_2(g_2(\mathbf{Y})),\ldots,M_p(g_p(\mathbf{Y})))^\top.
\end{equation}
We also use $h_i(g(\mathbf{Y})),i=1,2,\ldots,p$ to represent the $i$-th dimension element of $ \mathbf{h}(g(\mathbf{Y}))$. We note that conditional on $g(\mathbf{Y})$, the simulation of the Poisson counts is independent of all other sources of randomness and is straightforward to implement. 

\subsection{Modeling Framework of DS-WGAN: Wassertein Distance and Discriminator} \label{sec:ds_wgan_discriminator}
In the previous section, the simulator generates $ \mathbf{h}(g(\mathbf{Y}))$ that is intended to have approximately the same distribution as the true distribution $\mu$. To evaluate how good a simulator is, we use the Wasserstein distance to quantify the distance between the distribution of $\mathbf{h}(g(\mathbf{Y}))$ and the empirical distribution $\hat{\mu}_n$ from data. For any given generator $g$, denote the distribution of $\mathbf{h}(g(\mathbf{Y}))$ as $\mu_{\mathbf{h}(g(\mathbf{Y}))}$. The Wasserstein distance is then given by
\begin{equation}
    \label{eq:w-dist}
    W\left(\hat{\mu}_n, \mu_{\mathbf{h}(g(\mathbf{Y}))}\right) =\inf _{\gamma \in \Pi\left(\hat{\mu}_n, \mu_{\mathbf{h}(g(\mathbf{Y}))}\right)} \mathbb{E}_{(\mathbf{X}_r, \mathbf{X}_g) \sim \gamma}[\Vert\mathbf{X}_r - \mathbf{X}_g\Vert_2],    
\end{equation}
where $\Pi\left(\hat{\mu}_{n}, \mu_{\mathbf{h}(g(\mathbf{Y}))}\right)$ denotes the set of all joint distributions whose marginals are respectively
$\hat{\mu}_{n}$ and $\mu_{\mathbf{h}(g(\mathbf{Y}))}$, and $\Vert\cdot\Vert_2$ denotes the $L_2$ norm. The notation of $(\mathbf{X}_r, \mathbf{X}_g) \sim \gamma$ represents that $(\mathbf{X}_r, \mathbf{X}_g)$ is a random vector that follows the distribution $\gamma$. With the Wasserstein distance in hand, the best generator within a class $\mathcal{G}_{\mathrm{NN}}$, whose structure will be specified in the next section, is given by the following optimization problem
\begin{equation}
    g^*\in \argmin_{g \in \mathcal{G}_{\mathrm{NN}}}W\left(\hat{\mu}_n, \mu_{\mathbf{h}(g(\mathbf{Y}))}\right). \label{eq:WGAN_distance_opt}
\end{equation}
However, the Wasserstein distance in high dimensions does not carry a closed-form computation. Alternatively, using the Kantorovich-Rubinstein duality (\cite{villani2008optimal}), we have 
$$
\begin{aligned}
    W\left(\hat{\mu}_n, \mu_{\mathbf{h}(g(\mathbf{Y}))}\right)=&\sup _{\|f\|_{L} \leq 1} \mathbb{E}_{\mathbf{X}_r \sim \hat{\mu}_n}[f(\mathbf{X})]-\mathbb{E}_{\mathbf{X}_g \sim \mu_{\mathbf{h}(g(\mathbf{Y}))}}[f(\mathbf{X}_g)]
    \\=&\sup _{\|f\|_{L} \leq 1} \mathbb{E}_{\mathbf{X}_r \sim \hat{\mu}_n}[f(\mathbf{X})]-\mathbb{E}_{\mathbf{Y} \sim \nu}[f(\mathbf{h}(g(\mathbf{Y})))],
\end{aligned}
$$
where $\nu$ represents a $p$-dimensional random vector with each element as an independent Uniform(0,1) random variable, the supremum is over all the 1-Lipschitz functions $f$ (i.e., $|f(\mathbf{x}_1) - f(\mathbf{x}_2)|\leq \Vert\mathbf{x}_1 - \mathbf{x}_2\Vert_2$ for any $\mathbf{x}_1$ and $\mathbf{x}_2$ in $\mathbb{R}^d$). The optimization problem in (\ref{eq:WGAN_distance_opt}) is then equivalent to 
\begin{equation}
g^*\in \argmin_{g \in \mathcal{G}_{\mathrm{NN}}}\sup _{\|f\|_{L} \leq 1} \mathbb{E}_{\mathbf{X} \sim \hat{\mu}_n}[f(\mathbf{X})]-\mathbb{E}_{\mathbf{Y} \sim \nu}[f(\mathbf{h}(g(\mathbf{Y})))]. \label{eq:WGAN_dual_opt}
\end{equation}

In the optimization problem (\ref{eq:WGAN_dual_opt}), the computation to find the best $f$ is not tractable. Instead of searching over the space of all $1$-Lipschitz functions, we use a neural network to approximate $f$ and search over all $1$-Lipschitz functions parameterized by the neural network architecture. Computationally, the $1$-Lipschitz constraint can be enforced by weight clipping or gradient penalty (\cite{gulrajani2017improved}). Such a mapping $f$ parameterized by neural network parameters is called \textit{discriminator}. The architecture of a discriminator adopts the same form of (\ref{eq:generator_NN_definition}), with the exception that $n_L = 1$, so that the output of a discriminator is a scalar.

To distinguish the parameters of the generator $g$ and the discriminator $f$, we use $\theta$ and $\omega$ to denote and summarize the parameters ($\mathbf{W}$ and $\mathbf{b}$) of the generator and the discriminator, respectively. In the following, we sometimes use the abbreviated notation $f_\omega$ and $g_\theta$, or $f(\,\cdot\,; \omega)$ and $g(\,\cdot\,; \theta)$. With  $n$ iid copies of data $\mathbf{X}_1,\mathbf{X}_2,\ldots,\mathbf{X}_n$, the optimization problem in (\ref{eq:WGAN_dual_opt}) is approximated by 
\begin{equation}
    \left(g_n^*, f_n^*\right) \in \argmin_{g_\theta \in \mathcal{G}_{\mathrm{NN}}} \max_{f_\omega \in \mathcal{F}_{\mathrm{NN}}} \E_{\mathbf{Y}\sim \nu} \left[f(\mathbf{h}(g(\mathbf{Y};\theta);\omega ))\right] - \frac{1}{n}\sum_{i=1}^n f(\mathbf{X}_i;\omega).
    \label{eq:WGAN_opt}
\end{equation}
where $\mathcal{F}_{\mathrm{NN}}$ denotes a class of discriminators.

{Additionally, we may add a transformation layer between the discriminator and the generated/real data, which constructs a task-specific distribution from the generated/real arrival count distribution. The discriminator, taking the transformed data as input, then focuses on information relevant for the application of interest, instead of capturing the arrival count distribution in general.}

\section{Statistical Theory of DS-WGAN} 

\label{sec:ds_wgan_statistical}
In this section, we prove that the DS-WGAN framework is able to learn distributions of a wide class of doubly stochastic Poisson distributions with random intensities, if the neural network architectures are appropriately chosen.
In the following part, we describe the specific assumptions regarding the joint distribution of the underlying random intensities, and the neural network architectures of the generator and discriminator.

\subsection{Assumptions and Theorem Statement}
Recall that the real distribution $\mu$ denotes the distribution of the vector of interval counts $\mathbf{X} = (X_1,X_2,\ldots,X_p)$, in which each $X_i$ is a doubly stochastic Poisson distributed random variable with a random rate $\Lambda_i$. The dependence structure within  $\mathbf{X}$ is determined by the dependence structure within the unobserved vector of random rates $\mathbf{\Lambda}=(\Lambda_1, \Lambda_2,\ldots,\Lambda_p)$. We do not assume any parametric structure on the joint distribution $\pi_\mathbf{\Lambda}$ of $\mathbf{\Lambda}$, but to support the theory analysis, we need some regularity conditions. We summarize the assumptions on  $\pi_\mathbf{\Lambda}$ as follows.

\begin{assumption}\label{assum:density}
The joint density function $\pi_\mathbf{\Lambda}$ satisfies the following regularity conditions.
\begin{itemize}
    \item[a.)] The support of $\pi_\mathbf{\Lambda}$ is compact, convex. Specifically, there exists a constant $C_\mathbf{\Lambda}>0$ such that the support of $\pi_\mathbf{\Lambda}$ is a subset of $[0,C_\mathbf{\Lambda}]^p$. 
    \item[b.)] The density of $\pi_\mathbf{\Lambda}$ is bounded away from 0 on its support. That is, the infimum of $\pi_\mathbf{\Lambda}$ over its support is positive. 
    \item[c.)] The density of $\pi_\mathbf{\Lambda}$ is differentiable and has uniformly bounded first order partial derivatives across its support. 
\end{itemize}
\end{assumption}
The convexity and compactness assumption of the support facilitates the successful finding of a transformation from the random seed to the desired distribution. The bounded-away-from-zero assumption is standard in the theory literature for Wasserstein distance and optimal transport. \footnotemark[1]\footnotetext[1]{Assumption a.) implies that the density of $\pi_{\mathbf{\Lambda}}$ is discontinuous at the boundary.} The smoothness of the density function across the support facilitates the analysis of the convergence rate for the generator, and has been standard in the non-parametric statistics literature.

Next, we specify the neural network architecture of the discriminator and the generator. Recall that the generator network architecture is given by equation (\ref{eq:generator_NN_definition}), 
where $\sigma$ is the ReLU activation function. We represent a class of such generators as
\begin{equation}
\label{eq:generator_class}
\begin{aligned}
    \mathcal{G}_{\text{NN}}(\kappa,L,P,K)&=\{g:[0,1]^p\to[0,2C_\mathbf{\Lambda}]^p~\Big|~ g \text{ in form (\ref{eq:generator_NN_definition}) with $L$ layers and max width $P$,}\\
    &\Vert \mathbf{W}_i\Vert_{\infty}\leq \kappa,\Vert \mathbf{b}_i\Vert_{\infty}\leq\kappa,\text{for }i=1,\ldots,L,\sum_{i=1}^L\Vert \mathbf{W}_i\Vert_0+\Vert \mathbf{b}_i\Vert_0\leq K\},
\end{aligned}
\end{equation}
among which the generator function with optimal neural network parameters, $g^*$, is selected via the training process. To theoretically provide convergence guarantee, in this section we specify the discriminator network architecture as 
\begin{equation}
\label{eq:discriminator_NN_definition1}
f(\mathbf{y};\mathbf{W},\mathbf{b})=\mathbf{W}_L \cdot \sigma\left(\mathbf{W}_{L-1}\cdots\sigma\left(\mathbf{W}_2\cdot \tilde{\sigma}\left(\mathbf{W}_{1} \mathbf{y}+\mathbf{b}_{1}\right)+\mathbf{b}_2\right) \cdots +\mathbf{b}_{L-1}\right)+\mathbf{b}_{L}.
\end{equation}
{Note that the ReLU activation function $\sigma$ is used in each layer starting from the second, but in the first layer we use the Softsign activation function $\tilde{\sigma}(x_i):=x_i/(|x_i|+1)$. Roughly speaking, the effect of  $\tilde{\sigma}$ is to map unbounded distributions in $\mathbb{R}^p_+$ into distributions bounded on $[0,1]^p$, while maintaining the discriminative power of the discriminator.  We will further explain the reason for using $\tilde{\sigma}$ in appendix \ref{sec:EC_discriminative power}.}

We represent a class of such discriminators as
\begin{equation}
\label{eq:discriminator_class}
\begin{aligned}
    \mathcal{F}_{\text{NN}}(\kappa,L,P,K,\varepsilon_f)&=\{f_\omega:\mathbb{R}^p\to\mathbb{R}~\Big|~ f_\omega \text{ in form (\ref{eq:discriminator_NN_definition1}) with $L+1$ layers and max width $P$,}\\
    &\Vert \mathbf{W}_i\Vert_{\infty}\leq \kappa,\Vert \mathbf{b}_i\Vert_{\infty}\leq\kappa,\text{for }i=1,\ldots,L,\sum_{i=2}^L\Vert \mathbf{W}_i\Vert_0+\Vert \mathbf{b}_i\Vert_0\leq K,\\
    &\vert f_\omega(x)-f_\omega(y)\vert\leq \Vert x-y\Vert+2\varepsilon_f,\forall x,y\in\mathbb{R}^p\}.
\end{aligned}
\end{equation}
In the following, we sometimes hide the parameters and use $\mathcal{F}_{\text{NN}}$ for short. {We briefly introduce the idea behind the $\varepsilon_f$-constraint (i.e., $\vert f_\omega(x)-f_\omega(y)\vert\leq \Vert x-y\Vert +2\varepsilon_f$). The $\varepsilon_f$-constraint ensures that the metric induced by $\mathcal{F}_{\text{NN}}$ is close to the Wasserstein distance. This further enables us to control the approximation error caused by the difference between these two metrics. Meanwhile, we add the additional $\varepsilon_f$ term instead of restricting $\mathcal{F}_{\text{NN}}$ to be Lipschitz, because the discriminative power of a Lipchitz class cannot be theoretically guaranteed in a desired way.
We will further explain throughout the proof in appendix \ref{ec:thm-dswgan}. Admittedly, the $\varepsilon_f$-constraint is not theoretically guaranteed using standard Wasserstain training algorithm with gradient penalty. We therefore state it as a theoretical possibility.} We next present the main theorem:

\begin{theorem}\label{thm:dswgan}
Suppose that $n$ copies of iid data are available and that the underlying joint density function $\pi_\mathbf{\Lambda}$ satisfies Assumption \ref{assum:density}. 
Let the discriminator and generator architectures be specified as $\mathcal{G}_{\text{NN}}(\bar{\kappa},\bar{L},\bar{P},\bar{K})$ and $\mathcal{F}_{\text{NN}}(\kappa,L,P,K,\varepsilon_f)$ given in (\ref{eq:generator_class}) and (\ref{eq:discriminator_class}), where
\begin{equation*}
    \bar{\kappa}=O(n^{1/p}),\quad \bar{L}=O(\log n),\quad \bar{P}=O(n^{1/p}),\quad \bar{K}=O(n^{1/p}\log n),
\end{equation*}
and
\begin{equation*}
    \kappa=O(n^{1/p}),\quad L=O(\log n),\quad P=O(n^{1/p}),\quad K=O(n^{1/p}\log n),\quad \varepsilon_f=O(n^{-1/p}).
\end{equation*}
Denote ${g}^*_n$ as the generator that optimizes the problem \begin{equation}
    \left(g_n^*, f_n^*\right) \in \argmin_{g_\theta \in \mathcal{G}_{\mathrm{NN}}} \max_{f_\omega \in \mathcal{F}_{\mathrm{NN}}} \E_{\mathbf{Y}\sim \nu} \left[f(\mathbf{h}(g(\mathbf{Y};\theta);\omega ))\right] - \frac{1}{n}\sum_{i=1}^n f(\mathbf{X}_i;\omega),
    \label{eq:WGAN_opt_proof}
\end{equation}
and let $\mu_n^*$ denote the distribution of the random variable
$\mathbf{h}({g}^*_n(\mathbf{Y}))$.
We have 
\begin{equation*}
    \E W(\mu_n^*,\mu) \leq C\cdot n^{-1/p}(\log n)^2,
\end{equation*}
in which $C$ is a constant that is independent of $n$ and depends on $p$. 

Further, as a generalization to the weakly dependent case, if the sample data $(\mathbf{X}_i)_{i=1}^n$ forms a stationary $\rho$-mixing sequence for some $\rho:\mathbb{N}\to\mathbb{R}^+$ satisfying $\sum_{i\geq 0}\rho_i<\infty$, we also have
\begin{equation*}
    \E W(\mu_n^*,\mu) \leq \tilde{C}\cdot n^{-1/p}(\log n)^2,
\end{equation*}
in which $\tilde{C}$ is a constant that is independent of $n$ and depends on $p$ and $\rho$.
\end{theorem}


We now discuss the implications of Theorem \ref{thm:dswgan}. Theorem \ref{thm:dswgan} shows that with appropriately chosen neural network structures, for an arbitrary doubly stochastic Poisson model, if the joint distribution of the underlying stochastic intensities has a smooth density, the DS-WGAN can provide statistically consistent estimated generators. The estimated generator, compounded by the Poisson counts simulator, can simulate new data that is consistent with the true distribution under the Wasserstein distance measurement. The result in Theorem \ref{thm:dswgan} also establishes a guaranteed upper bound on the convergence rate and its dependence on the dimension $p$. This is the first statistical theory on using neural networks to estimate and simulate general doubly stochastic Poisson process models. Admittedly, when $p$ is large, this rate at the order of $n^{-1/p}$ can be slow, and we hope to point out such order is comparable to the best achievable rate for general non-parametric density estimators without stronger smoothness or sparsity assumptions. 

\begin{remark}
When there are an abundant amount of data and the practitioners do not have a specific choice of $p$ in mind, there is a natural question of what is the best choice of $p$ to recommend. A too large $p$ will lead each interval to have too few data points, increasing the variance. A too small $p$ will lead each interval to be too long, which overlooks more of the time-varying features and therefore increasing the bias. An elegant bias-variance trade-off for non-homogeneous Poisson process is introduced in \cite{yuoptimally}. The extension of the analysis in \cite{yuoptimally} to the doubly stochastic Poisson processes appears to be an appealing but challenging topic, because of the complications brought by the correlation structure. We plan to leave this extension as future work.
\end{remark}
\begin{remark}We briefly explain why statistical convergence is guaranteed only by neural networks far bigger than what we use in the numerical experiments. Since we do not make any parametric assumptions on the functional form of the underlying distribution, the theorem needs to provide convergence guarantee for the worst-case scenario. Further, few properties can be used to simplify the theoretical construction of the approximating neural networks. For this reason, the neural network functions $f_\theta$ and $g_\theta$ that are theoretically constructed as solutions to (\ref{eq:WGAN_opt_proof}) may be very different from (and far more complicated than) what is searched and adopted by the networks in the numerical experiments. We also note that certain training algorithms can guide the neural networks to find solutions that are less complicated than the theoretically constructed $f_\theta$ and $g_\theta$, but the effect of algorithms is not incorporated in our theoretical framework of deriving statistical convergence.
\end{remark}

\section{Challenges in the Optimization Problem for Model Estimation}
\label{sec:model_estimation_ds_wgan}
In this section, we provide an algorithm to solve the model estimation optimization problem of DS-WGAN. We discuss and address a unique challenge  in the estimation of DS-WGAN, compared to the model estimation of the classical WGAN. 

Given data $\mathbf{X}_1,\mathbf{X}_2,\ldots,\mathbf{X}_n$, the model estimation of DS-WGAN is to estimate the neural network parameters $\theta$ and $\omega$ for the generator $g(\cdot;\theta)$ and the discriminator $f(\cdot;\omega)$ in the optimization problem (\ref{eq:WGAN_opt}). The associated optimization problem is a non-convex two-player minimax game problem. How to effectively perform model estimation (or training) for such problems attracts keen attention in the machine learning and deep learning area. Specifically, for the model estimation of the classical WGAN without the doubly stochastic feature, variants of stochastic gradient descent (SGD) routines are popular and shown to produce stable performance. See \cite{gulrajani2017improved} for example. The model estimation of DS-WGAN also adopts the SGD routine. 

In order to implement SGD based optimization algorithms, gradients of the objective function with respect to neural network parameters need to be effectively evaluated. We first specify the notation needed for representing the gradients. The gradient of a critical component $f(\mathbf{h}(g(\mathbf{Y};\theta);\omega ))$ in the objective function with respect to the discriminator model parameters $\omega$ is given by $\nabla f_\omega(\mathbf{x};\omega)|_{\mathbf{x}=\mathbf{h}(g(\mathbf{Y};\theta))}$. The gradient of the critical component $f(\mathbf{h}(g(\mathbf{Y};\theta);\omega ))$ with respect to the generator model parameters $\theta$ is given by
\begin{align}
    \nabla_\theta f(\mathbf{h}(g(\mathbf{Y};\theta));\omega)
    =&\nabla_{\mathbf{x}}f(\mathbf{x};\omega)|_{\mathbf{x}=\mathbf{h}(g(\mathbf{Y};\theta))}\cdot\nabla_\theta \mathbf{h}(g(\mathbf{Y};\theta)) \nonumber
    \\=&\nabla_{\mathbf{x}}f(\mathbf{x};\omega)|_{\mathbf{x}=\mathbf{h}(g(\mathbf{Y};\theta))}\cdot\nabla_{\mathbf{\Lambda}}\mathbf{h}(\mathbf{\Lambda})|_{\mathbf{\Lambda}=g(\mathbf{Y};\theta)}\cdot \nabla_\theta g(\mathbf{Y};\theta). \label{eq:backprop_three_terms}
\end{align}
This gradient is evaluated using backpropagation (BP). According to  the chain rule, BP goes backward from the discriminator neural network, through the Poisson counts simulator, and to the generator neural network. The red dashed line in Figure \ref{fig:fram_ds_wgan_backprob} illustrates this gradient propagation evaluation of \eqref{eq:backprop_three_terms}.  We remark that, for the optimization problem \ref{eq:WGAN_opt}, we initially aim for an approximation of the gradient of the expectation,  $\nabla_\theta\mathbb{E}[f(\mathbf{h}(g(\mathbf{Y};\theta)),\omega)]$. This is is not equal to $\mathbb{E}[\nabla_\theta f(\mathbf{h}(g(\mathbf{Y};\theta)),\omega)]$, which can be approximated by taking average of $\nabla_\theta f(\mathbf{h}(g(\mathbf{Y};\theta)),\omega)$ over copies of the random variable $\mathbf{Y}$. Therefore, in the following we do not aim to provide
an unbiased estimator, or to give theoretical justification for the gradient estimator we derive, but instead to provide a heuristic evaluation that would work for practical use. 

\begin{figure}[ht!]
\centering

\tikzset{every picture/.style={line width=0.75pt}} 

\tikzset{every picture/.style={line width=0.75pt}} 

\tikzset{every picture/.style={line width=0.75pt}} 

\begin{tikzpicture}[x=0.75pt,y=0.75pt,yscale=-1,xscale=1]

\draw [color={rgb, 255:red, 118; green, 118; blue, 118 }  ,draw opacity=1 ][line width=0.75]    (168.99,78.37) -- (206.2,78.37) ;
\draw [shift={(208.2,78.37)}, rotate = 180] [color={rgb, 255:red, 118; green, 118; blue, 118 }  ,draw opacity=1 ][line width=0.75]    (10.93,-3.29) .. controls (6.95,-1.4) and (3.31,-0.3) .. (0,0) .. controls (3.31,0.3) and (6.95,1.4) .. (10.93,3.29)   ;
\draw [color={rgb, 255:red, 118; green, 118; blue, 118 }  ,draw opacity=1 ][line width=0.75]    (219.82,78.37) -- (257.03,78.37) ;
\draw [shift={(259.03,78.37)}, rotate = 180] [color={rgb, 255:red, 118; green, 118; blue, 118 }  ,draw opacity=1 ][line width=0.75]    (10.93,-3.29) .. controls (6.95,-1.4) and (3.31,-0.3) .. (0,0) .. controls (3.31,0.3) and (6.95,1.4) .. (10.93,3.29)   ;
\draw [color={rgb, 255:red, 118; green, 118; blue, 118 }  ,draw opacity=1 ][line width=0.75]    (48,78.37) -- (85.21,78.37) ;
\draw [shift={(87.21,78.37)}, rotate = 180] [color={rgb, 255:red, 118; green, 118; blue, 118 }  ,draw opacity=1 ][line width=0.75]    (10.93,-3.29) .. controls (6.95,-1.4) and (3.31,-0.3) .. (0,0) .. controls (3.31,0.3) and (6.95,1.4) .. (10.93,3.29)   ;
\draw [color={rgb, 255:red, 118; green, 118; blue, 118 }  ,draw opacity=1 ][line width=0.75]    (391,197.17) -- (428.58,159.91) ;
\draw [shift={(430,158.5)}, rotate = 495.25] [color={rgb, 255:red, 118; green, 118; blue, 118 }  ,draw opacity=1 ][line width=0.75]    (10.93,-3.29) .. controls (6.95,-1.4) and (3.31,-0.3) .. (0,0) .. controls (3.31,0.3) and (6.95,1.4) .. (10.93,3.29)   ;
\draw [color={rgb, 255:red, 118; green, 118; blue, 118 }  ,draw opacity=1 ][line width=0.75]    (391,78.5) -- (428.58,115.76) ;
\draw [shift={(430,117.17)}, rotate = 224.75] [color={rgb, 255:red, 118; green, 118; blue, 118 }  ,draw opacity=1 ][line width=0.75]    (10.93,-3.29) .. controls (6.95,-1.4) and (3.31,-0.3) .. (0,0) .. controls (3.31,0.3) and (6.95,1.4) .. (10.93,3.29)   ;
\draw [color={rgb, 255:red, 118; green, 118; blue, 118 }  ,draw opacity=1 ][line width=0.75]    (511.99,137.37) -- (549.2,137.37) ;
\draw [shift={(551.2,137.37)}, rotate = 180] [color={rgb, 255:red, 118; green, 118; blue, 118 }  ,draw opacity=1 ][line width=0.75]    (10.93,-3.29) .. controls (6.95,-1.4) and (3.31,-0.3) .. (0,0) .. controls (3.31,0.3) and (6.95,1.4) .. (10.93,3.29)   ;
\draw [color={rgb, 255:red, 118; green, 118; blue, 118 }  ,draw opacity=1 ][line width=0.75]    (340.55,78.37) -- (377.76,78.37) ;
\draw [shift={(379.76,78.37)}, rotate = 180] [color={rgb, 255:red, 118; green, 118; blue, 118 }  ,draw opacity=1 ][line width=0.75]    (10.93,-3.29) .. controls (6.95,-1.4) and (3.31,-0.3) .. (0,0) .. controls (3.31,0.3) and (6.95,1.4) .. (10.93,3.29)   ;
\draw  [color={rgb, 255:red, 153; green, 0; blue, 1 }  ,draw opacity=1 ][fill={rgb, 255:red, 244; green, 204; blue, 205 }  ,fill opacity=1 ][line width=0.75]  (37.07,28.12) .. controls (37.07,28.12) and (37.07,28.12) .. (37.07,28.12) -- (47.25,28.12) .. controls (47.25,28.12) and (47.25,28.12) .. (47.25,28.12) -- (47.25,128.62) .. controls (47.25,128.62) and (47.25,128.62) .. (47.25,128.62) -- (37.07,128.62) .. controls (37.07,128.62) and (37.07,128.62) .. (37.07,128.62) -- cycle ;
\draw  [color={rgb, 255:red, 56; green, 118; blue, 30 }  ,draw opacity=1 ][fill={rgb, 255:red, 217; green, 234; blue, 211 }  ,fill opacity=1 ][line width=0.75]  (208.9,28.12) .. controls (208.9,28.12) and (208.9,28.12) .. (208.9,28.12) -- (219.08,28.12) .. controls (219.08,28.12) and (219.08,28.12) .. (219.08,28.12) -- (219.08,128.62) .. controls (219.08,128.62) and (219.08,128.62) .. (219.08,128.62) -- (208.9,128.62) .. controls (208.9,128.62) and (208.9,128.62) .. (208.9,128.62) -- cycle ;
\draw  [color={rgb, 255:red, 118; green, 118; blue, 118 }  ,draw opacity=1 ][line width=0.75]  (87.46,36.67) .. controls (87.46,31.95) and (91.29,28.12) .. (96.01,28.12) -- (160.36,28.12) .. controls (165.08,28.12) and (168.9,31.95) .. (168.9,36.67) -- (168.9,120.08) .. controls (168.9,124.8) and (165.08,128.62) .. (160.36,128.62) -- (96.01,128.62) .. controls (91.29,128.62) and (87.46,124.8) .. (87.46,120.08) -- cycle ;
\draw  [color={rgb, 255:red, 118; green, 118; blue, 118 }  ,draw opacity=1 ][line width=0.75]  (259.27,36.67) .. controls (259.27,31.95) and (263.1,28.12) .. (267.82,28.12) -- (332.17,28.12) .. controls (336.89,28.12) and (340.72,31.95) .. (340.72,36.67) -- (340.72,120.08) .. controls (340.72,124.8) and (336.89,128.62) .. (332.17,128.62) -- (267.82,128.62) .. controls (263.1,128.62) and (259.27,124.8) .. (259.27,120.08) -- cycle ;
\draw  [color={rgb, 255:red, 10; green, 83; blue, 148 }  ,draw opacity=1 ][fill={rgb, 255:red, 201; green, 218; blue, 248 }  ,fill opacity=1 ][line width=0.75]  (380.51,28.12) .. controls (380.51,28.12) and (380.51,28.12) .. (380.51,28.12) -- (390.69,28.12) .. controls (390.69,28.12) and (390.69,28.12) .. (390.69,28.12) -- (390.69,128.62) .. controls (390.69,128.62) and (390.69,128.62) .. (390.69,128.62) -- (380.51,128.62) .. controls (380.51,128.62) and (380.51,128.62) .. (380.51,128.62) -- cycle ;
\draw  [color={rgb, 255:red, 118; green, 118; blue, 118 }  ,draw opacity=1 ][line width=0.75]  (430.42,95.67) .. controls (430.42,90.95) and (434.24,87.12) .. (438.96,87.12) -- (503.31,87.12) .. controls (508.03,87.12) and (511.86,90.95) .. (511.86,95.67) -- (511.86,179.08) .. controls (511.86,183.8) and (508.03,187.62) .. (503.31,187.62) -- (438.96,187.62) .. controls (434.24,187.62) and (430.42,183.8) .. (430.42,179.08) -- cycle ;
\draw  [color={rgb, 255:red, 10; green, 83; blue, 148 }  ,draw opacity=1 ][fill={rgb, 255:red, 103; green, 199; blue, 215 }  ,fill opacity=0.31 ][line width=0.75]  (380.41,146.12) .. controls (380.41,146.12) and (380.41,146.12) .. (380.41,146.12) -- (390.59,146.12) .. controls (390.59,146.12) and (390.59,146.12) .. (390.59,146.12) -- (390.59,246.62) .. controls (390.59,246.62) and (390.59,246.62) .. (390.59,246.62) -- (380.41,246.62) .. controls (380.41,246.62) and (380.41,246.62) .. (380.41,246.62) -- cycle ;
\draw [color={rgb, 255:red, 153; green, 0; blue, 1 }  ,draw opacity=1 ] [dash pattern={on 4.5pt off 4.5pt}]  (591.08,124.18) .. controls (591.08,111.18) and (584.08,111.18) .. (558.08,111.78) .. controls (532.08,112.37) and (485.7,111.65) .. (476,111.4) .. controls (466.3,111.15) and (445.6,103.4) .. (431.6,89) .. controls (417.6,74.6) and (409.2,57) .. (384.8,57) .. controls (360.52,57) and (193.56,58.17) .. (118.21,58.18) ;
\draw [shift={(117.08,58.18)}, rotate = 360] [color={rgb, 255:red, 153; green, 0; blue, 1 }  ,draw opacity=1 ][line width=0.75]    (10.93,-3.29) .. controls (6.95,-1.4) and (3.31,-0.3) .. (0,0) .. controls (3.31,0.3) and (6.95,1.4) .. (10.93,3.29)   ;
\draw [color={rgb, 255:red, 153; green, 0; blue, 1 }  ,draw opacity=1 ] [dash pattern={on 4.5pt off 4.5pt}]  (241.33,161) -- (172.35,161) ;
\draw [shift={(170.35,161)}, rotate = 360] [color={rgb, 255:red, 153; green, 0; blue, 1 }  ,draw opacity=1 ][line width=0.75]    (10.93,-3.29) .. controls (6.95,-1.4) and (3.31,-0.3) .. (0,0) .. controls (3.31,0.3) and (6.95,1.4) .. (10.93,3.29)   ;

\draw (90.18,67.87) node [anchor=north west][inner sep=0.75pt]  [font=\fontsize{0.6em}{0.72em}\selectfont] [align=left] {\begin{minipage}[lt]{56.229375000000005pt}\setlength\topsep{0pt}
\begin{center}
{\fontfamily{ptm}\selectfont Neural-network}\\{\fontfamily{ptm}\selectfont based generator }$\displaystyle g$
\end{center}

\end{minipage}};
\draw (271.99,68.37) node [anchor=north west][inner sep=0.75pt]  [font=\fontsize{0.6em}{0.72em}\selectfont] [align=left] {\begin{minipage}[lt]{39.280625pt}\setlength\topsep{0pt}
\begin{center}
{\fontfamily{ptm}\selectfont Poisson counts}\\{\fontfamily{ptm}\selectfont simulator $\mathbf{h}$}
\end{center}

\end{minipage}};
\draw (2.66,12.58) node [anchor=north west][inner sep=0.75pt]  [font=\fontsize{0.6em}{0.72em}\selectfont] [align=left] {\begin{minipage}[lt]{56.24pt}\setlength\topsep{0pt}
\begin{center}
{\fontfamily{ptm}\selectfont Random noises $\mathbf{Y}$}
\end{center}

\end{minipage}};
\draw (175.99,1.58) node [anchor=north west][inner sep=0.75pt]  [font=\fontsize{0.6em}{0.72em}\selectfont] [align=left] {\begin{minipage}[lt]{58.120625000000004pt}\setlength\topsep{0pt}
\begin{center}
{\fontfamily{ptm}\selectfont Simulated Poisson}\\{\fontfamily{ptm}\selectfont rates }$\displaystyle g( \mathbf{Y};\theta )$
\end{center}

\end{minipage}};
\draw (337.1,1.58) node [anchor=north west][inner sep=0.75pt]  [font=\fontsize{0.6em}{0.72em}\selectfont] [align=left] {\begin{minipage}[lt]{72.549375000000005pt}\setlength\topsep{0pt}
\begin{center}
{\fontfamily{ptm}\selectfont Simulated arrival counts}\\{\fontfamily{ptm}\selectfont  vector }$\displaystyle \mathbf{h}( g( \mathbf{Y};\theta ))$
\end{center}

\end{minipage}};
\draw (333,252.5) node [anchor=north west][inner sep=0.75pt]  [font=\fontsize{0.6em}{0.72em}\selectfont] [align=left] {\begin{minipage}[lt]{75.76875000000001pt}\setlength\topsep{0pt}
\begin{center}
{\fontfamily{ptm}\selectfont Real arrival counts vector}\\{\fontfamily{ptm}\selectfont from training data set}
\end{center}

\end{minipage}};
\draw (162.5,157) node [anchor=north west][inner sep=0.75pt]  [font=\fontsize{0.6em}{0.72em}\selectfont,color={rgb, 255:red, 153; green, 0; blue, 1 }  ,opacity=1 ] [align=left] {{\fontfamily{ptm}\selectfont `` \ \ \ \ \ \ \ \ \ \ \ \ \ \ \ \ \ \ \ \ \ \ \ \ \ \ \ \ \ \ \ \ \ \ '' : Gradient backpropagation}};
\draw (430.67,127.37) node [anchor=north west][inner sep=0.75pt]  [font=\fontsize{0.6em}{0.72em}\selectfont] [align=left] {\begin{minipage}[lt]{59.389375pt}\setlength\topsep{0pt}
\begin{center}
{\fontfamily{ptm}\selectfont Neural-network}\\{\fontfamily{ptm}\selectfont based discriminator f}
\end{center}

\end{minipage}};
\draw (548.28,129.58) node [anchor=north west][inner sep=0.75pt]  [font=\fontsize{0.6em}{0.72em}\selectfont] [align=left] {\begin{minipage}[lt]{58.439375000000005pt}\setlength\topsep{0pt}
\begin{center}
{\fontfamily{ptm}\selectfont {\fontsize{1em}{1.2em}\selectfont Objective function}}\\{\fontfamily{ptm}\selectfont {\fontsize{1em}{1.2em}\selectfont (Equation (10))}}
\end{center}

\end{minipage}};

\end{tikzpicture}

\caption{Backpropagation diagram (red dashed line) on the gradient evaluation of $f(\mathbf{h}(g(\mathbf{Y};\theta);\omega ))$ with respect to the parameters $\theta$ in the generator neural network $g$.} 
\label{fig:fram_ds_wgan_backprob}
\end{figure}
For the gradient evaluation of $ \nabla_\theta f(\mathbf{h}(g(\mathbf{Y};\theta));\omega)$ in equation (\ref{eq:backprop_three_terms}), the product of three parts are involved. The first part $\nabla_{\mathbf{x}}f(\mathbf{x};\omega)|_{\mathbf{x}=\mathbf{h}(g(\mathbf{Y};\theta))}$ and the third part $\nabla_\theta g(\mathbf{Y};\theta)$ can be effectively computed by BP, which is standard in the literature and can be conveniently implemented in libraries such as PyTorch. The key challenge is the second part $\nabla_{\mathbf{\Lambda}}\mathbf{h}(\mathbf{\Lambda})|_{\mathbf{\Lambda}=g(\mathbf{Y};\theta)}$, which is not well-defined. Specifically, for the Poisson counts simulator $\mathbf{h}$ as defined in (\ref{eq:poisson_count_simulator}), the input variable $\mathbf{\Lambda}=(\Lambda_1,\Lambda_2,\ldots,\Lambda_p)$ and the output variable $\mathbf{h}(\mathbf{\Lambda})=(M_1(\Lambda_1),M_2(\Lambda_2),\ldots,M_p(\Lambda_p))$ are both vectors of size $p\times 1$. Recall that the Poisson count $M_{1}\left(\Lambda_1\right)$ is a random function of a random variable $\Lambda_1$. Therefore, the derivative $\frac{\partial}{\partial \Lambda_{1}} M_{1}\left(\Lambda_1\right)$ adopts a sample-path definition in the BP routine. In fact, because $M_1$ outputs discrete variables from $\{0,1,2,\cdots\}$ in each sample path, the derivative $\frac{\partial}{\partial \Lambda_{1}} M_{1}\left(\Lambda_1\right)$ is either zero or infinity.\footnotemark[2]\footnotetext[2]{
The need to take derivative of continuous functions with respect to discrete variables arise in the use of generative adversarial networks in text generation and variational autoencoder with discrete latent variable. A reparametrization trick using Gumbel-Softmax is discussed in \cite{jang2016categorical} and \cite{maddison2016concrete} to approximate the derivative of continuous functions with respect to discrete variables. The challenge in our case, differently, is to take derivative of discrete functions with respect to continuous variables.} To avoid gradients of zero or infinity, we provide an approximation to replace the derivative $\frac{\partial}{\partial \lambda} M_{1}\left(\lambda\right)$. That is, 
$$\frac{\partial}{\partial \lambda} M_{1}\left(\lambda\right) \overset{\Delta}{\approx} 1+ \frac{ M_1(\lambda)- \lambda}{2\lambda}.$$
Using such approximation in all $p$ dimensions gives the approximated $\nabla_{\mathbf{\Lambda}}\mathbf{h}(\mathbf{\Lambda})$.

We next provide reasons for using this approximation. By applying Berry-Esseen theorem, we have 
\begin{equation*}
    |F_\lambda(x)-\Phi(x)|\leq \frac{C}{\sqrt{\lambda}},
\end{equation*}
where $F_\lambda(x)$ is the cumulative distribution function of $M_1(\lambda)/\sqrt{\lambda}$, $\Phi(x)$ is the c.d.f. of the standard normal distribution, and $C<2.5$ according to \cite{shevtsova2011absolute}. Consider approximating the Poisson random variable $M_1(\lambda)$ with a normal random variable with distribution $\mathcal{N}(\lambda, \lambda)$, denoted as $Z_\lambda$. The difference in cumulative distribution function between these two random variables is also bounded by $C\mathbb{E}X_1^3/\sqrt{\lambda}$. Also, taking derivative of $Z_\lambda$ w.r.t. $\lambda$ yields
\begin{equation*}
\frac{dZ_\lambda}{d\lambda}=\frac{d}{d\lambda}(\lambda+\sqrt{\lambda}Z)=1+\frac{Z}{2\sqrt{\lambda}}=1+\frac{\sqrt{\lambda}Z}{2\lambda}=1+\frac{Z_\lambda-\lambda}{2\lambda},
\end{equation*}
where $Z$ is the standard $\mathcal{N}(0,1)$ random variable. Our gradient approximation method is to plug in $M_1(\lambda)$ to replace $Z_\lambda$ in the equation. 

We note that other gradient approximations may be used, such as variants of finite difference and likelihood ratio methods. We refer to \cite{l1990unified}, \cite{vazquez1991comparing} and \cite{l1994convergence} for an overview of the methods.
We hope to point out that our approximation does not give an unbiased gradient estimator. The difficulty in developing an unbiased gradient estimator lies in that the Poisson simulator $\mathbf{h}$ is compounded with functions $f(\cdot)$ and $g(\mathbf{Y};\theta)$.
It is worth further investigation whether an unbiased gradient estimator  can be derived and efficiently integrated in the training algorithms. For example, one possible idea is take conditional expectation conditional on $\mathbf{\Lambda}$ and then compute the gradient of the conditional expectation.\footnotemark[3]\footnotetext[3]{We thank an anonymous reviewer for suggesting this idea of exploring an unbiased gradient estimator.}

We next list the SGD algorithm for the model estimation as the following Algorithm \ref{alg:ds_wgan}. This algorithm integrates the use of gradient penalty to enforce the Lipschitz constraint on the discriminator. The algorithm updates the parameters of the generator and discriminator neural networks in an iterative manner. 

\begin{algorithm}[ht!] 
\caption{Model estimation of DS-WGAN}
\label{alg:ds_wgan}
\begin{algorithmic}[1]
\REQUIRE The gradient penalty coefficient $\zeta$, 
the batch size $m$, the number of iterations of the discriminator per generator iteration $n_{\text{dis}}$,
Adam hyper-parameters $\alpha$, $\beta_1$ and $\beta_2$, \textcolor{black}{maximum iteration step $n_{\text{maxit}}$.}\\
\REQUIRE $w_{0}$, initial discriminator parameters. $\theta_{0}$, initial generator's parameters.
\FOR{\textcolor{black}{$k=1,2,\ldots,n_\text{maxit}$}}
    \FOR{$t=1,2,  \ldots, n_{\text {dis }}$}
        \STATE Sample a batch of real arrival count vectors from the training data set, $\{\mathbf{x}^{(i)}\}^{m}_{i=1}\sim \hat{\mu}_n$.
        \STATE Sample a batch of random noises independently, $\mathbf{y}^{(i)}\sim \nu,\text{ for }i=1,2,\ldots,m$.
        \STATE $\tilde{\mathbf{x}}^{(i)}\leftarrow\mathbf{h}(g(\mathbf{y}^{(i)};\theta)),\text{ for }i=1,2,\ldots,m$.
        \STATE Sample random number $\varepsilon^{(i)} \sim U[0,1],\text{ for }i=1,2,\ldots,m.$
        \STATE  $\hat{\mathbf{x}}^{(i)} \leftarrow \varepsilon^{(i)} \mathbf{x}^{(i)}+(1-\varepsilon^{(i)}) \tilde{\mathbf{x}}^{(i)},\text{ for }i=1,2,\ldots,m.$
        \STATE $\text{Gradient penalty}\leftarrow \zeta \cdot\left(\left\|\nabla_{\hat{\mathbf{x}}^{(i)}} f\left(\hat{\mathbf{x}}^{(i)} ; \omega\right)\right\|_{2}-1\right)^{2}$
        \STATE $L_{\omega}\leftarrow \sum_{i=1}^m f\left(\tilde{\mathbf{x}}^{(i)} ; \omega\right)-\sum_{i=1}^m f\left(\mathbf{x}^{(i)} ; \omega\right)+\text{Gradient penalty}$
        
        \STATE $\omega \leftarrow \omega+\operatorname{Adam}\left(L_{\omega}, \omega, \alpha, \beta_{1}, \beta_{2}\right)$
    \ENDFOR

    \STATE Sample a batch of random noises independently, $\mathbf{y}^{(i)}\sim \nu,\text{ for }i=1,2,\ldots,m$.
    \STATE $\tilde{\mathbf{x}}^{(i)}\leftarrow\mathbf{h}(g(\mathbf{y}^{(i)};\theta)),\text{ for }i=1,2,\ldots,m$.
    \STATE $L_\theta \leftarrow \sum_{i=1}^m f\left(\mathbf{h}\left(g\left(\mathbf{y}^{(i)};\theta\right)\right);\omega\right)$
    \STATE  $\theta \leftarrow \theta -  \operatorname{Adam}\left(\nabla_{\theta} L_\theta, \theta, \alpha, \beta_{1}, \beta_{2}\right)$
\ENDFOR
\end{algorithmic}
\end{algorithm}

\section{Numerical Experiments}
\label{sec:experiments}
\textcolor{black}{In this section, we test the performance of the DS-WGAN framework, using two sets of synthetic data (Section \ref{sec:exper_dspp_cir} and  \ref{sec:exper_pierre}) and two sets of real data (Section \ref{sec:exper_callcenter_real} and Section \ref{sec:exper_bikeshare_real})}. In Section \ref{sec:exper_dspp_cir}, we introduce run-through-queue experiments that demonstrate the performance of DS-WGAN from an operational performance point of view. {In Section  \ref{sec:exper_callcenter_real} and \ref{sec:exper_bikeshare_real}, we also compare the performance of DS-WGAN with one of the state-of-the-art methods, PGnorta, introduced by \cite{oreshkin2016rate}. Additionally, in section \ref{sec:exper_downsample} we demonstrate how reducing the model resolution $p$ influences the estimation results.} All the examples in this section consider a single call type. The code written for this section can be found at \url{https://github.com/DDDOH/Doubly-Stochastic-Simulator}.

\subsection{DSPP with Diffusion Intensity and Run-Through-Queue Experiments}
\label{sec:exper_dspp_cir}
In this subsection, we use a set of synthetic arrival data and run-through-queue experiments to test the performance of DS-WGAN. We note that one major advantage of using synthetic data is that we have a known ground truth to compare with.
\subsubsection{Underlying Model for Synthetic Data: DSPP with Diffusion Intensity}\label{sec:cir_model}
The underlying arrival process $N=(N(t):t\in[0,T])$ is a doubly stochastic Poisson process (DSPP) whose intensity process $\lambda=(\lambda(t):t\in[0,T])$ is given by a diffusion process called Cox-Ingersoll-Ross (CIR) process. Specifically, the intensity process is given by 
\begin{equation}
\label{eq:CIR_model}
    \mathrm{d} \lambda(t)=\kappa(R(t)-\lambda(t)) \mathrm{d} t+\sigma R(t)^{\alpha} \lambda(t)^{1/2}\mathrm{d} B(t), 
\end{equation}
where $\kappa$, $\sigma$, $\alpha$ are positive constants and $B(t)$ is a standard Brownian motion that is generated independent of other randomness in $(N(t):t\in[0,T])$. The process $(R(t):t\in[0,T])$ is a deterministic function of time that captures the time-of-day effect. To fully reflect the doubly stochastic nature, the initial value $\lambda(0)$ is set as $R(0)\cdot\text{Gamma}(\beta,\beta)$, in which for any positive parameters $a$ and $b$, $\text{Gamma}(a,b)$ denotes a gamma-distributed random variable with mean $a / b$ and variance $a / b^{2}$. This DSPP model with a non-stationary CIR process as the intensity process is an extension of the model provided in Section 3 of \cite{zhang2014scaling}, in which they $R(t)$ is set as a constant. As shown in \cite{zhang2014scaling}, this DSPP model with CIR intensity process is aligned with some real data sets and well captures the time-varying covariance structure presented in those real data sets. We therefore generate synthetic data from this underlying model to test the performance of DS-WGAN.  

\begin{figure}[ht!]
\centering
\includegraphics[scale=0.5]{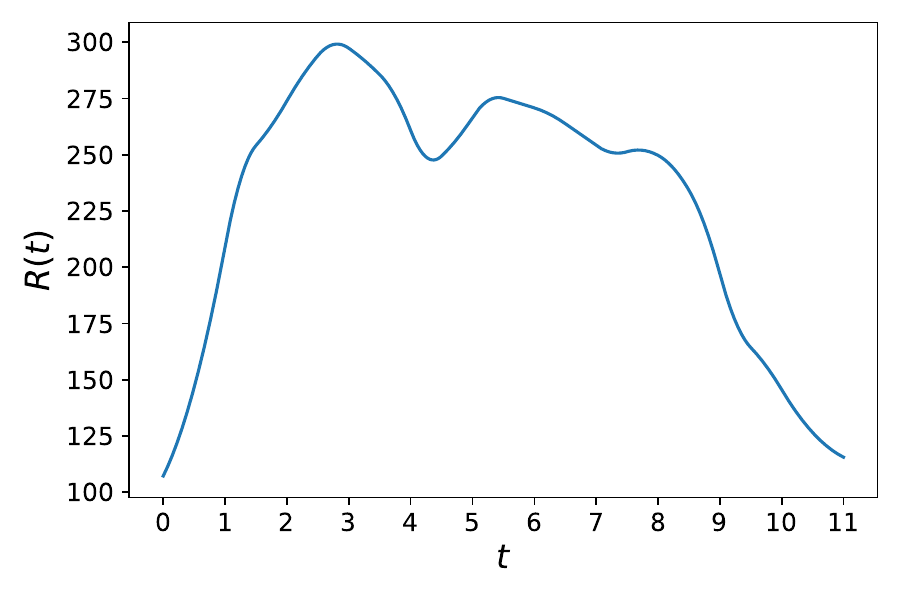}
\caption{The figure of $R(t)$ as a function of $t$}
\label{fig:R_t}
\end{figure}

The model parameters are set as $\kappa=0.2,\sigma=0.4,\alpha=0.3,\beta=100$ and the time-of-day pattern function $(R(t):t\in[0,T])$ is given in Figure \ref{fig:R_t}, which roughly calibrates the time-of-day pattern in a call center data set used in Section 6.2 of \cite{l2018modeling}. This center operates 11 hours a day so $T$ is set as $11$. For the data generation process, we simulate $n=300$ replications of $N=(N(t):t\in[0,T])$. In each replication, the underlying CIR process is simulated using the Euler–Maruyama discretization method (\cite{kloeden2013numerical}) at a resolution of $\delta = 0.001$. The generation of Poisson process conditional on the simulated intensity process is through the thinning method (\cite{lewis1979simulation}). For the $i$-th replication of $N=(N(t):t\in[0,T])$, where $i=1,2,\ldots,n$, we collect the consecutive half-hour interval counts and record as $\mathbf{X}_i = (X_{i1},X_{i2},\ldots,X_{ip})$, in which $p=22$. 

The data set $\{\mathbf{X}_i\}_{i=1}^n$ is used as the input data for DS-WGAN. In the DS-WGAN framework, the parametrization of the generator neural network is given by $L=4$, $\tilde{n}=\left(n_{1}, n_{2},n_{3},n_4\right)=(512,512,512,22)$. The parametrization of the discriminator neural network is given by $L=4$, and $\tilde{n}=\left(n_{1}, n_{2},n_{3},n_4\right)=(512,512,512,1)$. We find that moderately changing the structure setting $L$ and $\tilde{n}$ will not change much of the performance, but it remains an open question about how small the values can be to ensure performances comparable to what can be achieved by the above structure setting. The initialization of all the elements of the weight matrices $\mathbf{W}_l$'s of both the generator and discriminator networks are given by independent Gaussian random variables with mean 0 and variance 0.1. The vectors $\mathbf{b}_l$'s are initialized as constant 3. The neural network training is carried out with 50,000 iterations using the Adam optimizer (\cite{kingma2014adam}). The parameters for Adam optimizer are set as $\beta_1=0.5, \beta_2=0.9$ and the generator batch size is set as 256. The gradient penalty coefficient is set as 0.5. The discriminator is updated 10 times per generator iteration. An exponentially decaying learning rate is used for both the generator and discriminator, ranging from from 1e-4 to 1e-6. The training is implemented with PyTorch. The training takes about 50 minutes with one Nvidia K80 GPU.

With the trained DS-WGAN model, we can generate new iid copies of the vector of half-hour counts using the trained generator and the Poisson counts simulator. We represent a vector of half-hour counts generated by the trained DS-WGAN model as $\tilde{\mathbf{X}} = (\tilde{X}_1,\tilde{X}_2,\ldots,\tilde{X}_p)$ with $p=22$. Given a copy of the generated vector $\tilde{\mathbf{X}}$, we can use the piecewise linear version of the arrival epochs simulator in Section \ref{sec:model_setup} to generate a full sample path of arrival epochs on $[0,T]$. This full sample path generated by DS-WGAN and the arrival epoch simulator is a realization of a point process, which is denoted as $\tilde{N} = \{\tilde{N}(t):t\in[0,T]\}$. Denote $\mathbf{X}=(X_1,X_2,\ldots,X_p)$ as the vector of half-hour counts that reflect the true underlying DSPP model. {We design  two run-through-queue experiments to demonstrate that the probability measure induced by $\tilde{N} = \{\tilde{N}(t):t\in[0,T]\}$ matches that of the true underlying process $N=\{N(t):t\in[0,T]\}$. This can imply that the joint distribution of $\tilde{\mathbf{X}}$ matches that of $\mathbf{X}$, and further, the piecewise-linear rate is a good approximation of the underlying arrival rate process. }

Section \ref{sec:infinite_server_queue} involves an infinite-server queue, with performance measure as the distribution of number of people in the system at different times. Section \ref{sec:many_server_queue} involves a more realistic many-server queue, with performance measure as the distribution of waiting time in system at different times. The general idea is that if the DS-WGAN generated arrival process $\tilde{N}$ and the true underlying process $N$ have the same probability measure, the queueing performances for them when they are run through the same queueing system should resemble, serving as a necessity check. When we have data on the specific arrival times, such comparison is available.

\subsubsection{Run-through-queue Experiments: Infinite-server Queue}
\label{sec:infinite_server_queue}
\begin{figure}[ht!]
\centering 
\subfigure[Mean of number of occupied servers]{
\begin{minipage}{0.4\textwidth}
\centering                                           \includegraphics[scale=0.4]{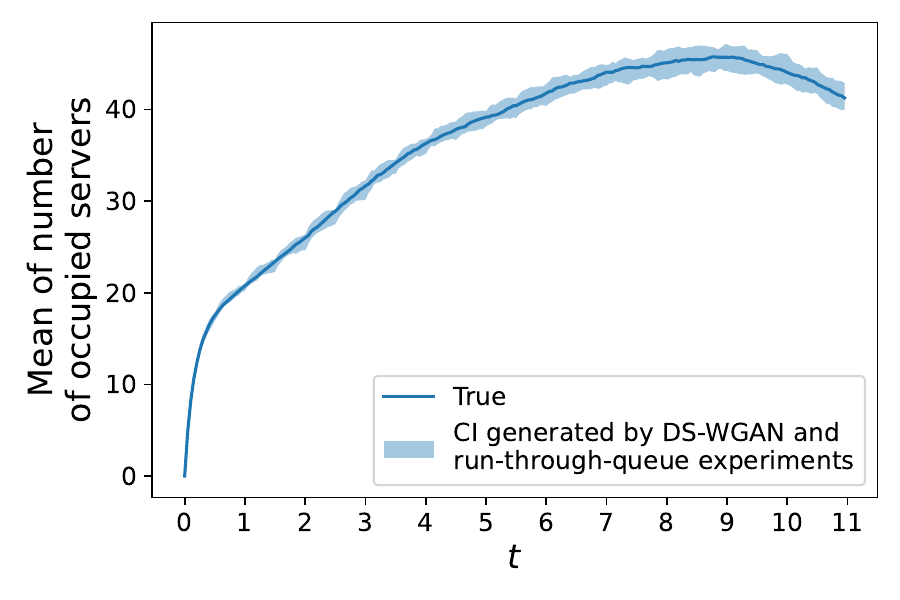}               
\end{minipage}}
\subfigure[Variance of number of occupied servers.]
{\begin{minipage}{0.4\textwidth}
\centering                                           \includegraphics[scale=0.4]{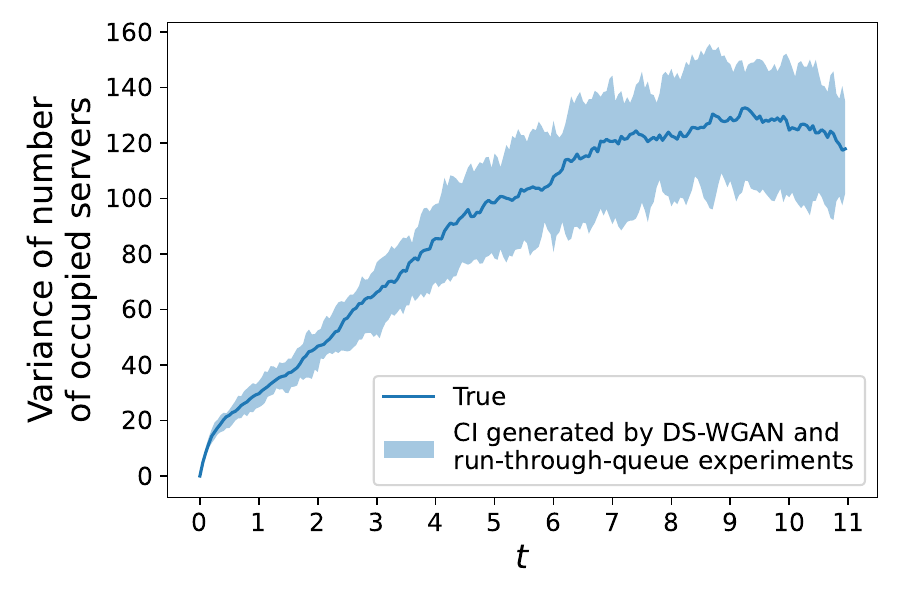}                
\end{minipage}}
\caption{Run-through queue experiment results of Section \ref{sec:infinite_server_queue}. Mean and variance of number of occupied servers at any time for the same Queue fed by the trained DS-WGAN model and the true underlying model. \textmd{The blue solid polylines are the true values. The light blue areas are the confidence intervals given by DS-WGAN model and run-through-queue experiments.}}
\label{fig:CIR}                                   
\end{figure}

Consider an infinite-server queue, in which the service time distribution is set as a log-normal distribution with mean 0.2 (in hours) and variance 0.1. We use the trained DS-WGAN model to generate $n=300$ iid copies of arrival process $\tilde{N} = \{\tilde{N}(t):t\in[0,T]\}$. We run each of the generated arrival processes into the infinite-server queue, generating $n=300$ independent realizations of the queueing operations on $[0,T]$. For results demonstration, we record the number of customers $V(t)$ in the system at time $t$, for which $t$ ranges over the end of each minute. We compute the expectation and variance of $V(t)$ as a function of time using the $n=300$ realizations of the queueing operations. We compute a 95\% confidence interval (CI) for both the expectations and variances by repeating the experiments for 100 replications, and plot the confidence interval in light blue in Figure \ref{fig:CIR}. We also plot the benchmark value of expectations of variances in blue solid polylines. The benchmark value is estimated from using the true underlying arrival process model, by independently simulating 3,000 replications of the queueing operations, discretized at the resolution of $\delta = 0.001$. Figure \ref{fig:CIR} shows that the queueing performance associated with the arrival process generated by the trained DS-WGAN model matches the benchmark. 

\subsubsection{Run-through-queue Experiments: Many-server Queue}
\label{sec:many_server_queue}
Compared to an infinite-server queue, a many-server queue with finite number of servers is a more realistic setting for many applications, especially in the service systems. In those systems, customers usually have to wait due to the limitation in service capacity. The probability distributions of customers waiting time are of high interest to the modeler, as they often represent customer satisfaction and largely impact customer retention. Driven by this need, we present a run-through-queue experiment using a many-server queue with two different staffing plans of servers. We remark that the purpose of this experiment is not to select an optimal staffing plan under some objective function. Instead, we aim to demonstrate that for a queue with a given staffing plan, the customer waiting time performance calculated from the DS-WGAN generated arrival process (which is fed into the queue) is close to the true benchmark. 

We consider a many-server queue, in which probability distribution of service time requirement for each customer is set as a log-normal distribution with mean 0.1 (in hours) and variance 0.1. The underlying customer arrival process, the synthetic data generation procedure, and the DS-WGAN model estimation procedure are the same as Section \ref{sec:cir_model} and Section \ref{sec:infinite_server_queue}. The only difference is in the queueing staffing setting. Instead of an infinite number of servers, we consider a finite number of servers. We select two staffing plans for which the number of servers are changing with time. Let $s(t)$ denote the number of servers staffed in the queueing system as a function of time, with $t\in[0,11]$. In both staffing plans, $s(t)$ is a piecewise constant function. Specifically,  $s(t)$ is constant with value $\{s_i\}$ on the time interval $[(i-1)*0.5, i*0.5)$, $i=1,2,\ldots,22$.

We next specify the value of $s_i$'s of the two staffing plans. Denote $R_i = \int_{(i-1)*0.5}^{i*0.5} R(t)\,dt$. Recall that $R(t)$ comes from the underlying CIR process (\ref{eq:CIR_model}). We remark that the same staffing plans are used on both the synthetic queueing data and the DS-WGAN-generated queueing data. The value of $R(t)$ is not known to or estimated by DS-WGAN, and the staffing plan is not an outcome of the estimation result. Instead, it is an auxiliary that helps to construct a distribution of interest from the generated data, in order to feature the estimation performance of DS-WGAN.

The first staffing plan in consideration is given by a square-root formula 
\begin{equation*}
    s_i = R_i \E(S) + \beta \, \sqrt{R_i\E(S)} 
\end{equation*}
in which $S$ in the random variable denoting the service time and $\beta = 1$. The second staffing plan is given by a formula considering the higher variability in arrivals for DSPP arrivals. This staffing plan is discussed in \cite{zhang2014scaling}. Specifically, the second staffing plan is 
\begin{equation*}
    s_i = R_i \E(S) + \beta \, {(R_i\E(S))}^{\frac{1}{2}+\alpha} 
\end{equation*}
in which $S$ in the random variable denoting the service time and $\beta = 1$. The parameter $\alpha$ is the variability parameter given in the random CIR arrival rate process, shown in equation (\ref{eq:CIR_model}). 

\begin{figure}[ht!]
\centering 
\begin{minipage}{\textwidth}
    \subfigure[Mean of the average waiting time $\Bar{W}_i$.]
    {\begin{minipage}{0.49\textwidth}
    \centering                                           \includegraphics[scale=0.4]{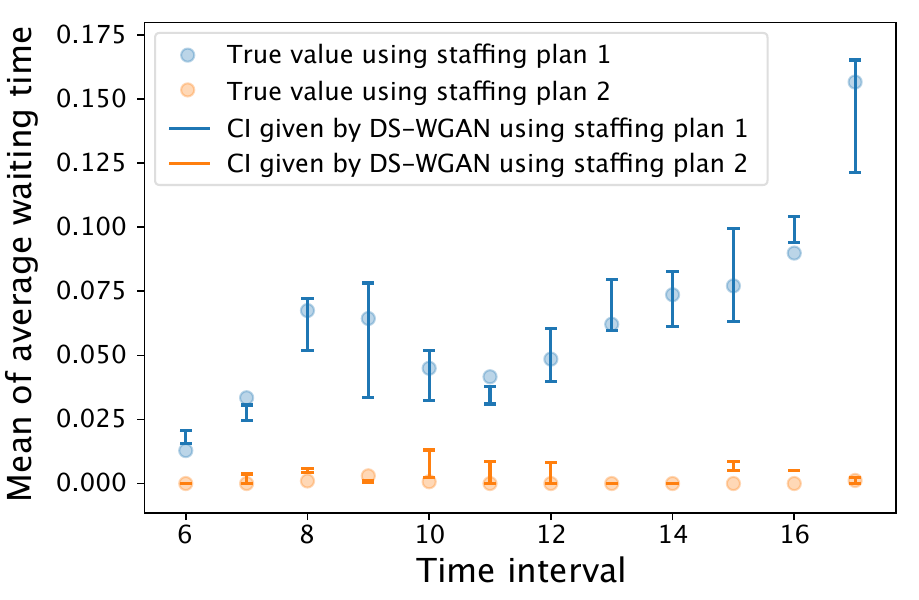}              \label{fig:multi_server_mean}
    \end{minipage}}
    \subfigure[Variance of the average waiting time $\Bar{W}_i$.]
    {\begin{minipage}{0.49\textwidth}
    \centering                                           \includegraphics[scale=0.4]{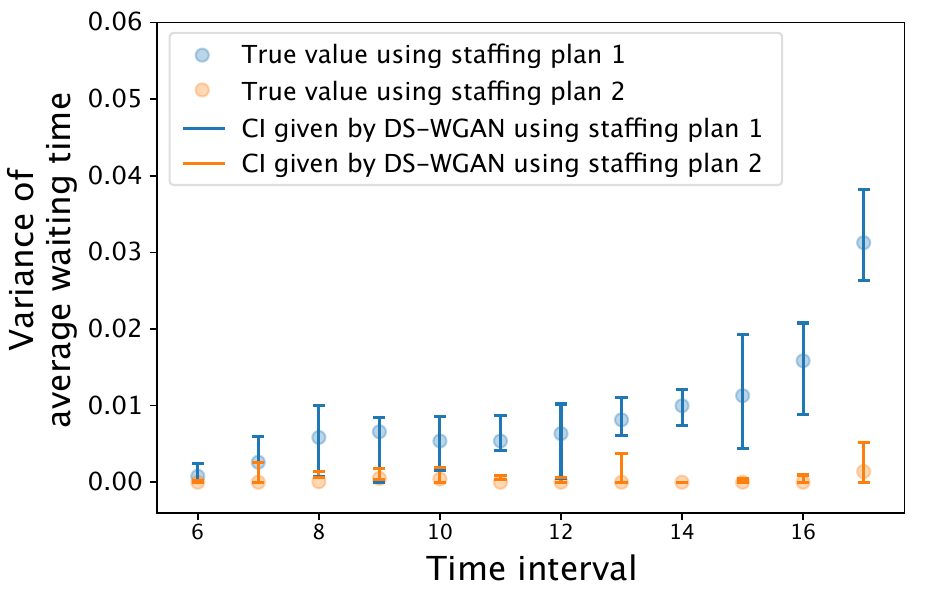}               \label{fig:multi_server_var}
    \end{minipage}}
\end{minipage}

\subfigure[80\% quantile of the average waiting time $\Bar{W}_i$.]
{\begin{minipage}{0.5\textwidth}
\centering                                           \includegraphics[scale=0.4]{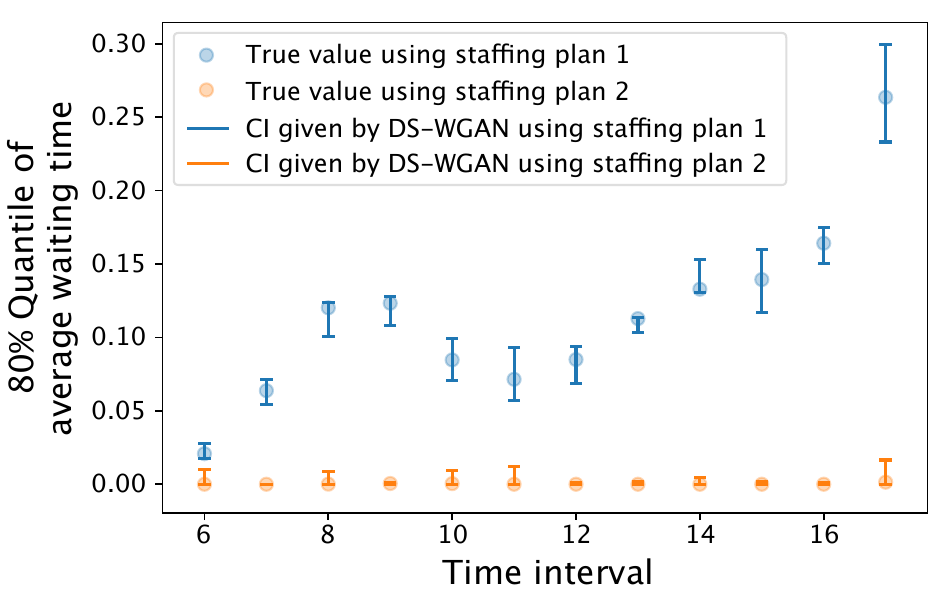}
\label{fig:multi_server_quantile}
\end{minipage}}
\caption{Run-through queue experiment results of Section \ref{sec:many_server_queue}. Mean, variance and quantile of average waiting time for customers arrived in each time interval, fed by the trained DS-WGAN model and the true underlying model. \textmd{The point scatters are the true value given by the underlying model. The vertical lines represents the CI's given by the DS-WGAN and run-through-queue experiments. The blue parts are the results using the first staffing plan, and the orange parts are the results when using the second staffing plan.}}
\label{fig:multi_server}
\end{figure}

We use the trained DS-WGAN model to generate $n=300$ iid copies of arrival process $\tilde{N} = \{\tilde{N}(t):t\in[0,T]\}$. We run each of the generated arrival processes into the many-server queue with the prescribed staffing plan, generating $n=300$ independent realizations of the queueing operations on $[0,T]$. For results demonstration, for each queueing operation realization, we record the average waiting time $\Bar{W}_i$ averaged over the waiting times of all customers that arrive at the system in that realization over the $i$-th time interval. Using the $n=300$ independent realizations, we compute the expectation, variance, and 80\% quantile of $\Bar{W}_i$. We then repeat this procedure 100 times to construct $95\%$ confidence interval (CI) for the expectations, variances, and quantiles, and plot the CI's in Figure \ref{fig:multi_server}. We also plot the benchmark value of the expectation, variance and quantile for comparison, which comes from the synthetic queueing data. The benchmark value is computed by independently simulating 3,000 replications of the queueing operations using the true underlying arrival process model, discretized at the resolution of $\delta = 0.001$. Figure \ref{fig:multi_server} suggests that, for both staffing plans,  the arrival processes generated by the trained DS-WGAN model using $n=300$ samples render close waiting time performances as compared to the  benchmark. 

\subsection{Performance Test on DS-WGAN using DSPP Model by \cite{oreshkin2016rate}}

\label{sec:exper_pierre}
In this subsection, we test the performance of DS-WGAN on a synthetic arrival data set. The underlying arrival process model is a general doubly stochastic Poisson process (DSPP) model introduced by Section 5 in the seminal paper \cite{oreshkin2016rate}. The random intensity model is given by a \textit{PGnorta} model named by \cite{oreshkin2016rate}, which has been shown to have strong flexibility to match the variance and covariance structure presented in several real arrival data sets. We next provide a brief summary of this model, but refer to Section 5 of \cite{oreshkin2016rate} for a detailed description of the model. 

\subsubsection{A Brief Summary: PGnorta Model}
\label{sec:PGnorta_model}

In the doubly stochastic PGnorta model introduced by \cite{oreshkin2016rate}, $\Lambda_j$ is given by $\lambda_j\cdot B_j$, where $\lambda_j$ is a deterministic base rate and $B_j$ is a random busyness factor. The vector $(B_1,B_2,\ldots,B_j)$ has a joint distribution, in which the marginal distribution of $B_j$ follows $\text{Gamma}(\alpha_j,\alpha_j)$, where $\text{Gamma}(\alpha_j,\alpha_j)$ represents a gamma distribution with mean one and variance $1/\alpha_j$. The correlation structure of $(B_1,B_2,\ldots,B_p)$ in the doubly stochastic PGnorta model is given by a normal copula. Specifically, there is an underlying joint multi-normal distributed random vector $\mathbf{Z}=\left(Z_{1}, \ldots, Z_{p}\right)$ with mean zero and covariance matrix $\mathbf{R}^{\mathrm{Z}}$. For $j=1,2,\ldots,p$, $B_j=G_j^{{-1}}(\Phi(Z_j))$, where $G_j^{-1}(\cdot)$ is the inverse cumulative distribution function of $\text{Gamma}(\alpha_j,\alpha_j)$ and $\Phi(\cdot)$ is the standard normal cumulative density function. 

We summarize the parameters to be determined in this model as $\Theta= \left\{\mathbf{R}^\mathrm{Z}, \{\alpha_j\}_{j=1}^p, \{\lambda_j\}_{j=1}^p \right\}$. To generate synthetic data using the doubly stochastic PGnorta Model, we set the model parameters as the estimation result for the real data set in Section 6.2 of \cite{oreshkin2016rate}, with $p=22$ where each time interval represents half hour.

\subsubsection{Evaluation of DS-WGAN}
\label{sec:eval_ds_wgan}
\begin{figure}[ht!]
\centering 
\begin{minipage}{\textwidth}
    \subfigure[Mean of arrival count in each time interval.]
    {\begin{minipage}{0.49\textwidth}
    \centering                                           \includegraphics[scale=0.4]{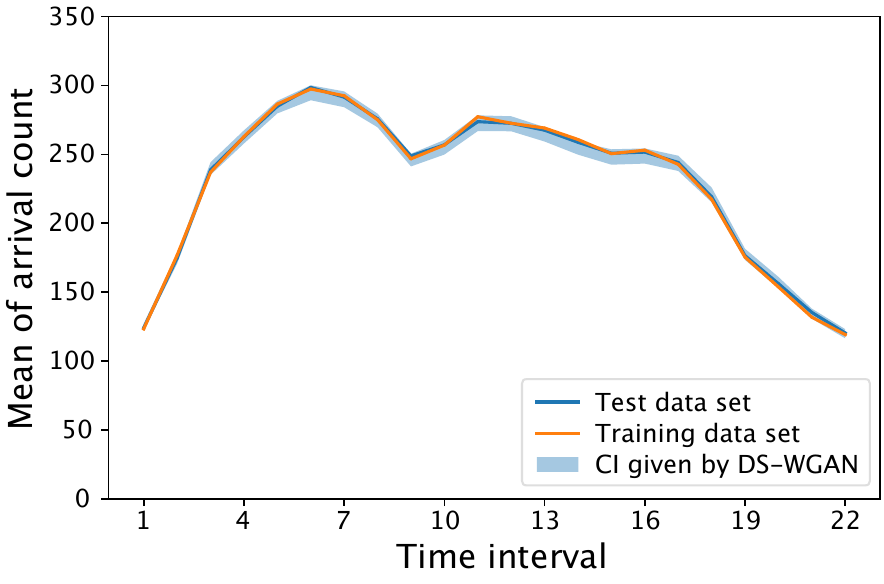}              \label{fig:exper_dswgan_mean}
    \end{minipage}}
    \subfigure[Variance of arrival count in each time interval.]
    {\begin{minipage}{0.49\textwidth}
    \centering                                           \includegraphics[scale=0.4]{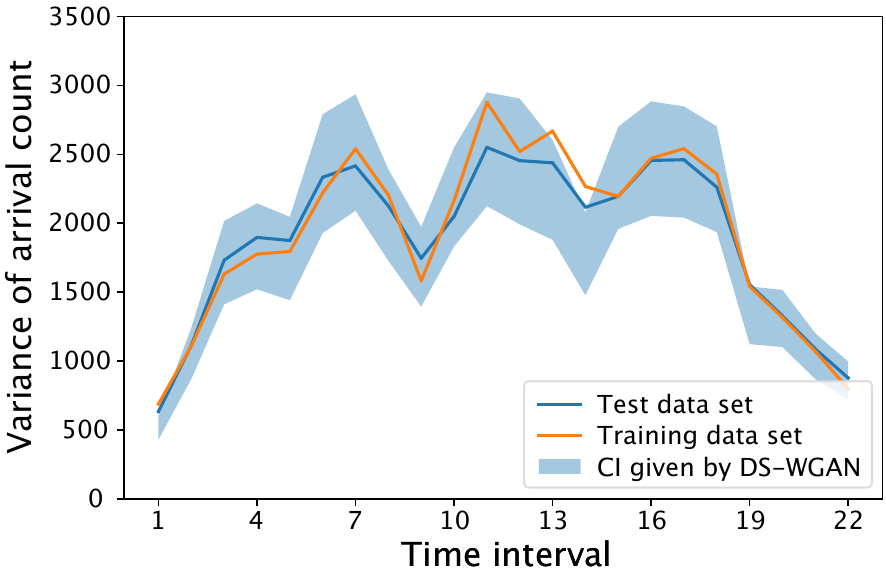}               \label{fig:exper_dswgan_var}
    \end{minipage}}
\end{minipage}

\subfigure[The correlation of past and future arrival counts.]
{\begin{minipage}{0.5\textwidth}
\centering                                           \includegraphics[scale=0.4]{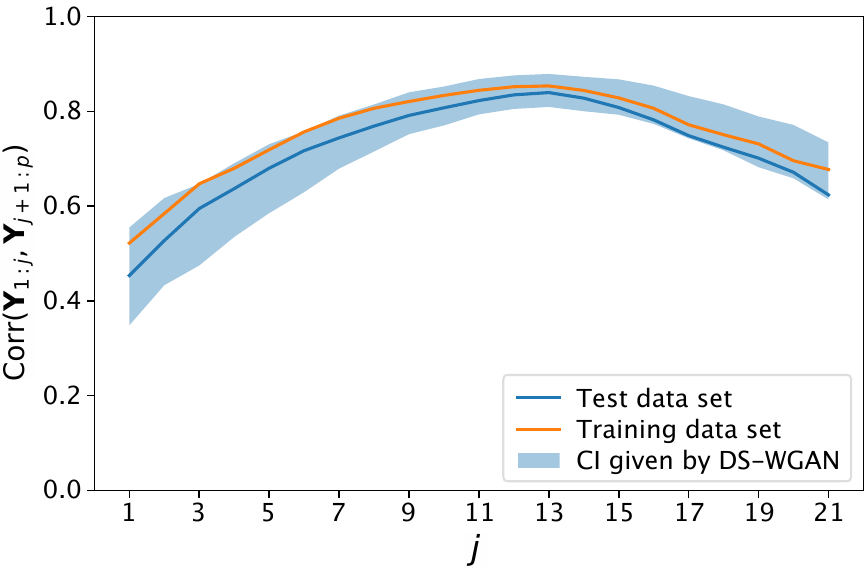}
\label{fig:exper_dswgan_corr}
\end{minipage}}
\caption{Experiment results of Section \ref{sec:exper_pierre}: Performance of DS-WGAN on synthetic data generated by \cite{oreshkin2016rate}. \textmd{The solid blue polylines are the marginal mean, marginal variance and correlation of past and future arrival counts computed on test data set. The solid oranges polylines are for the training data set, and the light blue areas are the confidence intervals given by DS-WGAN model.}}
\label{fig:exper_dswgan}
\end{figure}

We simulate $n=300$ iid copies of arrival count vectors using the calibrated doubly stochastic PGnorta model. These 300 iid copies of synthetic data are used as the training set. We describe and plot some summary statistics of this synthetic data set before discussing the training of DS-WGAN. We compute the marginal mean, marginal variance of arrival count for each time interval, as shown by the solid blue polylines in Figure \ref{fig:exper_dswgan_mean} and \ref{fig:exper_dswgan_var}. To illustrate the correlation structure of the arrival count vector, we adopt the concept of \textbf{correlation of past and future arrival count} used by \cite{oreshkin2016rate}. To prepare the notation for this concept, for an arrival count vector $\mathbf{X}=(X_1,X_2,\ldots,X_p)$, we define $Y_{j:j+d-1}=X_{j}+\cdots+X_{j+d-1}$ for $j \geqslant 1$ and $d \leqslant p-j+1$. This is the total count of arrivals in $d$ successive time intervals starting from $j$-th time interval. The correlation of past and future arrival count at the end of $j$-th time interval is defined as the correlation between the total count of arrivals in the first $j$ periods ($Y_{1:j}$) and the total count of arrivals in the remaining $p-j$ periods ($Y_{j+1:p}$), and is denoted as $\text{Corr}(\mathbf{Y}_{1:j},\mathbf{Y}_{j+1:p})$. We compute $\text{Corr}(\mathbf{Y}_{1:j},\mathbf{Y}_{j+1:p})$ for the training data set as a function of $j$. The result is plotted by blue solid line in Figure \ref{fig:exper_dswgan_corr}.

With the training set, we train the DS-WGAN model with the same neural network parametrization and training configurations as in Section \ref{sec:exper_dspp_cir}. The trained generator of DS-WGAN model is denoted as $g_\text{n}^*$. The arrival count vectors generated are thus $\mathbf{h}(g_\text{n}^*(\mathbf{Y}))$. Using the trained generator, we simulate $300$ iid replications of $\mathbf{h}(g_\text{n}^*(\mathbf{Y}))$, and then compute the marginal mean, marginal variance of arrival count for each time interval, as well as the correlation of past and future arrival count. We repeat this generation process 100 times to compute a 95\% confidence interval for each of the computed statistics. The confidence intervals are shown as light blue areas in Figure \ref{fig:exper_dswgan}.

Independently from the training set, we generate 20,000 more iid copies of arrival count vectors from the true underlying doubly stochastic PGnorta model. We use them as the test data set. For the test data set, the marginal mean, marginal variance of arrival count for each time interval, as well as the correlation of past and future arrival count are computed and plotted as orange solid polylines in Figure \ref{fig:exper_dswgan}.

The DS-WGAN generated confidence intervals for the three sets of statistics (marginal mean, marginal variance, and correlation of past and future arrival count) cover that of the test and the training data sets on almost every time interval. This experiment suggests that the DS-WGAN framework can perform well in terms of capturing these three sets of statistics, when the data sample size is only $n=300$. This data sample size requirement represents roughly a year length of data in many service systems operation, which is a moderate and reasonable requirement in the contexts of operations research and management science. 

\subsection{Testing DS-WGAN on a Call Center Dataset}
\label{sec:exper_callcenter_real}
In this subsection, we use a real arrival data set from an Oakland call center to test the performance of DS-WGAN. The data comes from an Oakland Public Works Call Center, in the city of Oakland in California, USA. We consider the call arrivals from 8AM to 4PM each day. Specifically, we have arrival counts data over each half hour ($p=16$ time intervals per day) from January 1, 2013 to December 31, 2017. Preliminary data analysis suggests that there is no significant day-of-week effect for weekdays, so we retain all the data from Monday to Friday. There's a total of 1,304 days retained. 

Outliers are removed according to 2.5\% and 97.5\% percentiles for the data set. Specifically, the 2.5\% and 97.5\% percentiles for the arrival count in each time interval are computed. Any arrival count vector with at least one dimension smaller than the 2.5\% percentile or larger than 97.5\% percentile in that dimension is regarded as outlier. Based on this criterion, a total percentage of 12\% of data are marked as outliers and removed. The training data set is used as input data to the DS-WGAN framework. For the training of DS-WGAN, the neural network parametrization and training configurations are set the same as the setting in section \ref{sec:exper_dspp_cir}, except that the input data vector dimension $p$ now is 12 and the gradient penalty coefficient is set to 5.

\begin{figure}[t]
    \centering

    \subfigure[Marginal mean]{
        \label{fig:callcenter.mean2.5}
        \includegraphics[width=0.33\textwidth]{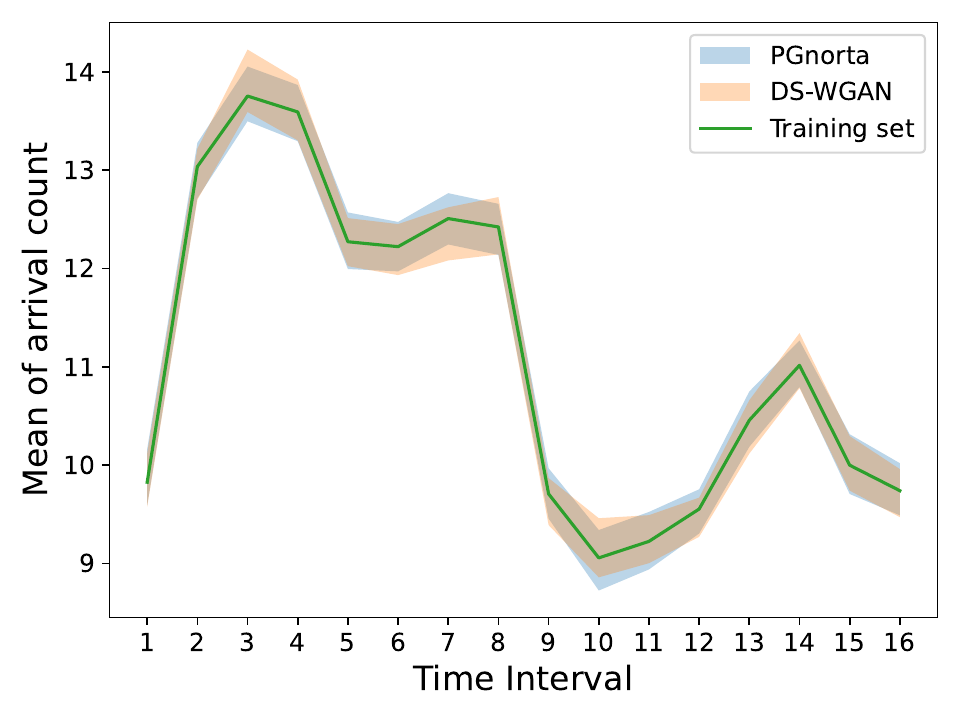}}
    \subfigure[Marginal variance]{
        \label{fig:callcenter.var2.5}\includegraphics[width=0.33\textwidth]{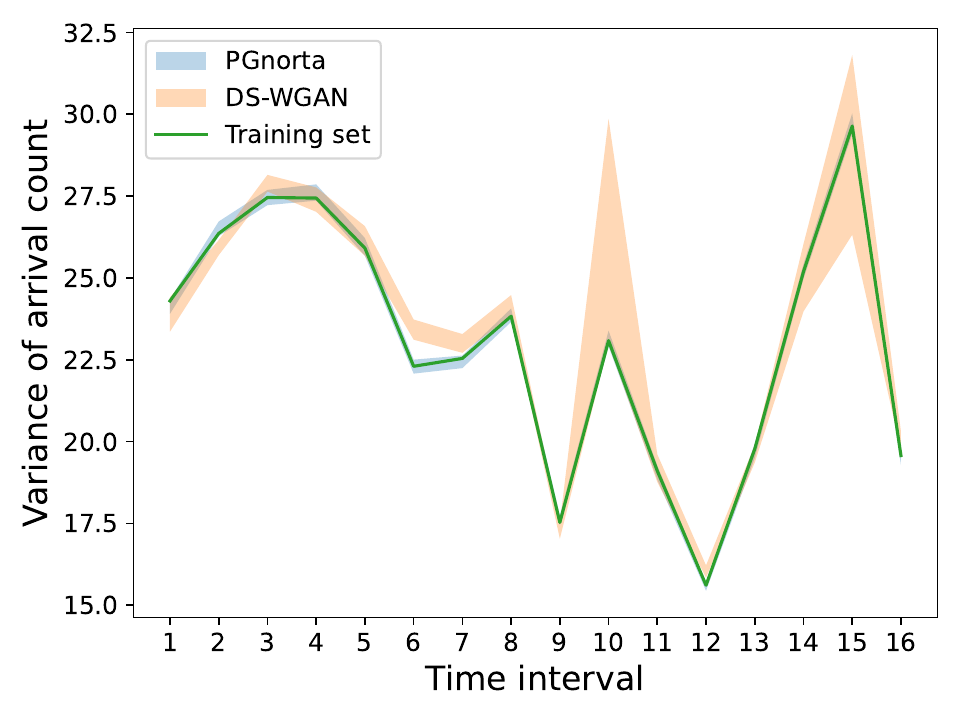}}

    \subfigure[Past and future correlation]{
        \label{fig:callcenter.corr2.5}
        \includegraphics[width=0.33\textwidth]{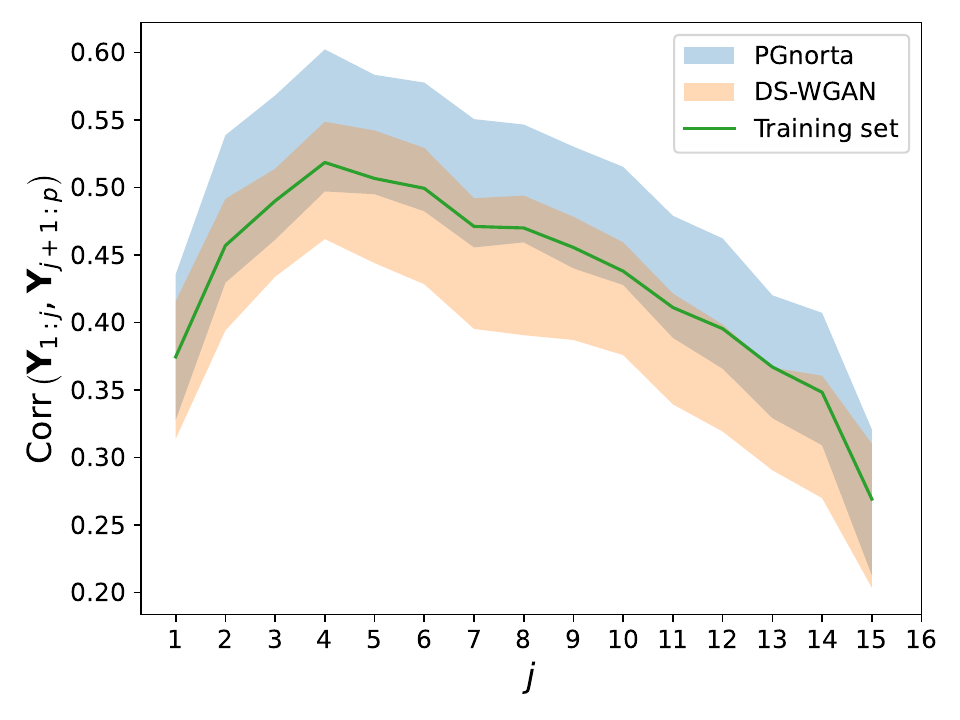}}
    \subfigure[Marginal Wasserstein distance]{
        \label{fig:callcenter.w_dist2.5}
        \includegraphics[width=0.33\textwidth]{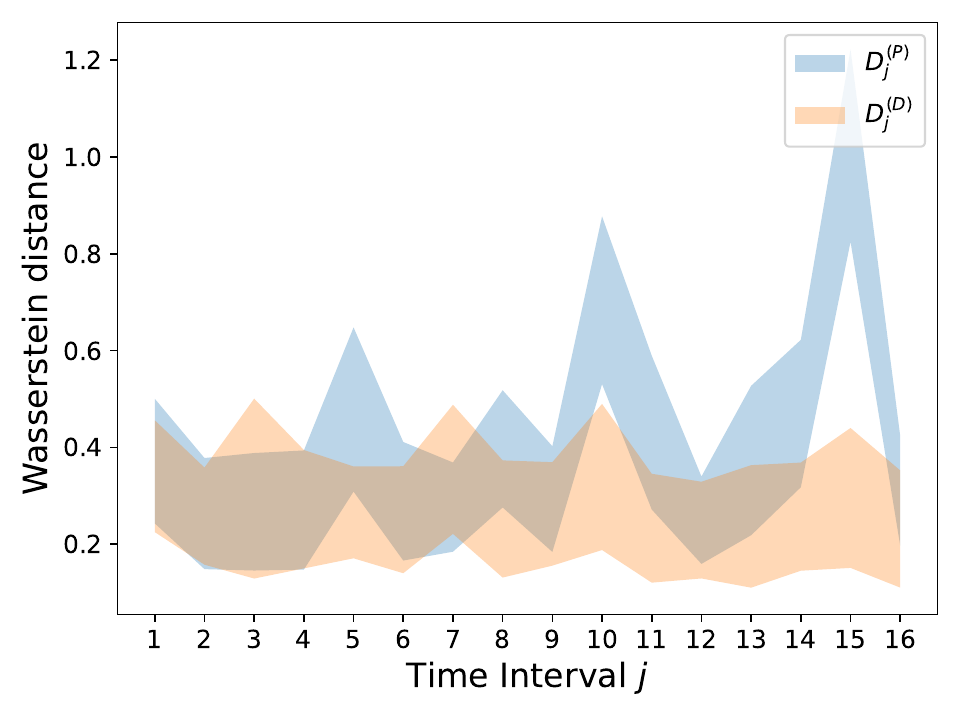}}

    \caption{Experiment results of Section \ref{sec:exper_callcenter_real}: Performance of PGnorta and DS-WGAN on real-world dataset from a call center.}
    \label{fig:callcenter_result}
\end{figure}

\subsubsection{Estimation and Evaluation}

\textcolor{black}{We train DS-WGAN on the call center dataset, and follow the procedure described in \cite{oreshkin2016rate} to estimate the PGnorta model. Using the trained DS-WGAN and PGnorta model, we collect $M=700$
\textit{i.i.d.} samples. Aside from the evaluation metrics used in section
\ref{sec:exper_pierre}, we further consider Wasserstein distance between the marginal distribution of the training set and samples collected from each model as an evaluation metric. The Wasserstein distance between one-dimensional probability distributions has a closed-form and can be effectively computed (see \cite{vallender1974calculation}). Denote the samples collected from PGnorta and DS-WGAN as $\{\mathbf{X}^{(PG)}_{i}\}_{i=1}^M$ and $\{\mathbf{X}^{(DS)}_{i}\}_{i=1}^M$ respectively. We compute
\begin{equation}
    \label{eq:w_dist_PGnorta}
    D^{(PG)}_{j} \stackrel{\text { def }}{=} W(\{X_{i,j}\}_{i=1}^N,\{X_{i,j}^{(PG)}\}_{i=1}^M)
\end{equation}
as the Wasserstein distance between $j$-th interval arrival count from the training set and samples collected from PGnorta model, and
\begin{equation}
    \label{eq:w_dist_DSWGAN}
    D^{(DS)}_j\stackrel{\text { def }}{=}W(\{X_{i,j}\}_{i=1}^N,\{X_{i,j}^{(DS)}\}_{i=1}^M)
\end{equation}
as the Wasserstein distance between $j$-th interval arrival count from the training set and samples collected from DS-WGAN model, for $j=1,2,\ldots,p$. A smaller Wasserstein distance indicates the model can fit the training set better, from the perspective of the marginal distribution. The sampling from DS-WGAN and PGnorta is repeated 100 times to compute a 95\% CI.} The results are given in Figure \ref{fig:callcenter_result}. {In this experiment, DS-WGAN and PGnorta have comparable
performances on the first three metrics. For the Wasserstein distance, we observe that PGnorta performs much better than DS-WGAN, which is an interesting phenomenon since DS-WGAN aims at using the Wasserstein distance itself as the optimization objective. Aside from that PGnorta is likely a very good fit for this dataset, it is possible that the discriminator, aiming at approximating the Wasserstein distance of the joint distributions, does not capture  the marginal distributions as well as PGnorta in this dataset, because of the limitation in its discriminative power.}

\subsection{Testing DS-WGAN on a Bike Share Dataset}
\label{sec:exper_bikeshare_real}

\begin{figure}[t]
    \centering

    \subfigure[Marginal mean]{
        \label{fig:bikeshare.mean}
        \includegraphics[width=0.33\textwidth]{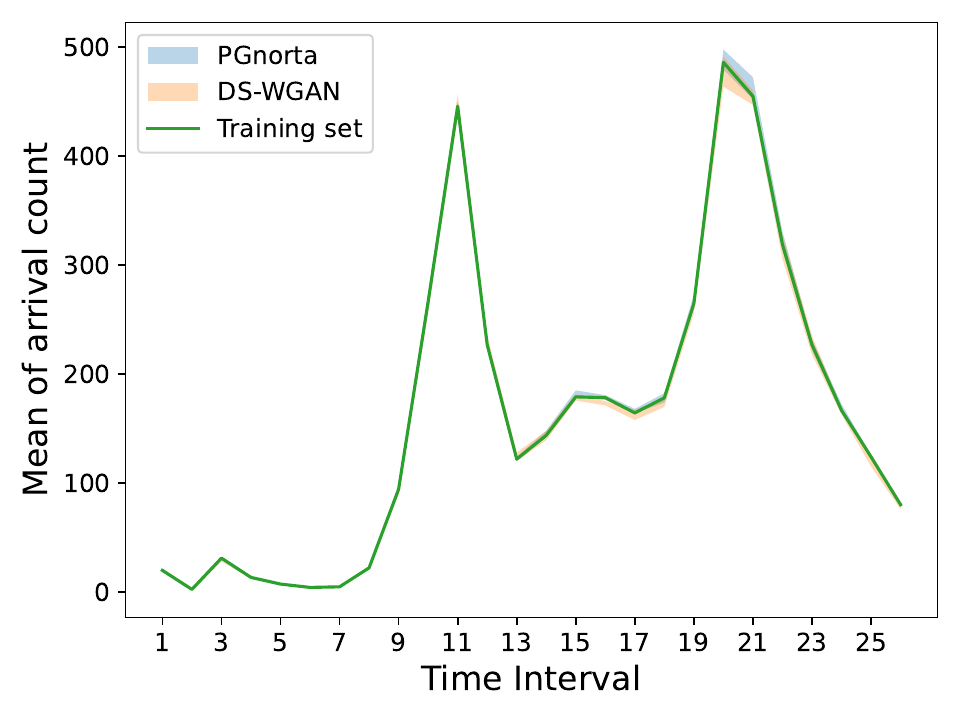}}
    \subfigure[Marginal variance]{
        \label{fig:bikeshare.var}\includegraphics[width=0.33\textwidth]{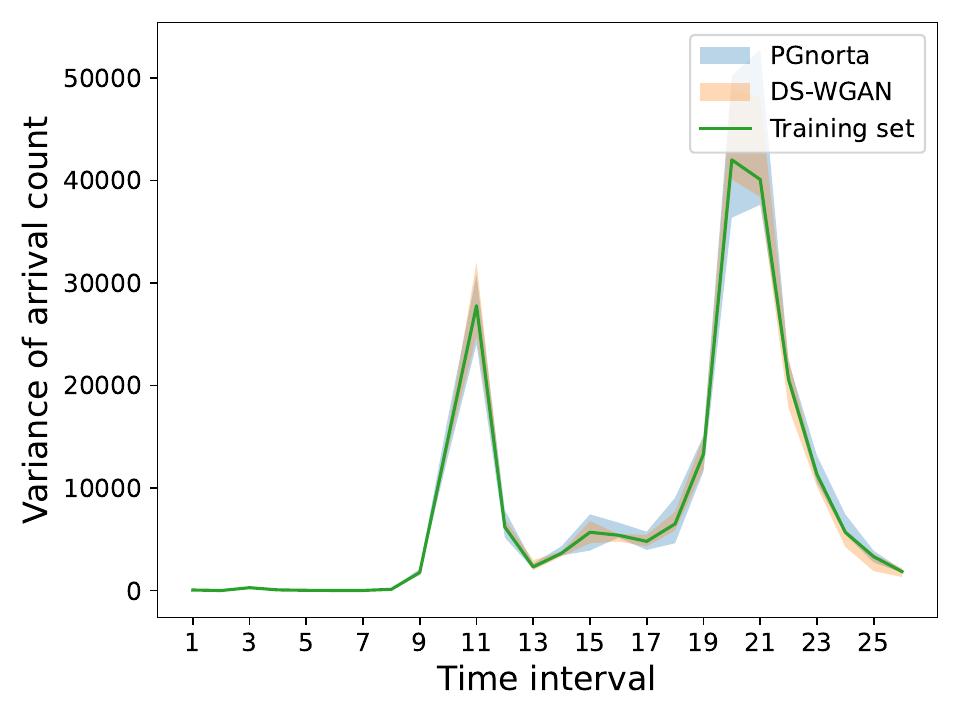}}

    \subfigure[Past and future correlation]{
        \label{fig:bikeshare.corr}
        \includegraphics[width=0.33\textwidth]{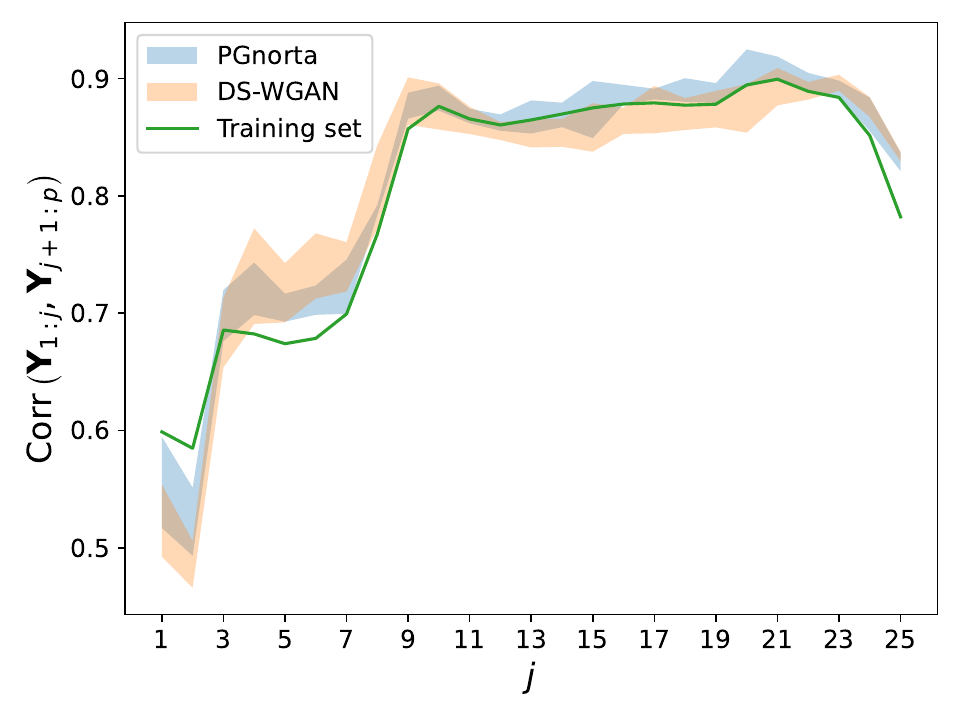}}
    \subfigure[Marginal Wasserstein distance]{
        \label{fig:bikeshare.w_dist}
        \includegraphics[width=0.33\textwidth]{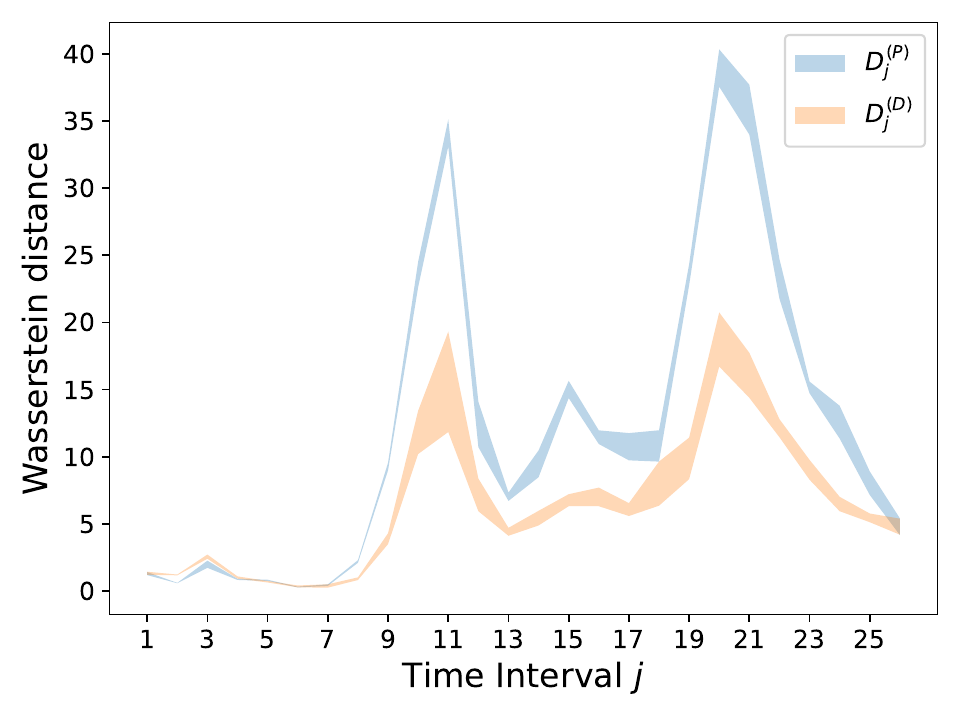}}

    \caption{Experiment results of Section \ref{sec:exper_bikeshare_real}: Performance of PGnorta and DS-WGAN on real-world dataset of bike-sharing.}
    \label{fig:real_bikeshare}
\end{figure}  

\begin{figure}[t]
    \centering
 \subfigure[8-th dimension]{
 \includegraphics[width=0.33\textwidth]{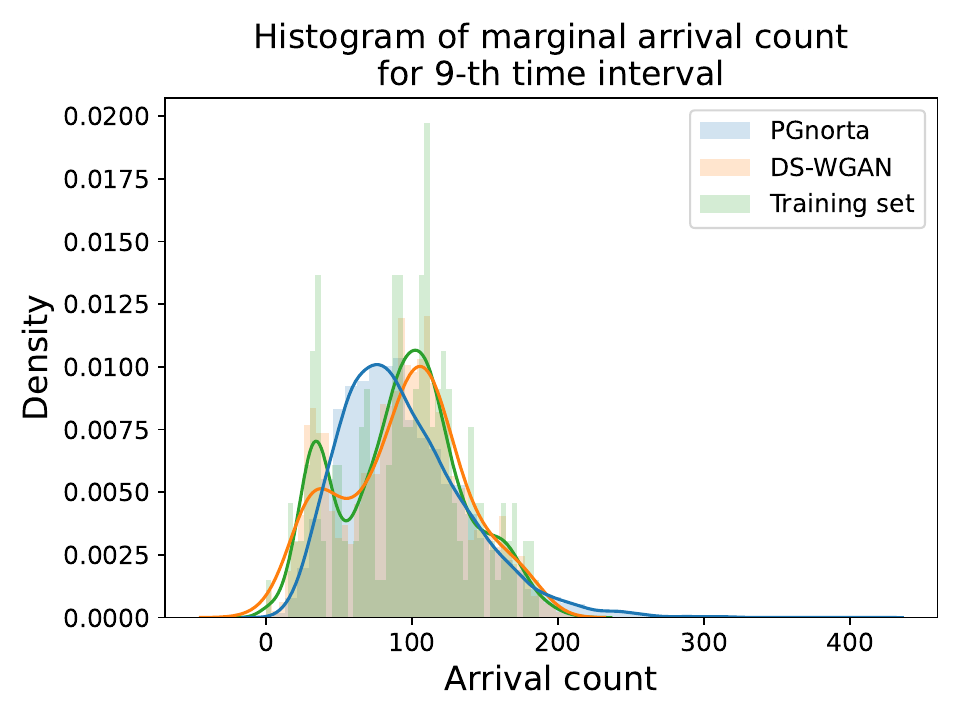}}
 \subfigure[10-th dimension]{
 \includegraphics[width=0.33\textwidth]{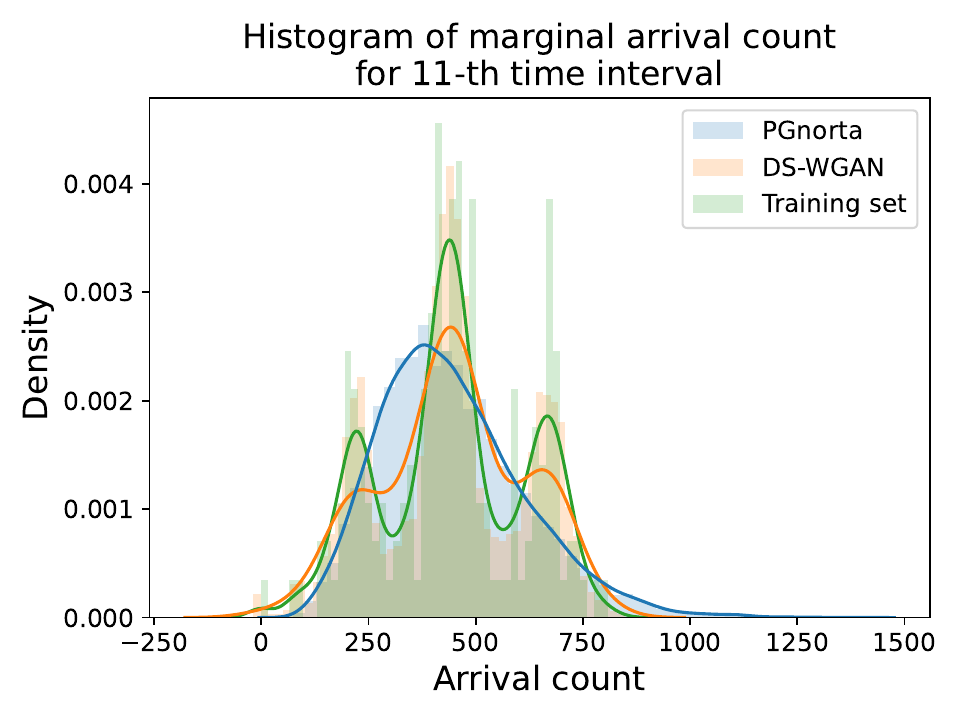}}
 \subfigure[12-th dimension]{
 \includegraphics[width=0.33\textwidth]{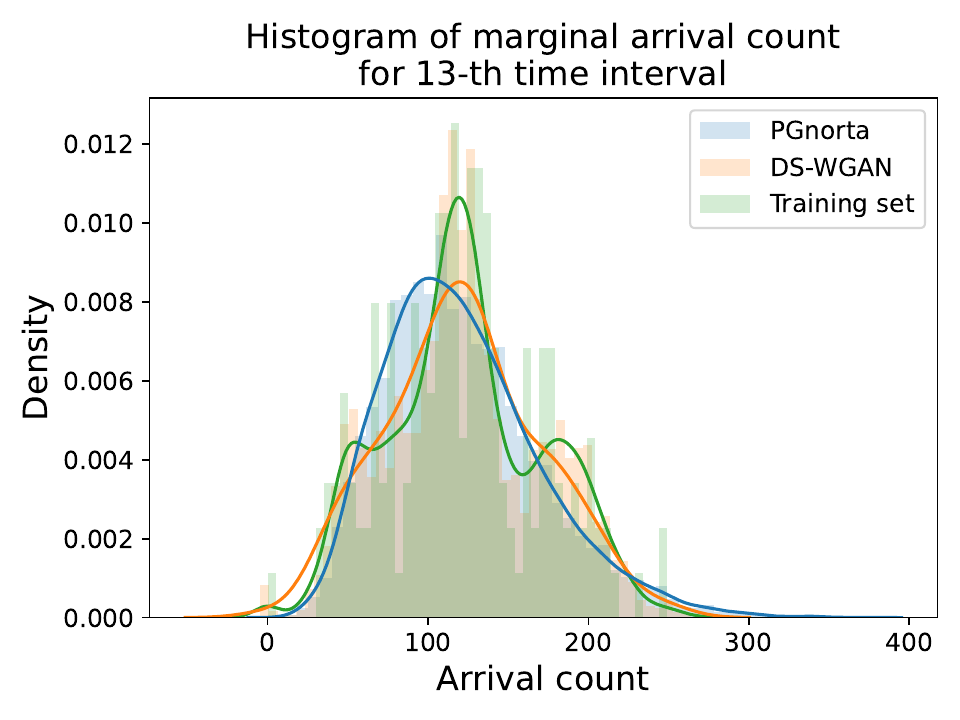}}
 \subfigure[21-th dimension]{
 \includegraphics[width=0.33\textwidth]{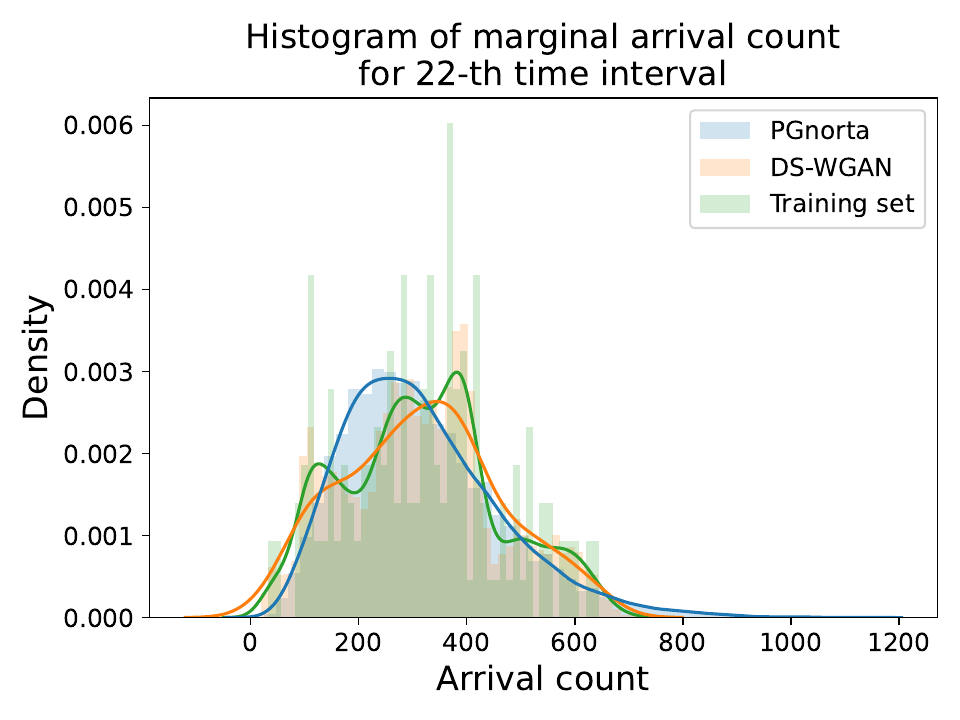}}
 
 \caption{Experiment result of Section \ref{sec:exper_bikeshare_real}: Marginal histograms for bike-sharing dataset}
 \label{fig:bike}
\end{figure}
In this subsection, we use a real arrival data set from a bike-sharing system to illustrate the performance of DS-WGAN.

This arrival data set comes from a bike-sharing system named Capital Bikeshare located in Washington, D.C. The dataset is provided by Kaggle [https://www.kaggle.com/c/bike-sharing-demand/data] (\cite{fanaee2014event}). This system operates 24 hours a day and customers can arrive to rent a bike at any time. The date range of the dataset is from January 20, 2011 to December 31, 2012. For the $i$-th day in this range, we have a 24-dimensional arrival count vector $\mathbf{x}^{(i)}$, whose $j$-th element represents the number of users who arrive and start to rent a bike in the $j$-th hour of this day. In this analysis, we focus on the hourly arrivals over the entire system and do not differentiate which specific location the customer arrivals at.

The data for weekends and holidays are removed. There is no significant day-of-week effect for weekdays suggested by preliminary data analysis, so all data from weekdays are retained. We note that out of all weekdays only about 65\% of days are recorded and available in the original data set. A total of 311 days are retained. We randomly select 2/3 of these days (208 days) as the training data set, and the remaining 103 days as the test data set. The training data set is used as input data to the DS-WGAN framework. The gradient penalty coefficient is set to 10. The parametrization of the generator neural network is given by $L=4$, $\tilde{n}=\left(n_{1}, n_{2},n_{3},n_4\right)=(512,512,512,24)$. The parametrization of the discriminator neural network is given by $L=4$, and $\tilde{n}=\left(n_{1}, n_{2},n_{3},n_4\right)=(512,512,512,24)$. The other training configurations are set the same as the setting in section \ref{sec:exper_dspp_cir}.

With the trained DS-WGAN and estimated PGnorta model, similar to Section \ref{sec:exper_callcenter_real}, we compute the marginal mean, marginal variance of arrival count for each time interval, the correlation of past and future arrival count, as well as the marginal Wasserstein distance. The computation is done for both the training data set and the test data set. The results are given in Figure \ref{fig:real_bikeshare}. {In this experiment, DS-WGAN and PGnorta have comparable
performances on the first three metrics, and DS-WGAN performs much better than PGnorta on the marginal Wasserstein metirc.} To find out a possible reason, we further look into a fifth metric, which is the marginal distribution (histogram) of the arrival counts in each dimension. In Figure \ref{fig:bike}, using the real
bikeshare dataset, we noticed that the marginal distribution of the arrival counts
in several dimension does show a multi-modal shape. In those cases, DS-WGAN may perform moderately better than PGnorta.

\section{Conclusion}\label{sec:conclude}
In this work, we propose a doubly stochastic simulator that is composed of a stochastic generative neural network and a classical Monte Carlo Poisson simulator. This particular doubly stochastic structure is designed to integrate the domain knowledge in the stochastic structure and the approximation power of neural networks. In addition, we provide statistical analysis and computational analysis for the estimation procedure of the proposed simulator. Numerical experiments are conducted to illustrate the feasibility and performance of the proposed framework. One potential future direction is as follows. In the current work, Wasserstein distance is used as a metric to match the general distribution of the simulation model with data. For certain applications where the simulation model is used as inputs, there may be specific parts of output distribution information, e.g., quantiles, that of interest. It may be useful to adjust the metric from the Wasserstein distance to a more focused metric that is tailored to different downstream tasks.

\ECSwitch


\ECHead{Appendix}

\section{Main Proof of Theorem \ref{thm:dswgan}} \label{ec:thm-dswgan}

The proof for the statistical theory of DS-WGAN utilizes the function approximation theory to analyze the statistical properties and convergence rate for neural-network based statistical estimator. \cite{bai2018approximability} and \cite{chen2020statistical} are two of the representative works in this area, and the statistical theory proved in our work is inspired by them. In the DS-WGAN framework, the generator is composed of a generative neural network and a Poisson counts simulator.

\subsection{Decomposition of Error}
First note that $g_n^*$ is the optimizer of the optimization problem (\ref{eq:WGAN_opt_proof}).We need to provide an upper bound for $ W(\mu_n^*,\mu)$.  
For two arbitrary $p$-dimensional probability measures $\mu_1$ and $\mu_2$, the Wasserstein distance $W(\mu_1,\mu_2)$ can be written as
$$
\begin{aligned}
    W\left(\mu_1,\mu_2\right)= \sup _{\|f\|_{L} \leq 1} \mathbb{E}_{\mathbf{X}_1 \sim \mu_1}[f(\mathbf{X}_1)]-\mathbb{E}_{\mathbf{X}_1 \sim \mu_2}[f(\mathbf{X}_2)],
\end{aligned}
$$
in which $\|f\|_{L} \leq 1$ represents that $f$ is a $1$-Lipschitz function. 
To provide the upper bound, we decompose $ W(\mu_n^*,\mu)$ into several error terms and upper bound them separately. We first have
\begin{equation}
\label{eq:decomp_statErr}
     W(\mu_n^*,\mu)\leq  W(\mu_n^*,\hat{\mu}_n)+\underbrace{W(\hat{\mu}_n,\mu)}_{\text{statistical error}}.
\end{equation}
To control for the first term on the right hand side,  $W(\mu_n^*,\hat{\mu}_n)$, we first discuss bounding an alternative expression, $d_{\mathcal{F}}(\mu_n^*,\hat{\mu}_n)$. Then it remains only to establish the relationship between  $W(\mu_n^*,\hat{\mu}_n)$ and $d_{\mathcal{F}}(\mu_n^*,\hat{\mu}_n)$. Observe that
\begin{align}
    &d_{\mathcal{F}}(\mu_n^*,\hat{\mu}_n)\leq d_{\mathcal{F}}(\mu_g,\hat{\mu}_n),\label{eq:decomp_optimal}\\
    &d_{\mathcal{F}}(\mu_g,\hat{\mu}_n)=W(\mu_g,\hat{\mu}_n)+\underbrace{d_{\mathcal{F}}(\mu_g,\hat{\mu}_n)-W(\mu_g,\hat{\mu}_n)}_{\text{discriminator approximation error }}\label{eq:decomp_d_approx},\\
    &W(\mu_g,\hat{\mu}_n)\leq \underbrace{W(\mu_g,\mu)}_{\text{generator approximation error}}+\underbrace{W(\mu,\hat{\mu}_n)}_{\text{statistical error}}.\label{eq:decomp_g_approx}
\end{align}
In the following, we specify the notations in the above inequalities (\ref{eq:decomp_statErr})-(\ref{eq:decomp_g_approx}), explain the reasons of inequality (\ref{eq:decomp_optimal}) (other inequalities either follow from the triangular inequality), and provide interpretations of the error terms. 
\begin{itemize}
    \item $\hat{\mu}_n$ is the empirical distribution of the data set with $n$ samples. The statistical error  $W(\mu,\hat{\mu}_n)$ in both (\ref{eq:decomp_statErr}) and (\ref{eq:decomp_g_approx}) arise from \begin{enumerate}
        \item
        limitation in sample size $n$
        \item weak correlation among the sample set, if we allow for it to exist.
    \end{enumerate}  
    \item  $d_\mathcal{F}$ in (\ref{eq:decomp_optimal}) is the distance defined by the discriminator functional class $\mathcal{F}_{\text{NN}}(\kappa,L,P,K,\varepsilon_f)$, namely,
    \begin{equation*}
    d_{\mathcal{F}}(v_1,v_2)=\sup_{f_\omega\in\mathcal{F}_{\text{NN}}}\mathbb{E}_{\mathbf{X}_1 \sim \nu_1}[f_\omega(\mathbf{X}_1)]-\mathbb{E}_{\mathbf{X}_1 \sim \nu_2}[f_\omega(\mathbf{X}_2)].
    \end{equation*}
    \item $\mu_n^*$ in (\ref{eq:decomp_statErr}) and (\ref{eq:decomp_optimal}) is the distribution of $\mathbf{h}({g}_n^*(\mathbf{Y}))$, and $\mu_g$ in (\ref{eq:decomp_optimal}) is the distribution of $\mathbf{h}({g}(\mathbf{Y}))$, where $g_n^*,g\in\mathcal{G}_{\text{NN}}$ are solutions to different optimization problems. Namely, $g^*_n$ is the optimal solution of the optimization problem 
\begin{equation}
\min_{g_\theta \in \mathcal{G}_{\mathrm{NN}}} \max_{f_\omega \in \mathcal{F}_{\mathrm{NN}}} \E_{\mathbf{Y}\sim \nu} \left[f(\mathbf{h}(g(\mathbf{Y};\theta);\omega ))\right] - \frac{1}{n}\sum_{i=1}^n f(\mathbf{X}_i;\omega),
    \label{eq:WGAN_opt_proof_EC}
\end{equation}
denoted as (\ref{eq:WGAN_opt_proof}) in the main text, and
    \begin{equation}
    g=\inf_{g_\theta\in\mathcal{G}_{\text{NN}}}W(\mathbf{h}({g_\theta}(\mathbf{Y})),\mu).
    \label{eq_EC_g}
    \end{equation}
 We note that equation (\ref{eq:WGAN_opt_proof_EC}) can be equivalently reformulated as 
 \begin{equation*}
\min_{g_\theta \in \mathcal{G}_{\mathrm{NN}}} d_{\mathcal{F}}(\mathbf{h}(g_\theta(\mathbf{Y})),\hat{\mu}_n).
\end{equation*}
Thus, the optimality of $g_n^*$ with regard to problem (\ref{eq:WGAN_opt_proof_EC}) is the reason for inequality (\ref{eq:decomp_optimal}). Note that $g_n^*$ is likely to be different from $g$ for the following reasons:
    \begin{enumerate}
        \item Optimization problem (\ref{eq:WGAN_opt_proof_EC}) aims to minimize the distance between the distribution of $\mathbf{h}(g_n^*(\mathbf{Y}))$ and $\hat{\mu}_n$, and optimization problem (\ref{eq_EC_g}) aims to minimize the distance between the distribution of $\mathbf{h}(g_n^*(\mathbf{Y}))$ and $\mu$.
        \item The difference between $\mathcal{F}_{\text{NN}}(\kappa,L,P,K,\varepsilon_f)$ and the $1$-Lip functional class also induces some difference between the two optimal solutions. This portion of error is absorbed in the discriminator approximation error in equation (\ref{eq:decomp_d_approx}) and the discriminative power error (\ref{eq:dis_error2}) in the following section.
    \end{enumerate}
\end{itemize} 

Before discussing the relationship between  $W(\mu_n^*,\hat{\mu}_n)$ and $d_{\mathcal{F}}(\mu_n^*,\hat{\mu}_n)$, and the methods for bounding the error terms, we introduce a neural network approximation theorem in the next subsection, which serves as the foundation of our proofs ahead.

\subsection{Neural Network Approximation}
\label{sec:NN_approx}
In this section, we introduce the
foundation of proof, which is the deep neural network approximation theory. We first present a theorem adapted from  \cite{elbrachter2019deep} of approximating bounded Lipschitz functions.
\begin{theorem}[Page 2589 of {\cite{elbrachter2019deep}}]	
Given any $\delta\in(0,1)$, there exists such a ReLU network with $\kappa=O(\delta^{-1})$, $L=O(\log\delta^{-1})$, $P=O(\delta^{-p})$ and $K=O(\delta^{-p}\log\delta^{-1})$ such that, for any bounded Lipschitz functions $f$ with domain bounded by $[0,1]^p$, if the weight parameters are properly chosen, the network yields a function $f_\omega$ such that $\Vert f-f_\omega\Vert_\infty\leq \delta$. The notations are defined as follows: $\kappa$ is the weight magnitude, $L$ is the number of layers, $P$ is the maximum width of the layers and $K$ is the number of neurons.
\label{thm-approx}
\end{theorem}

\subsection{Discriminative Power of the Discriminator}
\label{sec:EC_discriminative power}

In this section, we discuss the relationship between $W(\mu_n^*,\hat{\mu}_n)$ and $d_{\mathcal{F}}(\mu_n^*,\hat{\mu}_n)$. 
We prove that the former distance can be upper bounded by the latter plus some statistical error terms. We remark that such a shift from Wasserstein distance metric to the discriminator network metric $d_{\mathcal{F}}(,)$ is necessary, because the optimality of $g_n^*$ is only guaranteed under the metric $d_\mathcal{F}(,)$. Further, proving that convergence of $W(\mu_n^*,\hat{\mu}_n)$ is implied by convergence of $d_{\mathcal{F}}(\mu_n^*,\hat{\mu}_n)$, we are in fact lower bounding the distance $d_\mathcal{F}(,)$, which can be interpreted as providing guarantee for the \textit{discriminative power} of the discriminator.

Note that, according to (\ref{eq:discriminator_NN_definition1}), the first layer of the discriminator class is given as $\tilde{\sigma}(\mathbf{W}_1\mathbf{y}+\mathbf{b}_1)$, where $\tilde{\sigma}(x_i)=x_i/(|x_i|+1)$, which is known as the Softsign activation function. We denote the transformation of the first layer as
\begin{equation*}
    \tau(\mathbf{y}):=\tilde{\sigma}(\mathbf{W}_1\mathbf{y}+\mathbf{b}_1),
\end{equation*}
and if such a transformation is performed on a random variable with distribution $\mu$, we denote the transformed distribution as $\tau[\mu]$. Specifically, note that if $\mathbf{W}_1=\mathbf{I}$ and $\mathbf{b}_1=\mathbf{0}$, we have $\tau=\tilde{\sigma}$. With these notations, we have
\begin{equation}
\begin{aligned}
    d_{\mathcal{F}}(\mu_n^*,\hat{\mu}_n)&\geq \tilde{d}_{\mathcal{F}}(\tilde{\sigma}\left[\mu_n^*\right],\tilde{\sigma}[\hat{\mu}_n])\\
    &=W(\tilde{\sigma}\left[\mu_n^*\right],\tilde{\sigma}[\hat{\mu}_n])-\underbrace{\left(W(\tilde{\sigma}\left[\mu_n^*\right],\tilde{\sigma}[\hat{\mu}_n])-\tilde{d}_{\mathcal{F}}(\tilde{\sigma}\left[\mu_n^*\right],\tilde{\sigma}[\hat{\mu}_n])\right)}_{\text{discriminative power error}},
\label{eq:dis_error2}
\end{aligned}
\end{equation}
where $\tilde{d}_\mathcal{F}(,)$ is the distance defined by the neural network functional class 
\begin{equation}
\label{eq:discriminator_class1}
    \begin{aligned}
    \tilde{\mathcal{F}}_{\text{NN}}(\kappa,L,P,K,\varepsilon_f)&=\{\tilde{f}_\omega:\mathbb{R}^p\to\mathbb{R}\vert f_\omega=\tilde{f}_\omega\circ\tau \text{ in form (\ref{eq:discriminator_NN_definition1}) with $L+1$ layers and max width $P$,}\\
    &\Vert W_i\Vert_{\infty}\leq \kappa,\Vert b_i\Vert_{\infty}\leq\kappa,\text{for }i=1,\ldots,L,\sum_{i=2}^L\Vert W_i\Vert_0+\Vert b_i\Vert_0\leq K,\\
    &\vert f_\omega(x)-f_\omega(y)\vert\leq \Vert x-y\Vert+2\varepsilon_f,\forall x,y\in\mathbb{R}^p\},
\end{aligned}
\end{equation}
in other words, $\tilde{\mathcal{F}}_{\text{NN}}$ is $\mathcal{F}_{\text{NN}}$ with the first layer $\tau$ removed. The first inequality of (\ref{eq:dis_error2}) arises from that $\tilde{d}_F$ is computed by taking supreme over a subset of $\mathcal{F}_{\text{NN}}$, namely, the functions in $\mathcal{F}_{\text{NN}}$ with the first layer restricted to be $\tau=\tilde{\sigma}$, namely $\mathbf{W}_1=I$ and $\mathbf{b}_1=0$. With guarantee for the approximation power of $\tilde{\mathcal{F}}_{\text{NN}}(\kappa,L,P,K,\varepsilon_f)$ provided in section \ref{sec:NN_approx},  the discriminative power error in (\ref{eq:dis_error2}) can be controlled as follows:
\begin{equation}
\label{eq:dis_2Eps}
\begin{aligned}
&W(\tilde{\sigma}[\mu_n^*],\tilde{\sigma}[\hat{\mu}_n]) - \tilde{d}_{\mathcal{F}}(\tilde{\sigma}[\mu_n^*],\tilde{\sigma}[\hat{\mu}_n])\\
=&\sup_{\Vert f\Vert_L\leq 1}\left[\mathbb{E}_{\mathbf{X}\sim \mu_n^*}f(\tilde{\sigma}(\mathbf{X}))-\mathbb{E}_{\mathbf{X}\sim\hat{\mu}_n}f(\tilde{\sigma}(\mathbf{X}))\right]
-\sup_{f_\omega\in\mathcal{F}_{\text{NN}}}\left[\mathbb{E}_{\mathbf{X}\sim \mu_n^*}f_\omega(\tilde{\sigma}(\mathbf{X}))-\mathbb{E}_{\mathbf{X}\sim\hat{\mu}_n}f_\omega(\tilde{\sigma}(\mathbf{X}))\right]\\
=&\inf_{f_\omega\in\mathcal{F}_{\text{NN}}}\sup_{\Vert f\Vert_L\leq 1}\mathbb{E}_{\mathbf{X}\sim\mu_n^*}\left[f(\tilde{\sigma}(\mathbf{X}))-f_\omega(\tilde{\sigma}(\mathbf{X}))\right]+\inf_{f_\omega\in\mathcal{F}_{\text{NN}}}\sup_{\Vert f\Vert_L\leq 1}\mathbb{E}_{\mathbf{X}\sim\hat{\mu}_n}\left[f_\omega(\tilde{\sigma}(\mathbf{X}))-f(\tilde{\sigma}(\mathbf{X}))\right]\\
\leq &\inf_{f_\omega\in\mathcal{F}_{\text{NN}}}\sup_{\Vert f\Vert_L\leq 1}2\Vert f-f_\omega\Vert_\infty\leq 2\delta.
\end{aligned}
\end{equation}

We next observe the first term $W(\tilde{\sigma}\left[\mu_n^*\right],\tilde{\sigma}[\hat{\mu}_n])$ on the right hand side of (\ref{eq:dis_error2}). We hope to bound $W(\mu_n^*,\hat{\mu}_n)$ with $W(\tilde{\sigma}\left[\mu_n^*\right],\tilde{\sigma}[\hat{\mu}_n])$. For $W(\mu_n^*,\hat{\mu}_n)$, note that for any joint distribution $\Gamma(\mu_n^*,\mu)$ and any constant $A>0$, we have
\begin{equation}
\label{eq:dis_decomp2parts}
\begin{aligned}
    \mathbb{E}_{(\mathbf{X},\mathbf{X}^\prime)\sim\Gamma(\mu_n^*,\mu)}\Vert\mathbf{X}-\mathbf{X}^\prime\Vert&\leq \sum_{i=1}^p\mathbb{E}_{(\mathbf{X},\mathbf{X}^\prime)\sim\Gamma(\mu_n^*,\mu)}\vert X_i-{X}^\prime_i\vert\\
    &\leq\sum_{i=1}^p\mathbb{E}_{(\mathbf{X},\mathbf{X}^\prime)\sim\Gamma(\mu_n^*,\mu)}\left[\vert X_i-X_i^\prime\vert\Big|X_i\leq A,X^\prime_i\leq A\right]\\
    &+\sum_{i=1}^p\left(\mathbb{E}_{\mathbf{X}\sim\mu_n^*}\left[ X_i|X_i> A\right]P(X_i>A)+\mathbb{E}_{\mathbf{X}^\prime\sim\mu}\left[X_i^\prime|X_i^\prime> A\right]P(X_i^\prime>A)\right)\\
    &:= E_1+E_2,
\end{aligned}
\end{equation}
where $E_1, E_2$ refer to the two summation terms respectively. We first have
\begin{align}
    E_1&\leq (1+A)^2\sum_{i=1}^p\mathbb{E}_{(\mathbf{X},\mathbf{X}^\prime)\sim\Gamma(\mu_n^*,\mu)}\vert \tilde{\sigma}(X_i)-\tilde{\sigma}(X_i^\prime)\vert\label{eq:dis_1+A}\\
    &\leq \sqrt{p}(1+A)^2\mathbb{E}_{(\mathbf{X},\mathbf{X}^\prime)\sim\Gamma(\mu_n^*,\mu)}\Vert \tilde{\sigma}(\mathbf{X})-\tilde{\sigma}(\mathbf{X}^\prime)\Vert.\label{eq:dis_cheb}
\end{align}
The reason for (\ref{eq:dis_1+A}) is that, when $0\leq x,y\leq A$ and $x\not=y$, $\vert\tilde{\sigma}(x)-\tilde{\sigma}(y)\vert/\vert x-y\vert$ is bounded above $(1+A)^{-2}$, and (\ref{eq:dis_cheb}) arises from Chebyshev inequality. To control $E_2$, note that $X_i$ and $X_i^\prime$ are Poisson random variables with intensities upper bounded by constant $2C_\mathbf{\Lambda}$, as specified in assumption \ref{assum:density}. Therefore, $E_2$ can be bounded with incomplete moments of Poisson distribution with intensity $2C_\mathbf{\Lambda}$. According to \cite{haight1967handbook}, incomplete Poisson moments can be calculated as
\begin{equation*}
    \sum_{i=x}^{\infty}ip_i(\lambda)=\lambda Q_{x-2}(\lambda),
\end{equation*}
where
\begin{equation*}
    Q_x(\lambda):=\sum_{i=x+1}^\infty p_i(\lambda)=\frac{1}{\Gamma(x+1)}\int_{0}^\lambda e^{-t}t^xdt.
\end{equation*}
Therefore, $E_2$ can be bounded by
\begin{equation}
    E_2\leq 2\cdot p\cdot 2C_\mathbf{\Lambda}\frac{1}{\Gamma(A-1)}\int_0^{2C_\mathbf{\Lambda}}e^{-t}t^{A-2}dt.
\label{eq:dis_incomplete}
\end{equation}
For some large enough $A$, we use Sterling's equation to bound the right hand side by $(2C_\mathbf{\Lambda}e/A)^{A}$. Summarizing (\ref{eq:dis_decomp2parts})-(\ref{eq:dis_incomplete}), we have
\begin{equation}
    \mathbb{E}_{(\mathbf{X},\mathbf{X}^\prime)\sim\Gamma(\mu_n^*,\mu)}\Vert\mathbf{X}-\mathbf{X}^\prime\Vert\leq \sqrt{p}(1+A)^2\mathbb{E}_{(\mathbf{X},\mathbf{X}^\prime)\sim\Gamma(\mu_n^*,\mu)}\Vert \tilde{\sigma}(\mathbf{X})-\tilde{\sigma}(\mathbf{X}^\prime)\Vert+\left(\frac{2C_\mathbf{\Lambda}e}{A}\right)^A.
    \label{eq_EC14}
\end{equation}
From (\ref{eq_EC14}), take infimum over all joint distributions $\Gamma(\tilde{\sigma}[\mu_n^*],\tilde{\sigma}[\mu])$, 
and then all joint distributions $\Gamma(\mu_n^*,\mu)$, we have
\begin{equation}
W(\mu_n^*,\mu)\leq \sqrt{p}(1+A)^2W(\tilde{\sigma}\left[\mu_n^*\right],\tilde{\sigma}\left[\mu\right])+\left(\frac{2C_\mathbf{\Lambda}e}{A}\right)^A.
\label{eq:dis_transformRelation}
\end{equation}
Also, note that
\begin{equation}
    W(\tilde{\sigma}\left[\mu_n^*\right],\tilde{\sigma}[\hat{\mu}_n])\geq W(\tilde{\sigma}\left[\mu_n^*\right],\tilde{\sigma}[\mu])-W(\tilde{\sigma}[\mu],\tilde{\sigma}[\hat{\mu}_n]).
\label{eq:dis_statErr}
\end{equation}

Finally, we have
\begin{equation}
\label{eq:dis_final}
\begin{aligned}
W(\mu_n^*,\hat{\mu}_n)&\leq W(\mu_n^*,\mu)+W(\mu,\hat{\mu}_n)\\
&\leq \sqrt{p}(1+A)^2W(\tilde{\sigma}\left[\mu_n^*\right],\tilde{\sigma}\left[\mu\right])+\left(\frac{2C_\mathbf{\Lambda}e}{A}\right)^A+W(\mu,\hat{\mu}_n)\\
&\leq 
\sqrt{p}(1+A)^2\left[W(\tilde{\sigma}\left[\mu_n^*\right],\tilde{\sigma}\left[\hat{\mu}_n\right])+W(\tilde{\sigma}\left[\hat{\mu}_n\right],\tilde{\sigma}\left[\mu\right])\right]+W(\mu,\hat{\mu}_n)+\left(\frac{2C_\mathbf{\Lambda}e}{A}\right)^A\\
&\leq \sqrt{p}(1+A)^2\left[\tilde{d}_{\mathcal{F}}(\tilde{\sigma}\left[\mu_n^*\right],\tilde{\sigma}\left[\hat{\mu}_n\right])+ 2\delta\right]+2\sqrt{p}(1+A)^2W(\mu,\hat{\mu}_n)+\left(\frac{2C_\mathbf{\Lambda}e}{A}\right)^A\\
&\leq \sqrt{p}(1+A)^2\left[d_{\mathcal{F}}(\mu_n^*,\hat{\mu}_n)+ 2\delta +2W(\mu,\hat{\mu}_n)\right]+\left(\frac{2C_\mathbf{\Lambda}e}{A}\right)^A
\end{aligned}
\end{equation}
where the last four inequalities are derived from (\ref{eq:dis_transformRelation}), (\ref{eq:dis_statErr}), (\ref{eq:dis_2Eps}) and (\ref{eq:dis_error2}), respectively. Besides, in the fourth inequality, we also used the fact that $W(\tilde{\sigma}\left[\hat{\mu}_n\right],\tilde{\sigma}\left[\mu\right])\leq W(\hat{\mu}_n,\mu)$. The order dependence of $\delta$ and $A$ on sample size $n$ will be derived to balance the error terms in the following subsections.

{
We give an additional remark on the Softsign activation function $\tilde{\sigma}(x)=x/(|x|+1)$. Admittedly, we do not strictly follow the theoretical setting in our numerical experiment. We use $\tilde{\sigma}$ in theoretical proof for the following reasons:
\begin{enumerate}
    \item $\tilde{\sigma}$ transforms the unbounded Poisson distribution to $[0,1]^p$. We use this transformation because existing ReLU-activated neural network approximation theory only guarantees approximability within bounded areas.
    \item $\tilde{\sigma}$ maintains discriminative power in the sense that $|\tilde{\sigma}(x)-\tilde{\sigma}(y)|/|x-y|\geq(1+A)^{-2}$, $\forall x,y\in[0,A]$ and $x\not=y$, which leads to equation (\ref{eq:dis_1+A}). In other words, $x,y$ can be distinguished by $\tilde{\sigma}$ when they are relatively small, and the probability that they get big is upper bounded by incomplete Poisson moment. 
\end{enumerate}
}

\subsection{Generator Approximation Error}
In this section, we derive the upper bound for the generator approximation error $W(\mu_g,\mu)$. Recall that $\mu_g$ is the distribution of $\mathbf{h}(g(\mathbf{Y}))$, where $\mathbf{h}$ is the Poisson simulator, and $\mathbf{Y}$ is the random input of the generator network $g$. Also,  the underlying real distribution $\mu$ can be realized as $\mu\overset{\mathcal{D}}{=}\tilde{\mathbf{h}}(\mathbf{\Lambda})$, where $\tilde{\mathbf{h}}\overset{\mathcal{D}}{=}\mathbf{h}$, and is independent of $\mathbf{\Lambda}$. 

Roughly speaking, we first show that the real distribution of the random intensity $\mathbf{\Lambda}$ can be approximated arbitrarily well by the output of some generator neural network, denoted as $g(\mathbf{Y})$, in other words, $W(g(\mathbf{Y}),\mathbf{\Lambda})\to 0$ as we increase the size of $\mathcal{G}_{\text{NN}}$. We then use a coupling method to upper bound $W(\mu_g,\mu)$ with $W(g(\mathbf{Y}),\mathbf{\Lambda})$. Since the distribution of  $\mathbf{\Lambda}$ has bounded support, the coupling method helps us to avoid discussing neural network approximation on unbounded domains, which so far lacks theoretical guarantee.

According to theorem $1$ of \cite{chen2020statistical}, for uniformly distributed variable $\mathbf{Y}\sim \pi_{\mathbf{Y}}=U[0,1]^p$, and random intensity $\mathbf{\Lambda}\sim \pi_{\mathbf{\Lambda}}$, if $\pi_{\mathbf{\Lambda}}$ satisfies assumption \ref{assum:density}, there exists a transformation $T$ which is differentiable and has uniformly bounded first order partial derivatives, such that $T[\pi_{\mathbf{Y}}]=\pi_\mathbf{\Lambda}$. Moreover, if the network structure yields a transformation $g$ such that $\max_{x\in[0,1]^p}\Vert g(x)-T(x)\Vert_{\infty}\leq \delta$, then by lemma $4$ of \cite{chen2020statistical}, we have $W(g(\mathbf{Y}),\mathbf{\Lambda})\leq \delta$. According to theorem \ref{thm-approx}, this can be achieved using a generator class $\mathcal{G}_{\text{NN}}(\bar{\kappa},\bar{L},\bar{P},\bar{K})$, where $\bar{\kappa}=O(\delta^{-1})$, $\bar{L}=O(\log\delta^{-1})$, $\bar{P}=O(\delta^{-p})$ and $\bar{K}=O(\delta^{-p}\log\delta^{-1})$.

We next use a coupling technique to extend the result to $W(\mu_g,\mu)$. As an overview of the main idea of coupling, note that by definition, $W(X,X_1)=W(X,X_2)$ as long as random variables $X_1$ and $X_2$ have the same distribution function, regardless of the source of their randomness. Therefore, in the following we shift between (compound) random variables with the same distribution and different source of randomness,  and thus different dependency relationships.


Recall that from the definition of the Wasserstein distance, 
\begin{equation*}
    W(g(\mathbf{Y}),\mathbf{\Lambda})=\inf_{\gamma\in\Pi(g(\mathbf{Y}),\mathbf{\Lambda})}\mathbb{E}_{ (X_1,X_2)\sim\gamma}[\Vert X_1-X_2\Vert_2],
\end{equation*}
where $\mathbf{\Lambda}$ is the underlying real random intensity, and $\mathbf{Y}$ is the random input of the generator network $g$. Therefore, $\forall k>0$, we can find a random vector $\mathbf{\Lambda}_k$ that (1) is independent of $\mathbf{h}$ (2) has the same probability distribution function as $\mathbf{\Lambda}$, and (3) is jointly distributed with ${g}(\mathbf{Y})$ such that the Wasserstein distance between ${g}(\mathbf{Y})$ and $\mathbf{\Lambda}_k$ is approximately achieved, meaning
\begin{equation}\label{eq:proof_setlambda}
    W({g}(\mathbf{Y}),\mathbf{\Lambda}_k) \leq \E [\|{g}(\mathbf{Y})- \mathbf{\Lambda}_k\|_2]\leq W({g}(\mathbf{Y}),\mathbf{\Lambda}_k) +\frac{1}{k}.
\end{equation}
We then have the following inequalities,
\begin{equation}
\begin{aligned}
    W(\mu_g,\mu)\, 
    &=W(\mathbf{h}(g(\mathbf{Y})),\mathbf{h}(\mathbf{\Lambda}_k))\\
    &=\, \sup _{\|f\|_{L} \leq 1} \mathbb{E}[f(\mathbf{h}(g(\mathbf{Y}))      )]-\mathbb{E}[f({\mathbf{h}}(\mathbf{\Lambda}_k))] \\
    &\le\, \sup _{\|f\|_{L} \leq 1} \mathbb{E}|f(\mathbf{h}(g(\mathbf{Y})))- f({\mathbf{h}}(\mathbf{\Lambda}_k)) |\\
    &\le\, \mathbb{E} \| \mathbf{h}(g(\mathbf{Y}))- {\mathbf{h}}(\mathbf{\Lambda}_k) \|_2 \\
    &=\, \E \sqrt{ \sum_{i=1}^{p}  (\mathbf{h}_i({g}(\mathbf{Y})) - {\mathbf{h}}_i(\mathbf{\Lambda}_k) )^2}\\
    &\le\,  \sum_{i=1}^{p} \E |\mathbf{h}_i({g}(\mathbf{Y})) - {\mathbf{h}}_i(\mathbf{\Lambda}_k)|, 
\end{aligned}
\label{eq_EC19}
\end{equation}
where the first equation arises from $\mu\overset{\mathcal{D}}{=}\mathbf{h}(\mathbf{\Lambda}_k)$. This is because $\tilde{\mathbf{h}}\overset{\mathcal{D}}{=}\mathbf{h}$, $\mathbf{\Lambda}\overset{\mathcal{D}}{=}\mathbf{\Lambda}_k$, $\tilde{\mathbf{h}}$ is independent of $\mathbf{\Lambda}$, and $\mathbf{h}$ is independent of $\mathbf{\Lambda}_k$. The inequalities are from definition and standard Cauchy inequalities. Note that for any $i=1,2,\ldots,p$, the difference $\mathbf{h}_i({g}(\mathbf{Y})) - {\mathbf{h}}_i(\mathbf{\Lambda}_k)$ is given by
\[
\mathbf{h}_i({g}(\mathbf{Y})) - {\mathbf{h}}_i(\mathbf{\Lambda}_k) = M_i({g}^{(i)}(\mathbf{Y})) - {M_i}(\mathbf{\Lambda}_k^{(i)})
\]
in which $(M_i(t):t\ge 0)$ is a unit-rate Poisson process whose randomness is independent of any other source of randomness, ${g}^{(i)}$ represents the $i$-th element of the output of ${g}$, and $\mathbf{\Lambda}^{(i)}_k$ is the $i$-th element of $\mathbf{\Lambda}_k$. Therefore, conditional on ${g}(\mathbf{Y})$ and $\mathbf{\Lambda}_k$, 
\[
\E (|M_i({g}^{(i)}(\mathbf{Y})) - {M_i}(\mathbf{\Lambda}^{(i)}_k)| \,\big\lvert\, {g}(\mathbf{Y}),\mathbf{\Lambda}_k) = |{g}^{(i)}(\mathbf{Y}) - \mathbf{\Lambda}_k^{(i)}|
\]
because of the independent and stationary increments property for unit-rate Poisson processes. As a result, continuing from (\ref{eq_EC19}),
\begin{align*}    W(\mu_g,\mu)\, &\le \,  \sum_{i=1}^{p} \E |\mathbf{h}_i({g}(\mathbf{Y})) - {\mathbf{h}}_i(\mathbf{\Lambda})|\\
     &\le \, \sum_{i=1}^{p} \E |{g}^{(i)}(\mathbf{Y}) - \mathbf{\Lambda}^{(i)}_k|\\
     &\le \sqrt{p} \, \E \|{g}(\mathbf{Y}) - \mathbf{\Lambda}_k\|_2\\
     &\leq\, \sqrt{p}\, \left(W({g}(\mathbf{Y}),\mathbf{\Lambda})+\frac{1}{k}\right),\numberthis \label{eq:proof_wassNwass}
\end{align*}
where the second last step follows the Chebyshev inequality and the last step follows (\ref{eq:proof_setlambda}). Taking $k\to\infty$, we have
\begin{equation*}
    W(\mu_g,\mu)\leq \sqrt{p}(W(g(\mathbf{Y})),\mathbf{\Lambda}).
\end{equation*}
Therefore, when $W({g}(\mathbf{Y}),\mathbf{\Lambda})\leq \delta$ is guaranteed by the generator class, we have $W(\mu_g,\mu)\leq \sqrt{p}\delta$.

\subsection{Discriminator Approximation Error}
In this section, we control the discriminator approximation error term $d_{\mathcal{F}}(\mu_g,\hat{\mu}_n) - W(\mu_g,\hat{\mu}_n)$. We have the following lemma, the proof of which is given in the next section \ref{append:lem}.
\begin{lemma}
	For the neural network functional class $\mathcal{F}_{\text{NN}}(\kappa,L,p,K,\varepsilon_f)$ defined by \ref{eq:discriminator_class} and the $1$-Lipchitz class denoted as
	\begin{equation*}
  \mathcal{F}_{W}:=\{f:\mathbb{R}^p\to\mathbb{R}|\Vert f\Vert_{L}\leq 1\} ,
	\end{equation*}
	we have $\forall f_\omega\in\mathcal{F}_{\text{NN}}$, $\forall B>0$, $\exists f\in\mathcal{F}_{W}$, such that $\Vert f-f_\omega\Vert_{L^{\infty}[0,B]^p}\leq 3\varepsilon_f$.
\label{lem:analysis}
\end{lemma}
Using lemma \ref{lem:analysis}, We can upper bound the discriminator approximation error term as follows. First, for two arbitrary distributions $\pi_1$ and $\pi_2$ with bounded support, we have
\begin{equation}
\label{eq:dis_approx_error_sec}
\begin{aligned}
&d_{\mathcal{F}}(\pi_1,\pi_2) - W(\pi_1,\pi_2)\\
=&\sup_{f_\omega\in\mathcal{F}_{\text{NN}}}\left[\mathbb{E}_{\mathbf{X}\sim\pi_1}f_\omega(\mathbf{X})-\mathbb{E}_{\mathbf{X}\sim\pi_2}f_\omega(\mathbf{X})\right]-
\sup_{\Vert f\Vert_L\leq 1}\left[\mathbb{E}_{\mathbf{X}\sim\pi_1}f(\mathbf{X})-\mathbb{E}_{\mathbf{X}\sim\pi_2}f(\mathbf{X})\right]\\
=&\inf_{\Vert f\Vert_L\leq 1}\sup_{f_\omega\in\mathcal{F}_{\text{NN}}}\mathbb{E}_{\mathbf{X}\sim\pi_1}\left[f_\omega(\mathbf{X})-f(\mathbf{X})\right]+
\inf_{\Vert f\Vert_L\leq 1}\sup_{f_\omega\in\mathcal{F}_{\text{NN}}}\mathbb{E}_{\mathbf{X}\sim\pi_2}\left[f(\mathbf{X})-f_\omega(\mathbf{X})\right]\\
\leq &\inf_{\Vert f\Vert_L\leq 1}\sup_{f_\omega\in\mathcal{F}_{\text{NN}}}2\Vert f-f_\omega\Vert_\infty\leq 6\varepsilon_f.
\end{aligned}
\end{equation}

At this point, we can naturally provide a further explanation of the $\varepsilon_f$-constraint assumption on $\mathcal{F}_{\text{NN}}$, i.e., $\vert f_\omega(x)-f_\omega(y)\vert\leq \Vert x-y\Vert+2\varepsilon_f$. First, the purpose of this assumption is to ensure that $d_\mathcal{F}(,)-W(,)$ can be upper bounded by $\varepsilon_f$ multiplied with some constant. In other words, we aim to provide the condition for deriving Lemma \ref{lem:analysis} and (\ref{eq:dis_approx_error_sec}). Second, we choose to restrict $\mathcal{F}_{\text{NN}}$ in this way instead of directly assuming it to be $1$-Lipschitz, because the $1$-Lipschitz restriction weakens the approximation power of $\mathcal{F}_{\text{NN}}$. More specifically, according to the neural network function approximation Theorem \ref{thm-approx}, although there exists a function $f_\omega=\tilde{f}_\omega\circ\tilde{\sigma}\in\mathcal{F}_{NN}(\kappa,L,P,K)$ that satisfies $\Vert \tilde{f}_\omega-f\Vert_\infty\leq \delta$, there is no guarantee for this $f_\omega$ to be $1$-Lipchitz. However, if $\tilde{f}_\omega$ is an approximation of the $1$-Lipschitz function $f$ such that $\Vert \tilde{f}_\omega-f\Vert_\infty\leq \delta$, we know that $f_\omega$ at least satisfies the $\delta$-restriction, namely, 
\begin{equation}
\begin{aligned}
    \vert f_\omega(x)-f_\omega(y)\vert&=\vert \tilde{f}_\omega\circ\tilde{\sigma}(x)-\tilde{f}_\omega\circ\tilde{\sigma}(y)\vert\\
    &\leq\vert \tilde{f}_\omega\circ\tilde{\sigma}(x)-f\circ\tilde{\sigma}(x)\vert+
    \vert f\circ\tilde{\sigma}(x)-f\circ\tilde{\sigma}(y)\vert+
    \vert \tilde{f}_\omega\circ\tilde{\sigma}(y)-f\circ\tilde{\sigma}(y)\vert\\
    &\leq\Vert \tilde{\sigma}(x)-\tilde{\sigma}(y)\Vert+2\delta\leq \Vert x-y\Vert+2\delta.
\end{aligned}
\end{equation} Therefore, taking $\varepsilon_f=\delta$ eliminates some possible inconsistency in the proof.

Note that equation (\ref{eq:dis_approx_error_sec}) requires $\pi_1$ and $\pi_2$ to have bounded support, but the support of distribution $\mu_g$ is unbounded due to the Poisson simulator $\mathbf{h}$.  Therefore, for arbitrary constant $B>0$, let $\mu_g^B$ be the distribution clipped under $B$, namely, let $F_\mu$ denote the cumulative distribution function of $\mu_g$, and $F_\mu^B$ the cumulative distribution function of $\mu_g^B$, we have
\begin{equation*}
    F_\mu^B(\mathbf{x})=\left\{\begin{array}{ll}
    F_\mu(\mathbf{x}),& x_i\leq B,\forall i=1,2,\ldots,p;  \\
    \\
    1, & \exists i \text{ s.t. } x_i>B.
    \end{array}\right.
\end{equation*}
Further, let $\mathbf{x}^B=(x_1^B,x_2^B\ldots,x_p^B)$, where $x_i^B=\min\{x_i,B\}$. With these, we decompose the discriminator approximation error $d_\mathcal{F}(\mu_g,\hat{\mu}_n)-W(\mu_g,\hat{\mu}_n)$ as follows:
\begin{equation}
    d_\mathcal{F}(\mu_g,\hat{\mu}_n)-W(\mu_g,\hat{\mu}_n)\leq d_\mathcal{F}(\mu_g^B,\mu_g)+d_\mathcal{F}(\mu_g^B,\hat{\mu}_n)-W(\mu_g^B,\hat{\mu}_n)+W(\mu_g^B,\mu_g).
\end{equation}
Note that $\hat{\mu}_n$ has bounded support since it is the empirical distribution. According to equation (\ref{eq:dis_approx_error_sec}), for arbitrary large enough $B$, $d_\mathcal{F}(\mu_g^B,\hat{\mu}_n)-W(\mu_g^B,\hat{\mu}_n)$ can always be bounded by $6\varepsilon_f$. We next control the bounding error terms $d_\mathcal{F}(\mu_g^B,\mu_g)$ and $W(\mu_g^B,\mu_g)$. Since
\begin{equation}
\begin{aligned}
    d_\mathcal{F}(\mu_g^B,\mu_g)&=\sup_{f_\omega\in\mathcal{F}_{\text{NN}}}\mathbb{E}_{\mathbf{X}\sim\mu_g}f_\omega(\mathbf{X})-\mathbb{E}_{\mathbf{X}\sim\mu_g^B}f_\omega(\mathbf{X})\\
    &=\sup_{f_\omega\in\mathcal{F}_{\text{NN}}}\mathbb{E}_{\mathbf{X}\sim\mu_g}\left[f_\omega(\mathbf{X})-f_\omega(\mathbf{X}^B)\right]\\
    &=\sup_{f_\omega\in\mathcal{F}_{\text{NN}}}\int_{\mathbb{R}_+^p\backslash[0,B]^p} [f_\omega(\mathbf{x})-f_\omega(\mathbf{x}^B)]d\mu_g(\mathbf{x})\\
    &\leq \sup_{f_\omega\in\mathcal{F}_{\text{NN}}}\int_{\mathbb{R}_+^p\backslash[0,B]^p}[\Vert \mathbf{x}-\mathbf{x}^B\Vert+\varepsilon_f] d\mu_g(\mathbf{x}),\\
\end{aligned}
\end{equation}
where the last inequality is due to the $\varepsilon_f$-restrction on $\mathcal{F}_\text{NN}$. By the tail order of the distribution $\mu_g$, which is the same as the tail order of the Poisson distribution, we can find $B>0$ such that $d_\mathcal{F}(\mu_g^B,\mu_g)<\varepsilon_f$, and $W(\mu_g^B,\mu_g)$ can be bounded similarly. Therefore, we have
\begin{equation}
    d_{\mathcal{F}}(\mathbf{h}(g(\mathbf{Y})),\hat{\mu}_n) - W(\mathbf{h}(g(\mathbf{Y})),\hat{\mu}_n)\leq 8\varepsilon_f.
\end{equation}

\subsection{Statistical Error}

We first restate a special case of Theorem $1$ of \cite{fournier2015rate} as lemma \ref{lem:fournier}. Note that $\mathcal{P}(\mathbb{R}^p)$ stands for the set of all probability measures on $\mathbb{R}^p$.
\begin{lemma}[\cite{fournier2015rate}, Theorem 1]
\label{lem:fournier}
Let $(X_i)_{i\geq 1}$ be random variables with common law $\mu$, and $\hat{\mu}_n:=n^{-1}\sum_{i=1}^n\delta_{X_i}$. Let $\mu\in\mathcal{P}(\mathbb{R}^p)$ with finite $k$-th moment denoted as $M_k(\mu)$($k>2$). There exists a constant $C$ depending on $p$ and $k$ such that, for all $n\geq 1$,
\begin{equation*}
    \mathbb{E}(W(\mu,\hat{\mu}_n))\leq C M_k^{1/k}(\mu)\times\left\{\begin{array}{ll}
      n^{-1/2}+n^{-(k-1)/k},   &\text{if }p<2, k\not=2;  \\
      \\
      n^{-1/2}\log(1+n)+ n^{-(k-1)/k},   & \text{if }p=2,k\not=2;
      \\
      \\
      n^{-1/p}+n^{-(k-1)/k}, & \text{if }p>2,k\not=\displaystyle{p/(p-1)}.
    \end{array}\right.
\end{equation*}
\end{lemma}
Since the multi-dimensional Poisson distribution with bounded random intensity have finite moment of any order, we can always find large enough $k$ such that $n^{-(k-1)/k}$ is not of leading order. Therefore, we have
\begin{equation}
    \mathbb{E}(W(\mu,\hat{\mu}_n))=\left\{
    \begin{array}{ll}
         O(n^{-1/2}),&\text{if }p=1;  \\
         \\ O(n^{-1/2})\log(1+n),&\text{if }p=2; \\
         \\
     O(n^{-1/p}),&\text{if }p>2.
    \end{array}\right.
    \label{eq:stat_err_fournier}
\end{equation}

Further, theorem 14 of \cite{fournier2015rate} upper bounds the Wasserstein statistical error for $\rho$-mixing stationary sequences. A stationary sequence of random variables $(X_n)_{n\geq 1}$ with common law $\mu$ is said to be $\rho$-mixing, for some $\rho:\mathbb{N}\to\mathbb{R}^+$ with $\rho_i\to 0$, if for all $f,g\in L^2(\mu)$ and all $i,j\geq 1$,
\begin{equation*}
    \text{Cov}(f(X_i),g(X_j))\leq \rho_{|i-j|} \sqrt{\text{Var}(f(X_i)) \text{Var}(g(X_j))}.
\end{equation*}
A special case of Theorem 14 is restated as lemma \ref{lem:founier_weak_correlation}.
\begin{lemma}[\cite{fournier2015rate}, Theorem 14]
\label{lem:founier_weak_correlation}
Consider a stationary sequence of random variables $(X_i) _{i\geq 1}$ with common law $\mu$, and $\hat{\mu}_n:=n^{-1}\sum_{i=1}^n\delta_{X_i}$. Assume that the sequence is $\rho$-mixing, for some $\rho:\mathbb{N}\to\mathbb{R}^+$ satisfying $\sum_{i\geq 0}\rho_i<\infty$. Let $\mu\in\mathcal{P}(\mathbb{R}^p)$ with finite $k$-th moment denoted as $M_k(\mu)(k>2)$. There exists a constant $C$ depending on $p$, $k$, $M_k(\mu)$ and $\rho$ such that, for all $n\geq 1$,
\begin{equation*}
    \mathbb{E}(W(\mu,\hat{\mu}_n))\leq C\left\{\begin{array}{ll}
      n^{-1/2}+n^{-(k-1)/k},   &\text{if }p<2, k\not=2;  \\
      \\
      n^{-1/2}\log(1+n)+ n^{-(k-1)/k},   & \text{if }p=2,k\not=2;
      \\
      \\
      n^{-1/p}+n^{-(k-1)/k}, & \text{if }p>2,k\not=\displaystyle{p/(p-1)}.
    \end{array}\right.
\end{equation*}
\end{lemma}
Similarly to equation (\ref{eq:stat_err_fournier}), we can take sufficiently large $k$, and derive
\begin{equation}
   \mathbb{E}(W(\mu,\hat{\mu}_n))=\left\{
    \begin{array}{ll}
         O(n^{-1/2}),&\text{if }p=1;  \\
         \\ O(n^{-1/2})\log(1+n),&\text{if }p=2; \\
         \\
     O(n^{-1/p}),&\text{if }p>2.
    \end{array}\right. 
\end{equation}
which gives us the same estimate as in the independent case. Since $p$ is expected to be bigger than $2$, in the following we only consider the convergence order for $p>2$.

\subsection{Summarizing and Balancing}
Summarizing equations (\ref{eq:decomp_statErr})-(\ref{eq:decomp_g_approx}) and (\ref{eq:dis_final}), we have
\begin{equation}
    \begin{aligned}
    W(\mu_n^*,\mu)\leq \sqrt{p}(1+A)^2(\text{GE}+\text{DAE}+\text{DPE}+3\times\text{SE})+\left(\frac{2C_\mathbf{\Lambda}e}{A}\right)^A,
    \end{aligned}
\label{eq:final}
\end{equation}
where GE stands for generator approximation error $W(\mu_g,\mu)$, DAE stands for discriminator approximation error $d_{\mathcal{F}}(\mu_g,\hat{\mu}_n)-W(\mu_g,\hat{\mu}_n)$, DPE stands for discriminative power error $W(\tilde{\sigma}\left[\mu_n^*\right],\tilde{\sigma}[\hat{\mu}_n])-\tilde{d}_{\mathcal{F}}(\tilde{\sigma}\left[\mu_n^*\right],\tilde{\sigma}[\hat{\mu}_n])$, and SE stands for statistical error $W(\mu,\hat{\mu}_n)$.

Since the statistical error is of order $O(n^{-1/p})$, we can take $O(n^{-1/p})$ as the order for all other error terms for balancing. According to the network approximation theorem \ref{thm-approx}, the orders of the parameters of the generator class $\mathcal{G}_{\text{NN}}(\bar{\kappa},\bar{L},\bar{P},\bar{K})$ are given as 
\begin{equation*}
    \bar{\kappa}=O(n^{1/p}),\quad \bar{L}=O(\log n),\quad \bar{P}=O(n^{1/p}),\quad \bar{K}=O(n^{1/p}\log n),
\end{equation*}
and the orders of the parameters of the discriminator class $\mathcal{F}_\text{NN}(\kappa,L,P,K)$ are given as
\begin{equation*}
    \kappa=O(n^{1/p}),\quad L=O(\log n),\quad P=O(n^{1/p}),\quad K=O(n^{1/p}\log n),\quad \varepsilon_f=O(n^{-1/p}).
\end{equation*}

We next balance the order of $A$. According to (\ref{eq:final}), we can take the order of $A$ to statisfy
\begin{equation*}
    A\log A=O(\log n),
\end{equation*}
which means the order of $A$ is between $O(\log n)^{1/2}$ and $O(\log n)$. This gives the final result,
\begin{equation*}
    W(\mu_n^*,\mu)\leq O(n^{-\frac{1}{p}}(\log n)^2).
\end{equation*}

\section{Proof of Lemma \ref{lem:analysis} }
\label{append:lem}
In this section, we present the proof of lemma \ref{lem:analysis}. We first restate the lemma.
\begin{lemma}
	For the neural network functional class $\mathcal{F}_{\text{NN}}(\kappa,L,P,K,\varepsilon_f)$ defined by \ref{eq:discriminator_class} and the $1$-Lipchitz class denoted as
	\begin{equation*}
	   \mathcal{F}_{W}:=\{f:\mathbb{R}^p\to\mathbb{R}|\Vert f\Vert_{L}\leq 1\} ,
	\end{equation*}
	we have $\forall f_\omega\in\mathcal{F}_{\text{NN}}$, $\forall B>0$, $\exists f\in\mathcal{F}_{W}$, such that $\Vert f-f_\omega\Vert_{L^{\infty}[0,B]^p}\leq 3\varepsilon_f$, where $\Vert\cdot \Vert_{L^{\infty}[0,B]^p}$ is the infinity norm on $[0,B]^p$.
\end{lemma}

\noindent\textbf{Proof}: For fixed $f_\omega\in\mathcal{F}_{\text{NN}}$, without loss of generality, we assume that $p=1, B=1$ and $f_\omega(0)=0$. We prove by contradiction. For $f\in\mathcal{F}_W$, let
\begin{equation*}
x(f):=\sup\{x\in[0,1]:|f_\omega(x^\prime)-f(x^\prime)|\leq 3\varepsilon_f,\forall x^\prime\leq x\},
\end{equation*}
and
\begin{equation}
f^*=\arg\sup_{f\in\mathcal{F}_W}x(f).
\label{eq-cond-sup}
\end{equation}

We first show that  $x^*:=\sup_{f\in\mathcal{F}_W}x(f)>0$ and that $f^*$ exists. Note that for $x^\prime<\varepsilon_f$, we have, by definition of the functional class $\mathcal{F}_{\text{NN}}(\kappa,L,P,K,\varepsilon_f)$,
\begin{equation*}
    \vert f_\omega(x^\prime)-f_\omega(0)\vert \leq \vert x^\prime\vert+2\varepsilon_f<3\varepsilon_f.
\end{equation*}
Therefore, taking $f(x)\equiv f_\omega(0)$ yields an approximation of $f_\omega$ with precision of $3\varepsilon_f$ on $[0,\varepsilon_f]$, implying that $x^*$ is at least $\varepsilon_f$. To demonstrate the existence of $f^*$, suppose that $\vert f_\omega(x)\vert<C$ on $[0,1]$. The $1$-Lipschitz functional class bounded by $2C$ on $[0,1]$, denoted as $\mathcal{F}_W^C$ is uniformly bounded and equicontinuous. Arzel$\grave{a}$-Ascoli theorem suggests that $\mathcal{F}_W^C$ is sequentially compact. Further, for any convergent sequence in $\mathcal{F}_W^C$, it can be verified that the limit also lies in $\mathcal{F}_W^C$. Therefore, $\mathcal{F}_W^C$ is a compact set. It remains to be proved that $x(f)$ is upper semi-continuous on $\mathcal{F}_W$, so that the supremum point $f^*$ exists. In fact, for a fixed element $f_0\in\mathcal{F}_W$, for all small enough $\varepsilon>0$, let
\begin{equation*}
    \delta=\sup_{x(f_0)\leq x\leq x(f_0)+\varepsilon}\left[\vert f_\omega(x)-f_0(x)\vert-3\varepsilon_f\right].
\end{equation*}
By definition of $x(f)$, we have $\delta>0$, and for all $f\in\mathcal{F}_W$ such that $\Vert f-f_0\Vert_{L^{\infty}[0,B]^p}< \delta$, we have $x(f)< x(f_0)+\varepsilon$. Therefor, $x(f)$ is upper semi-continuous, and the existence of $f^*$ is proved.

We next return to the  contradiction assumption, which suggests that $x^*<1$. Note that both $f_\omega$ and $f$ are continuous. Therefore, by the definition of $x^*$, we have
\begin{equation*}
|f^*(x^*)-f_w(x^*)|=3\varepsilon_f.
\end{equation*}
Without loss of generality, let $f_\omega(x^*)=f^*(x^*)+3\varepsilon_f$. Also,
\begin{equation}
\forall \Delta_+>0, \exists x_{\Delta}\in(x^*,\min\{x^*+{\Delta_+},1\}), \text{ s.t. } f_\omega(x_\Delta)>f^*(x^*)+(x_\Delta-x^*)+3\varepsilon_f. 
\label{eq-cond-low}
\end{equation}

Now, we claim that $\exists\Delta_->0$,
\begin{equation*}
\frac{f^*(x^*)-f^*(x)}{x^*-x}=1,\quad \forall x\in[x^*-\Delta_-, x^*).
\end{equation*}
If this is contradicted, we can find some $\Delta<\varepsilon_f$, and
\begin{equation*}
f^*(x^*)-\Delta<f^*(x^*-\Delta)<f^*(x^*)+3\varepsilon_f-\Delta.
\end{equation*} 
In this case, we can move $f^*$ upward on $[x^*-\Delta, x^*]$ by modifying it into 
\begin{equation*}
\tilde{f}^*(x)=\left\{\begin{array}{ll}
f^*(x),&x\in[0,x^*-\Delta];\\
\\ 
f^*(x^*-\Delta)+x-(x^*-\Delta),&x\in(x^*-\Delta,x^*],
\end{array}\right.
\end{equation*}
so that $\tilde{f}^*(x^*)>f^*(x^*)$, and $\Vert\tilde{f}^*(x^*)\Vert_{L}\leq 1$ still holds. Also, note that $\forall x\in[x^*-\Delta,x^*]$,
\begin{equation*}
\begin{aligned}
&f_\omega(x)\geq f_\omega(x^*)+x-x^*-3\varepsilon_f=f^*(x^*)+x-x^*>\tilde{f}^*(x)+(x^*-x)-3\varepsilon_f,\\
& f_\omega(x)\leq f^*(x)+3\varepsilon_f\leq \tilde{f}^*(x)+3\varepsilon_f.
\end{aligned}
\end{equation*}
so $x(\tilde{f}^*)>x(f^*)$, which contradicts (\ref{eq-cond-sup}). Therefore, the claim is valid.

Let 
\begin{equation*}
x_1=\inf\{x\geq0~\Big|~\frac{f^*(x^*)-f^*(x)}{x^*-x}=1\}
\end{equation*}
Note that, from (\ref{eq-cond-low}) we have $f_\omega(x_\Delta)>f^*(x_1)+(x_\Delta-x_1)+3\varepsilon_f$, therefore, $f_\omega(x_1)>f^*(x_1)$, which implies that $x_1>0$ since we can set $f_\omega(0)=f^*(0)$. In this case, we can find $\Delta^\prime>0$, such that
\begin{equation*}
f^*(x_1)-\Delta^\prime< f^*(x_1-\Delta^\prime)\leq f^*(x_1)+3\varepsilon_f-\Delta^\prime.
\end{equation*}
We can similarly define $\tilde{f}^*$ by moving $f^*$ upward on $[x_1-\Delta^\prime,x^*]$, namely,
\begin{equation*}
\tilde{f}^*(x)=\left\{\begin{array}{ll}
f^*(x),&x\in[0,x_1-\Delta^\prime];\\
\\ 
f^*(x_1-\Delta^\prime)+x-(x_1-\Delta^\prime),&x\in(x_1-\Delta^\prime,x^*],
\end{array}\right.
\end{equation*}
and verify that $\Vert\tilde{f}^*(x)\Vert_L\leq 1$ and $x(\tilde{f}^*)>x(f^*)$, which also contradicts (\ref{eq-cond-sup}). The proof is complete.

\section{Specifications of Two Versions of the Arrival Epochs Simulator}
\label{ec:arrival_epochs_simulator}
In this part, we give the specific formulation and simulation algorithm of the two versions of the arrival epochs simulator. Specifically the \textit{arrival epochs simulator based on piecewise constant rate} and the \textit{continuous piecewise linear arrival epochs simulator}. Both these two versions of arrival epochs simulator takes an arrival count vector $\mathbf{x}=(x_1,x_2,\ldots,x_p)$ as input, which records the arrival counts in the consecutive $p$ time intervals in $[0,T]$. The total number of arrivals within $[0,T]$ is $\sum_{i=1}^p x_i$. The arrival epochs simulator assigns an arrival epoch for each of these $\sum_{i=1}^p x_i$ arrivals.

Given the arrival count vector $\mathbf{x}$, the realization of the underlying random intensity process can be inferred. The main difference between the two versions of arrival epochs simulator is that the arrival epochs simulator based on piecewise constant rate approximates the realized intensity process as a piecewise constant intensity function, while the piecewise linear arrival epochs simulator approximates the realized intensity process as a piecewise linear intensity function. Once the intensity process is inferred, within each time interval, the exact arrival epochs are assigned according to the intensity process over that time interval. We next discuss the details of the two arrival epochs simulator, respectively in \ref{sec:pc_arriavl_epochs_simulator} and \ref{sec:pc_arrival_epochs_simulator}.

\subsection{Arrival Epochs Simulator based on Piecewise Constant Rate}
\label{sec:pc_arriavl_epochs_simulator}
The arrival epochs simulator based on piecewise constant rate approximates the underlying rate function $\lambda(t), t\in[0,T]$ as a piecewise constant function, with the specific form
$$
\lambda(t)=\lambda_j,\text{ for }\frac{(j-1)\cdot T}{p}<t\le \frac{jT}{p},\text{ for }j=1,2,\ldots,p,
$$
where $\lambda_j$ is a constant for $j$-th period $\left(\frac{(j-1)T}{p},\frac{jT}{p}\right]$. Under this approximation, given the arrival count $x_j$ in the $j$-th time interval, the arrival epochs are order statistics of $x_j$ iid uniform random variables supported on the $j$-th time interval. Therefore, the arrival epochs simulator simulates $x_j$ copies of iid uniform random variables supported on the $j$-th time interval, orders them, and returns them as the arrival epochs for the $j$-th time interval. Each time interval is handled separately and independently in the same pattern. We summarize the procedure for the arrival epochs simulator based on piecewise constant rate in Algorithm \ref{alg:pc_aes}.

\renewcommand{\algorithmicensure}{\textbf{Output: }}
\renewcommand{\algorithmicrequire}{\textbf{Require: }}

\begin{algorithm}[ht!] 
\caption{Arrival epochs simulator based on piecewise constant rate.}\label{alg:pc_aes}
\algorithmicrequire The arrival count vector $\mathbf{x}=(x_1,x_2,\ldots,x_p)$, the length of the time range $T$.

\algorithmicensure A list $E$ that contains the sequential arrival epochs for all the arrivals in time range $[0,T].$
\begin{algorithmic}[1]
\STATE $E\leftarrow\Phi$.
\FOR{$j=1,2, \ldots, p$}
\STATE Simulate $x_j$ iid copies of $\text{Uniform}\left(\frac{(j-1)T}{p},\frac{jT}{p}\right)$, denoted as $\{s_{ji}\}_{i=1}^{x_j}$.
\STATE $E\leftarrow E\cup\{s_{ji}\}_{i=1}^{x_j}$.
\ENDFOR
\STATE Sort $E$.
\RETURN $E$.
\end{algorithmic}
\end{algorithm}

\subsection{Arrival Epochs Simulator based on Piecewise Linear Rate}
\label{sec:pc_arrival_epochs_simulator}

In this subsection, we introduce an arrival epochs simulator based on piecewise linear arrival rate, which is a standard practice in the simulation literature and toolbox. 

If the underlying random intensity process lies in the continuous function space of $C[0,T]$, the aforementioned arrival epochs simulator may be unrealistic. In this case, even though it is impossible to recover the full continuous function of the realized intensity process by just observing the interval counts, we consider a \textit{continuous arrival epochs simulator based on piecewise linear rate} to avoid the implausible discontinuities caused by the first arrival epochs simulator. Specifically, we fit a continuous piecewise linear intensity function given the interval count vector $(x_1,x_2,\ldots,x_p)$. The fitting procedure follows that provided in \cite{zhengglynn2017}. Then, for a given time interval $\left(\frac{(i-1)T}{p},\frac{iT}{p}\right]$, we randomly simulate and sort $x_i$ copies of iid random variables according to a probability distribution supported on that time interval, whose density function is the normalized fitted intensity function on $\left(\frac{(i-1)T}{p},\frac{iT}{p}\right]$. See \ref{ec:arrival_epochs_simulator} for the computational details of the arrival epochs simulators.

For the arrival epochs simulator based on piecewise linear rate, the underlying intensity function is approximated by a piecewise linear function, where $\{\lambda(t): 0\le t\le T\}$ is specified such that  $\lambda(t)$ is linear over each of the time intervals $\left(0,\frac{T}{p}\right], \left(\frac{T}{p},\frac{2T}{p}\right], \ldots, \left(\frac{(p-1)T}{p},T\right]$. In particular, we use $\lambda_i$ to denote the intensity at the boundaries of all the time intervals. That is, $\lambda_j = \lambda(\frac{jT}{p})$ for $j=0,1,\ldots,p$. The full piecewise linear function can be determined by the intensity values on the boundaries of all the time intervals. Specifically,
$$
\lambda(t)=\lambda_{j-1}+\left(\frac{\lambda_j-\lambda_{j-1}}{T/p}\right)\left(t-\frac{(j-1)T}{p}\right),\text{ for }t_{j-1}\le t\le t_j, 1\le j\le p.
$$

Given the arrival count vector $\mathbf{x}$, the piecewise linear intensity function can be inferred using a maximum likelihood estimation approach introduced by \cite{zhengglynn2017}. Specifically, this approach takes the arrival count vector as inputs and simultaneously returns the estimations for $\lambda_j$'s for $j=0,1,\ldots,p$. The estimations are denoted as  $\hat{\lambda}_j$'s for $j=0,1,\ldots,p$. Then, the full intensity function can be estimated through interpolation of the estimated intensity values $\hat{\lambda}_j$'s. Specifically,
$$
\begin{aligned}
\hat{\lambda}(t)&=\hat{\lambda}_{j-1}+\left(\frac{\hat{\lambda}_j-\hat{\lambda}_{j-1}}{T/p}\right)\left(t-\frac{(j-1)T}{p}\right)
\\&=\left(\frac{\hat{\lambda}_j-\hat{\lambda}_{j-1}}{T/p}\right)t + \hat{\lambda}_{j-1}j-\hat{\lambda}_j(j-1), \text{ for }t\in \left[(j-1) \frac{T}{p},j\frac{T}{p}\right].
\end{aligned}
$$
With the estimated intensity function in hand, for each time interval, the exact arrival epochs can be simulated using the inversion method on each time interval. The detailed procedure of the arrival epochs simulator based on piecewise linear rate is given in  Algorithm \ref{alg:pl_aes}.

\renewcommand{\algorithmicensure}{\textbf{Output: }}
\renewcommand{\algorithmicrequire}{\textbf{Require: }}

\begin{algorithm}[ht!] 
\caption{Arrival epochs simulator based on piecewise linear rate.}\label{alg:pl_aes}
\algorithmicrequire The arrival count vector $\mathbf{x}=(x_1,x_2,\ldots,x_p)$, the length of the time interval $T$.

\algorithmicensure A list $E$ that contains the arrival epochs for all the arrivals in time interval $[0,T].$
\begin{algorithmic}[1]
\STATE $E\leftarrow\Phi$

\STATE Compute the estimators $\hat{\lambda}_j$ for $j=0,1,\ldots,p$ using the MLE approach.

\FOR{$j=1,2, \ldots, p$}

\STATE $m\leftarrow \frac{\hat{\lambda}_j-\hat{\lambda}_{j-1}}{T/p}$
\STATE $b\leftarrow\hat{\lambda}_{j-1}j-\hat{\lambda}_j(j-1)$
\STATE $t_0\leftarrow (j-1)\frac{T}{p}$
\STATE $t_1\leftarrow j\frac{T}{p}$

\STATE The estimated intensity function $\hat{\lambda}(t)$ over $j$-th interval is given by 
$$\hat{\lambda}(t)=mt+b, \text{for }t\in [t_0,t_1].$$

\STATE Define the integral of $\hat{\lambda}(t)$ over the time interval $[t_0,t]$ with $t$ taking value within $[t_0,t_1]$ as 
$$
\begin{aligned}
\Lambda(t)&\triangleq\int_{t_0}^t \hat{\lambda}(t)\,dt,
\\&=\frac{mt^2}{2}+bt-(\frac{mt_0^2}{2}+bt_0),\text{ for }t\in[t_0,t_1].
\end{aligned}
$$
\STATE Compute $\Lambda(t_0)=0$ and $\Lambda(t_1)=\frac{mt_1^2}{2}+bt_1-(\frac{mt_0^2}{2}+bt_0)$.

\STATE Simulate $x_j$ iid copies of $\text{Uniform}\left(\Lambda(t_0),\Lambda(t_1)\right)$, denoted as $\{u_{ji}\}_{i=1}^{x_j}$.

\STATE $s_{ji}\leftarrow \Lambda^{-1}(u_{ji})=\frac{1}{m}\left(-b+\sqrt{b^2+2m\left(\frac{mt_0^2}{2}+bt_0+u_{ji}\right)}\,\right),\text{ for }i=1,2,\ldots,x_j.$

\STATE $E\leftarrow E\cup\{s_{ji}\}_{i=1}^{x_j}$.
\ENDFOR
\STATE Sort $E$. 
\RETURN $E$.
\end{algorithmic}
\end{algorithm}

\section{What-if Simulation Using DS-WGAN}
\label{sec:what_if}
Suppose that the iid data $\mathbf{X_1},\mathbf{X_2},\ldots,\mathbf{X}_n$ were collected from exogenous user arrivals to a service system over $n$ days of operations in the past. Consider a what-if planning decision in the service system for the future three months, with the objective to satisfy a certain level of user satisfaction, say, on average waiting time. For example, the question can be \textbf{what is an appropriate workers scheduling design if the expected number of daily user arrivals in the future three months is $20\%$ larger, compared to that in the historical data?}

{To answer the what-if question and do the planning decision for future, there is a need to simulate the vector of hourly counts (and then the arrival epochs) that reflect the 20\% daily volume increase while realistically preserve the correlation structure of the underlying random intensity process. As an example, the optimization method by simulation and cutting plane (\cite{avramidis2010optimizing}) requires simulation of the system dynamics.} This need for simulation cannot be satisfied by re-sampling the arrivals from the historical data or by adding a multiplier to the hourly counts,  because the increased scale should be multiplied on the random intensity level, before the generation of Poisson counts. In this case, the DS-WGAN framework is a natural fit for the task. By feeding the data into DS-WGAN, a generator $\hat{g}_n$ can be estimated to learn the joint distribution of the underlying random intensities. To reflect the $20\%$ scale increase, DS-WGAN can generate new data as 
\begin{equation} \label{eq:what-if-simulation}
\tilde{\mathbf{X}}_1=\mathbf{h}(1.2 \cdot {g}^*_n(\mathbf{Y_1})),\, \tilde{\mathbf{X}}_2=\mathbf{h}(1.2 \cdot {g}^*_n(\mathbf{Y_2})),\,\ldots 
\end{equation}
We further remark that the standard WGAN framework is not able to facilitate the aforementioned what-if simulation task. DS-WGAN can be applied to the what-if problems because it not only captures underlying distributions of data as effectively as the WGAN framework, but also exploits the doubly stochastic structure.

\section{Estimating from a Down-sampled Dataset}
\label{sec:exper_downsample}
\begin{figure}[t]
    \centering

    \subfigure[Marginal mean]{
        \label{fig:downsample_cir_result.mean}
        \includegraphics[width=0.33\textwidth]{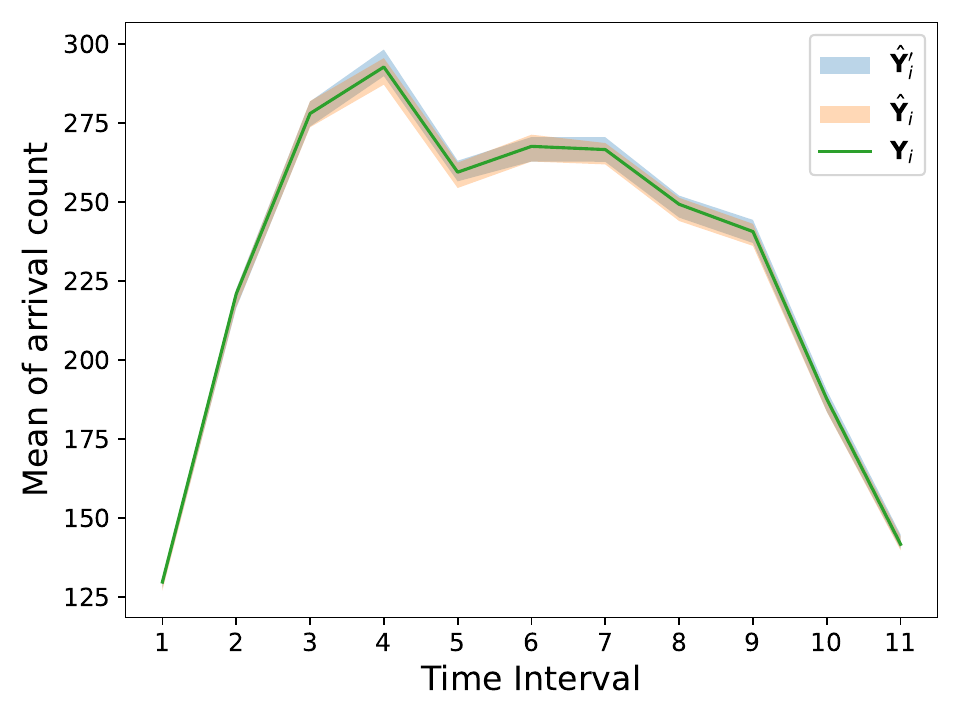}}
    \subfigure[Marginal variance]{
        \label{fig:downsample_cir.var}\includegraphics[width=0.33\textwidth]{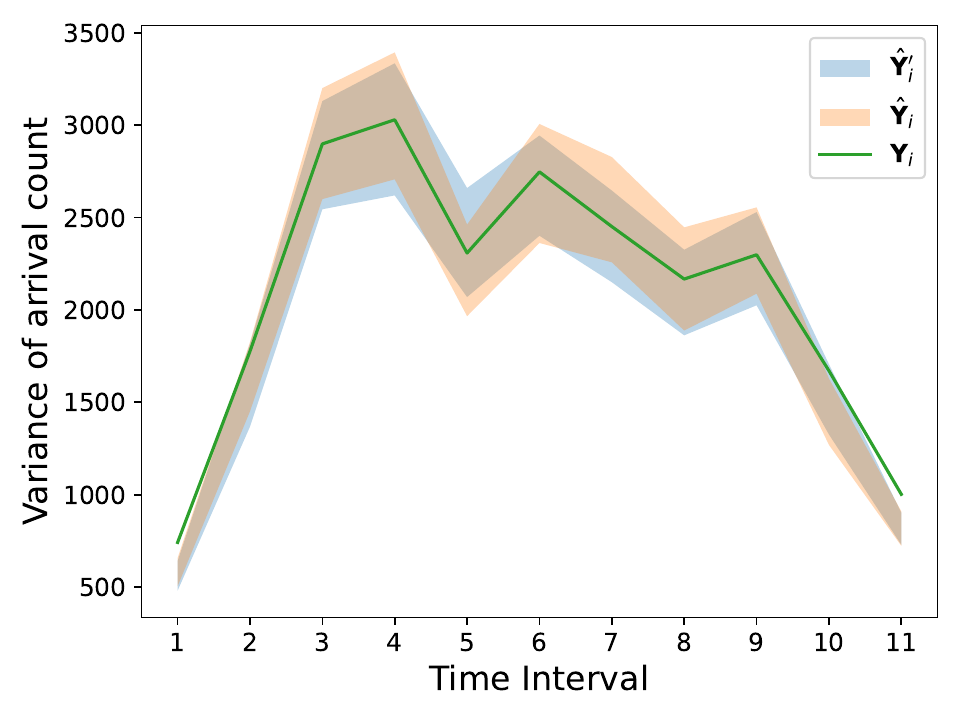}}

    \subfigure[Past and future correlation]{
        \label{fig:downsample_cir.corr}
        \includegraphics[width=0.33\textwidth]{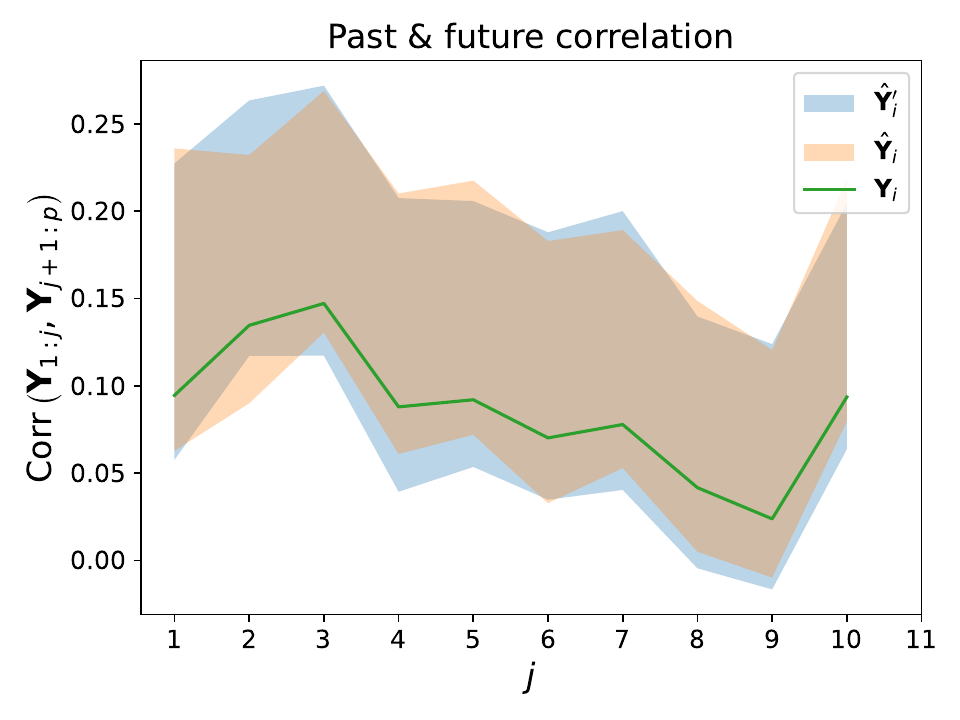}}
    \subfigure[Marginal Wasserstein distance]{
        \label{fig:downsample_cir.w_dist}
        \includegraphics[width=0.33\textwidth]{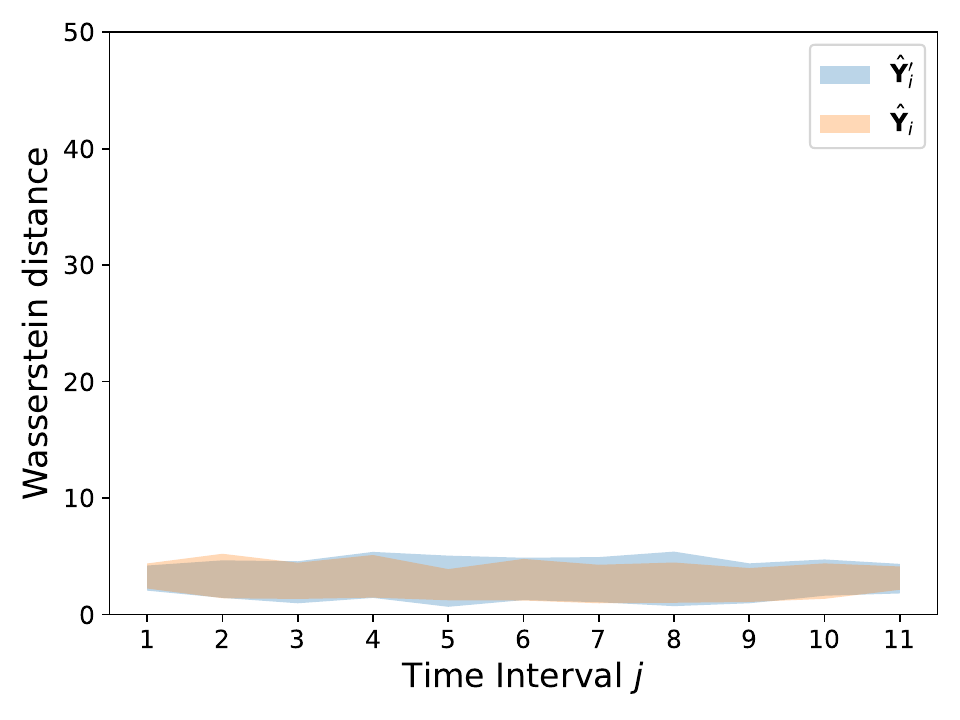}}

    \caption{Experiment results of Section \ref{sec:exper_downsample}: Compare the summary statistics of DS-WGAN trained on original data and downsampled data, CIR model}
    \label{fig:downsample_cir}
\end{figure}

\begin{figure}[h]
    \centering

    \subfigure[Marginal mean]{
        \label{fig:downsample_pgnorta_result.mean}
        \includegraphics[width=0.33\textwidth]{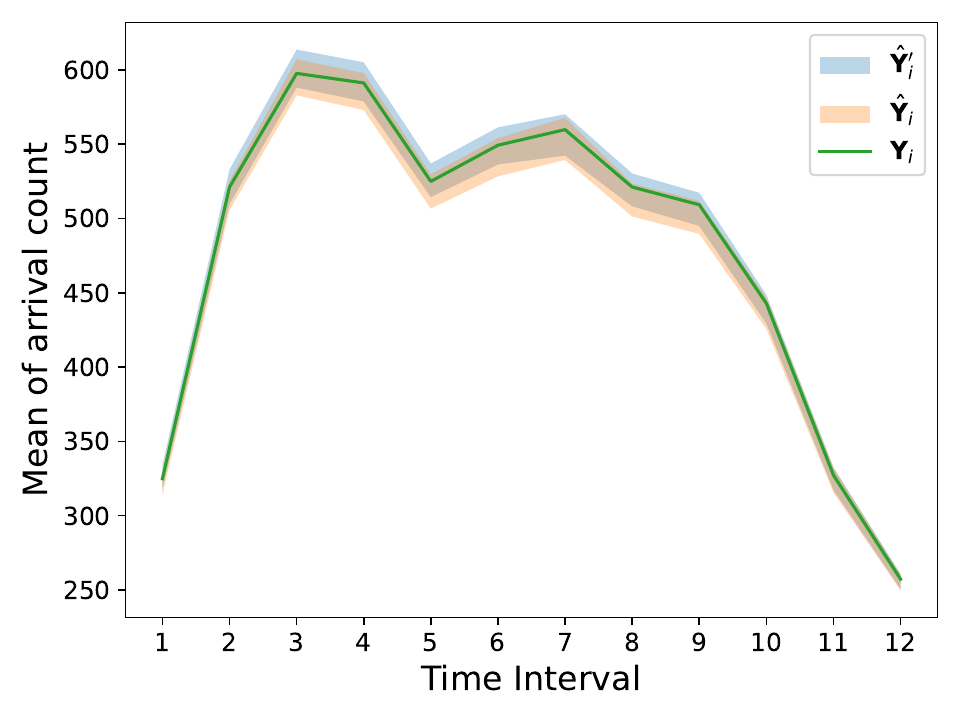}}
    \subfigure[Marginal variance]{
        \label{fig:downsample_pgnorta.var}\includegraphics[width=0.33\textwidth]{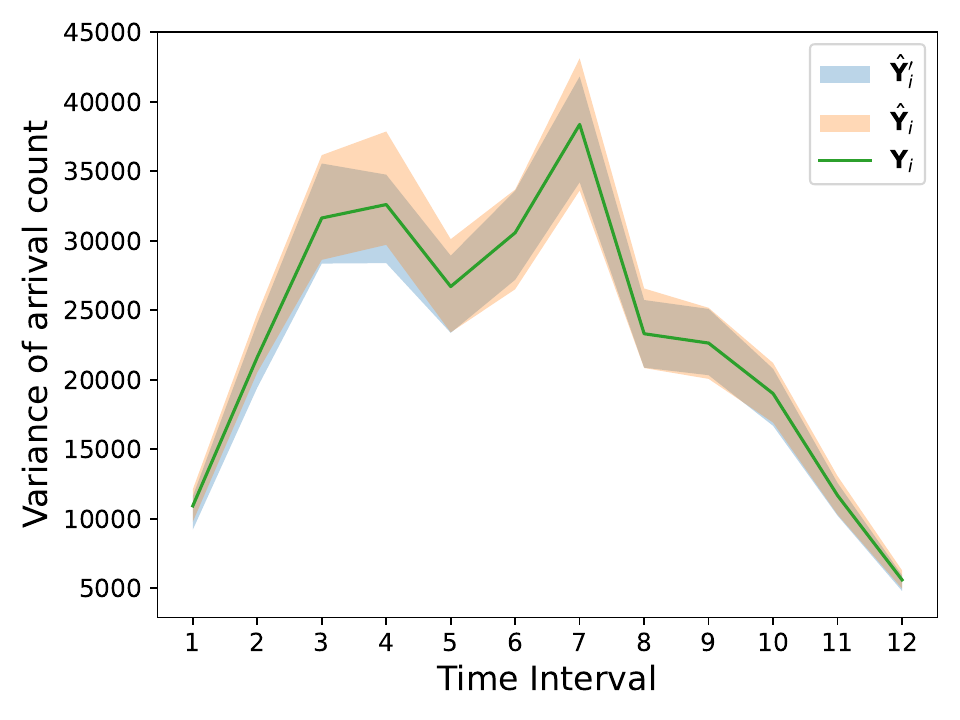}}

    \subfigure[Past and future correlation]{
        \label{fig:downsample_pgnorta.corr}
        \includegraphics[width=0.33\textwidth]{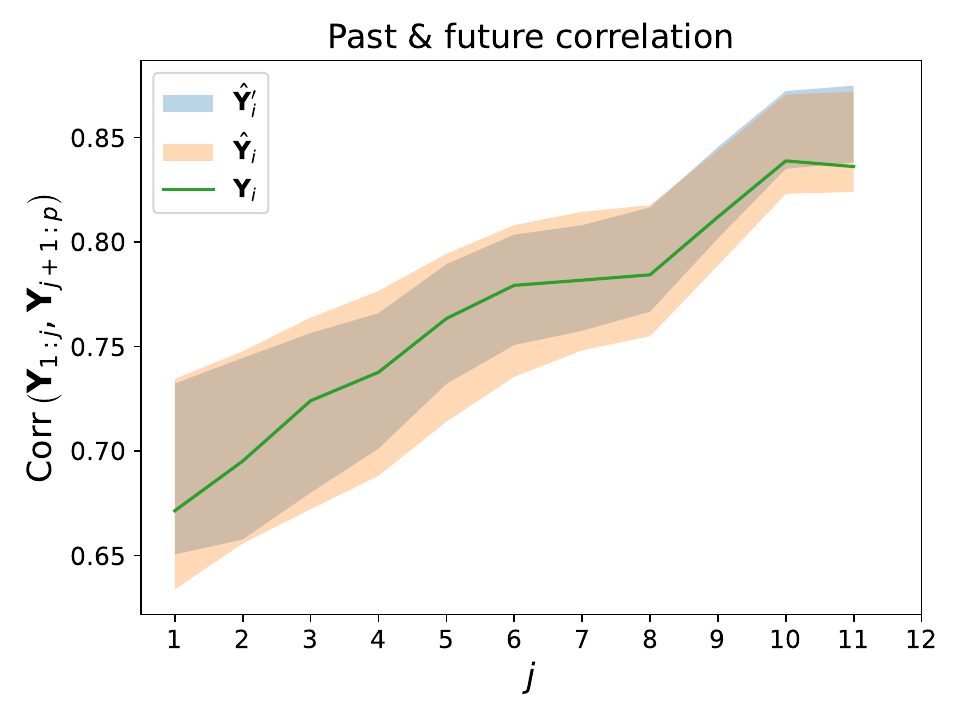}}
    \subfigure[Marginal Wasserstein distance]{
        \label{fig:downsample_pgnorta.w_dist}
        \includegraphics[width=0.33\textwidth]{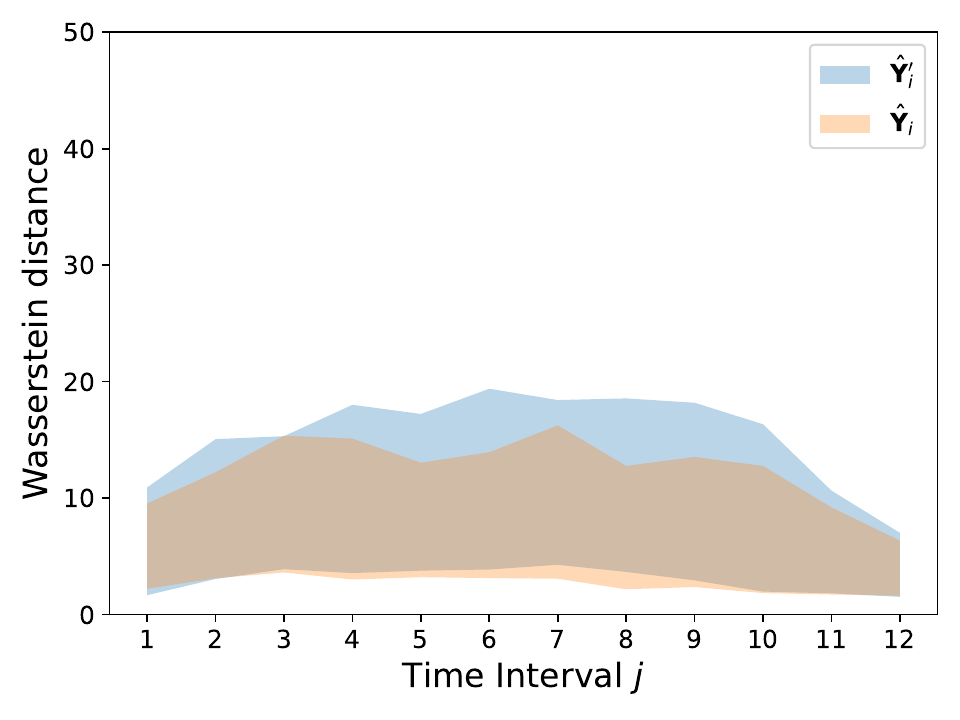}}

    \caption{Experiment results of Section \ref{sec:exper_downsample}: Compare the summary statistics of DS-WGAN trained on original data and downsampled data, PGnorta dataset}
    \label{fig:downsample_pgnorta}
\end{figure} 
\textcolor{black}{In many real service systems, the arrival data is recorded in great detail. For example, a bike-sharing company might record the start time of each order, and a call center may keep track of the number of inbound calls every minute. Although one viable approach is to model the original dataset with high time resolution directly, it might happen that the decision maker primarily cares about some statistics on a relatively low time resolution (the mean and standard deviation of hourly arrival count, for example). We can aggregate the original dataset to a downsampled dataset (which contains only hourly arrival count), then train our DS-WGAN model upon the downsampled dataset.}

\textcolor{black}{Consider the case that a decision maker only cares about some statistics for hourly arrival count. We can aggregate the original dataset $\left\{\mathbf{X}_{i}\right\}_{i=1}^{N}$, where $\mathbf{X}_i\triangleq(X_{i,1},X_{i,2},\ldots,X_{i,2n}),$ into hourly level resolution. Specifically, compute the downsampled dataset as $\left\{\mathbf{Y}_{i}\right\}_{i=1}^{N}$, where $\mathbf{Y}_i\triangleq(Y_{i,1},Y_{i,2},\ldots,Y_{i,n}), \text{ with } Y_{i,j} = X_{i,2j-1} + Y_{i,2j},j=1,2,\ldots,n$. We consider the following two options for modeling the dataset:
\begin{enumerate}
  \item Estimate DS-WGAN model on the original dataset directly. The output of the DS-WGAN simulator will be arrival count vectors of $2n$ dimensions. Denote the DS-WGAN simulator output as $\hat{\mathbf{X}}_i=(\hat{X}_{i,1},\hat{X}_{i,2},\ldots,\hat{X}_{i,2n})$. Aggregate $\hat{\mathbf{X}}_i$ into $\hat{\mathbf{Y}}'_i$ by $\hat{Y}'_{i,j}\triangleq \hat{{X}}_{i,2j-1}+\hat{X}_{i,2j},j=1,2,\ldots,n$.
  \item Estimate DS-WGAN model on the downsampled dataset $\left\{\mathbf{Y}_{i}\right\}_{i=1}^{N}$. Denote the DS-WGAN simulator output as $\hat{\mathbf{Y}}_i$, which are $n$-dimensional random vectors.
\end{enumerate}
Since the decision maker only cares about hourly arrival count, we compare the distribution of $\hat{\mathbf{Y}}'_i$ and $\hat{\mathbf{Y}}_i$ with $\mathbf{Y}_i$, in terms of marginal mean, marginal variance, marginal Wasserstein distance, as well as past-future correlation for hourly arrival count. }

We consider a service system whose arrival follows the CIR process in Section \ref{sec:cir_model}, and set the duration of operation each day as 11 hours. Another service system we adopted is the PGnorta model in Section \ref{sec:PGnorta_model}, which operates 12 hours each day, with the half-hour arrival count following a PGnorta model with preset parameters. The experiment results are given in Figure \ref{fig:downsample_cir} and \ref{fig:downsample_pgnorta}. We repeat the sampling of DS-WGAN model for 100 times to compute a 95\% confidence interval for the summary statistics of $\hat{\mathbf{Y}_i}$ and $\hat{\mathbf{Y}'_i}$. As shown in the results, both the summary statistics of $\hat{\mathbf{Y}_i}$ and $\hat{\mathbf{Y}'_i}$ can match  $\mathbf{Y}_i$, with only negligible differences.

{We give a brief remark on aggregation of periods. If one only concerns the statistics of the arrival count distribution on a lower time resolution (hourly statistics in our experiment) when we have access to a higher time resolution dataset (half-hourly resolution), it is likely that using the original or aggregated dataset does not make much difference in the estimation outcome of the arrival count distribution. Also, aggregating periods before training reduces the subsequent work and noise. That said, if our ultimate goal is to estimate a task-specific distribution constructed from the arrival count distribution(e.g.,the distribution of waiting time), aggregating the periods can result in loss of information and thus estimation bias. These aspects should be considered for deciding whether or not to aggregate the periods.}


\section{Discussion on Removing Different Percentages of Outliers}

In Section 6.3, an experiment is done on a call center data set where outliers are removed according to 2.5\% and 97.5\% percentiles for the data set. (Specific criterion for outliers is described in Section 6.3.) As a result, about 12\% of days of data were removed. In this appendix, we examine the experiment results when fewer of data are marked as outliers and removed. 

\begin{figure}[t]
    \centering

    \subfigure[Marginal mean]{
        \label{fig:callcenter.mean2}
        \includegraphics[width=0.33\textwidth]{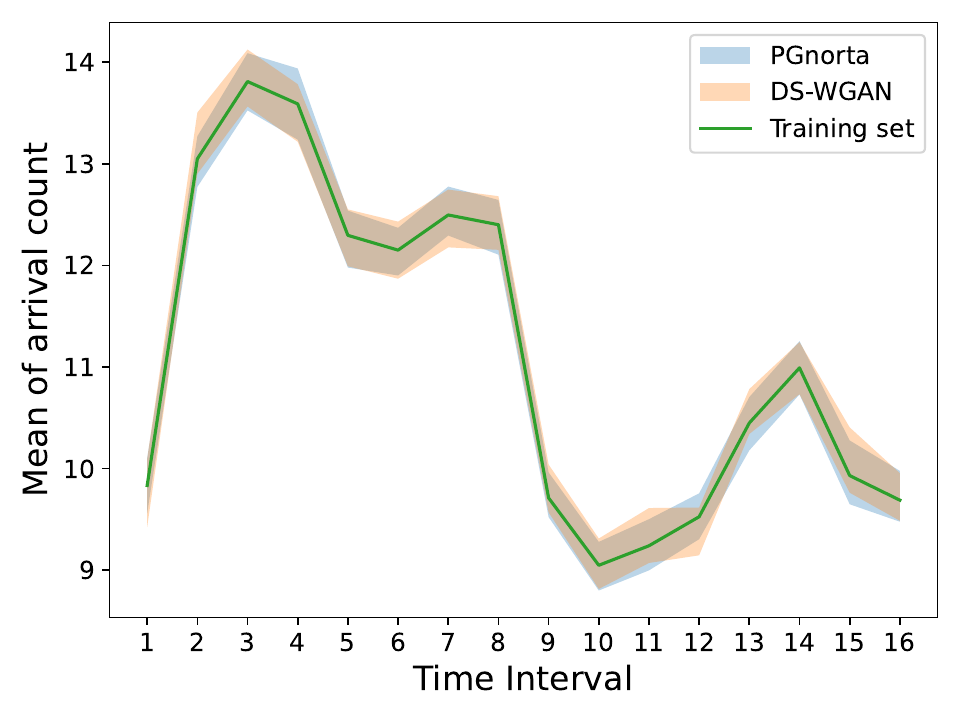}}
    \subfigure[Marginal variance]{
        \label{fig:callcenter.var2}\includegraphics[width=0.33\textwidth]{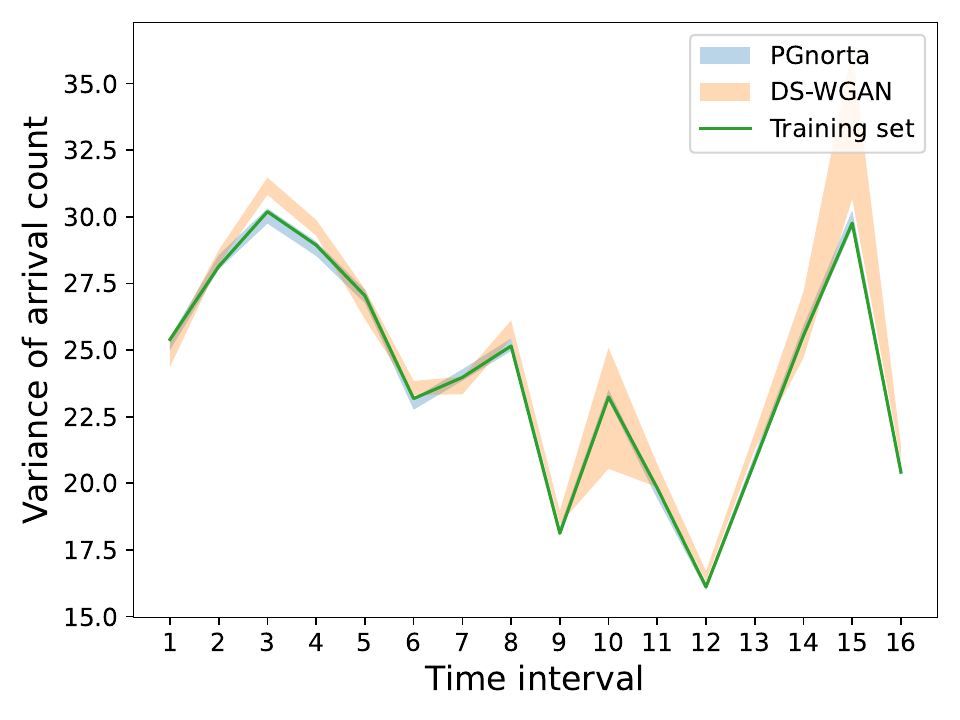}}

    \subfigure[Past and future correlation]{
        \label{fig:callcenter.corr2}
        \includegraphics[width=0.33\textwidth]{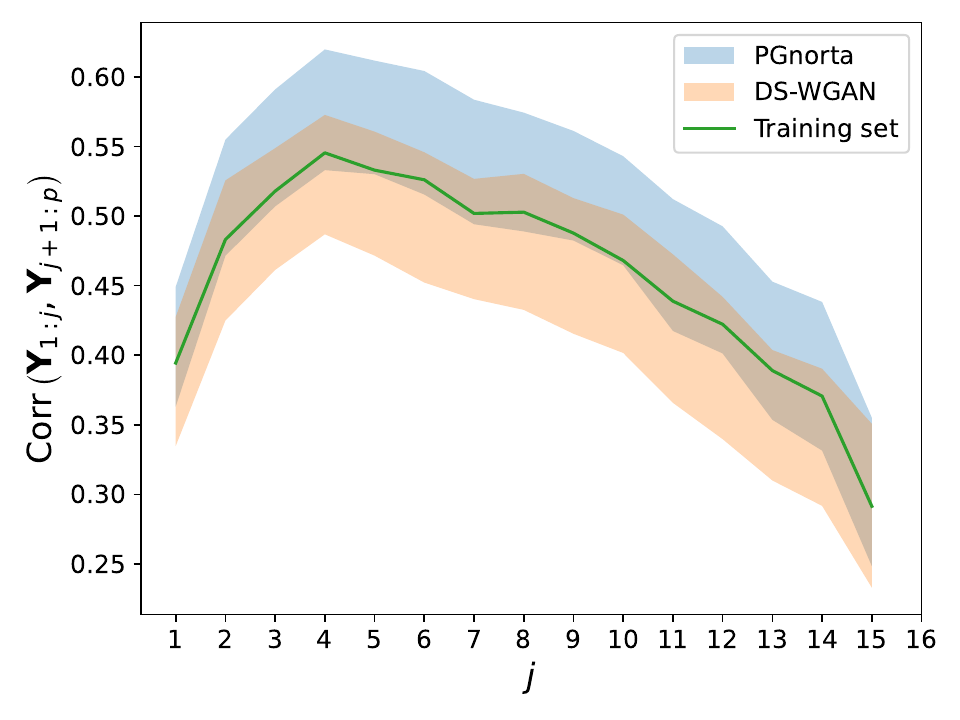}}
    \subfigure[Marginal Wasserstein distance]{
        \label{fig:callcenter.w_dist2}
        \includegraphics[width=0.33\textwidth]{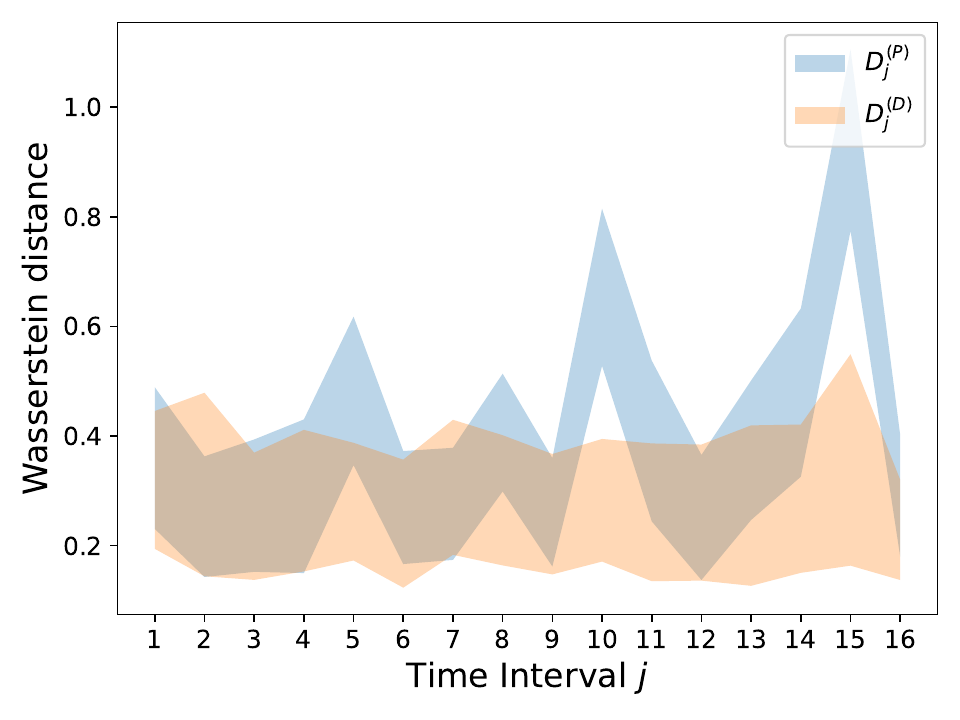}}

    \caption{Experiment results of Section \ref{sec:exper_callcenter_real}: with $2\%$ upper and lower quantiles of outliers removed.}
    \label{fig:callcenter_result_2}
\end{figure}

\begin{figure}[t]
    \centering

    \subfigure[Marginal mean]{
        \label{fig:callcenter.mean0.5}
        \includegraphics[width=0.33\textwidth]{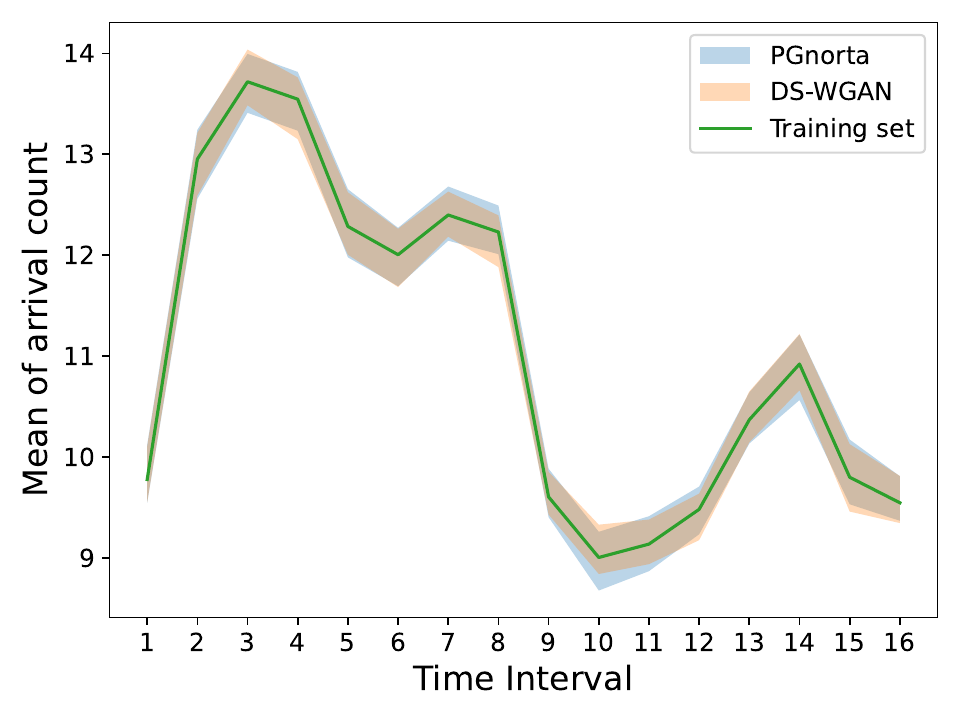}}
    \subfigure[Marginal variance]{
        \label{fig:callcenter.var0.5}\includegraphics[width=0.33\textwidth]{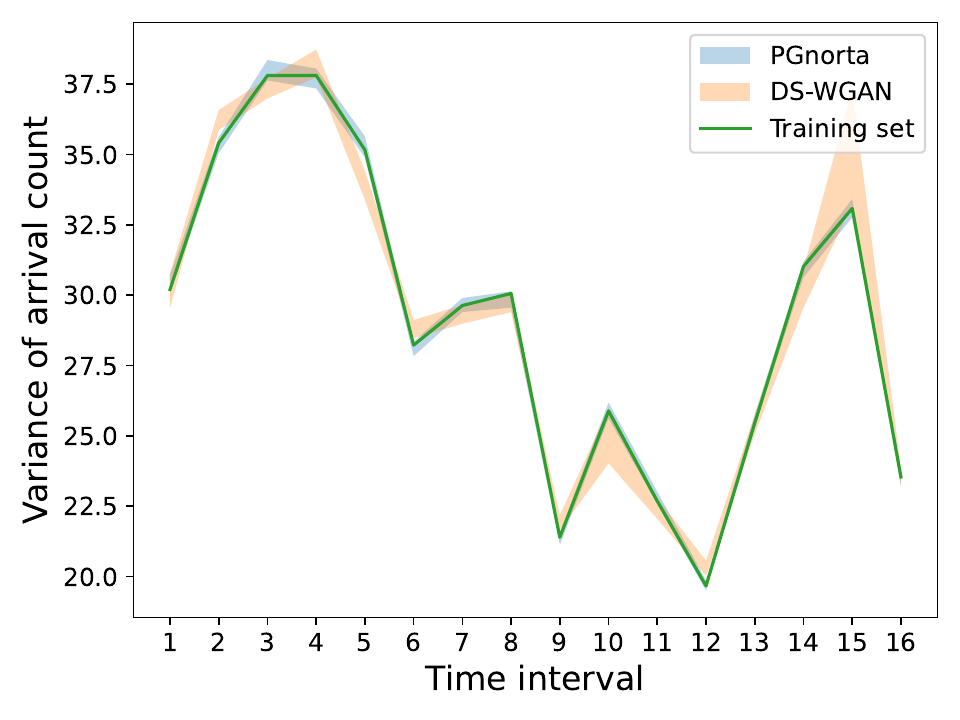}}

    \subfigure[Past and future correlation]{
        \label{fig:callcenter.corr0.5}
        \includegraphics[width=0.33\textwidth]{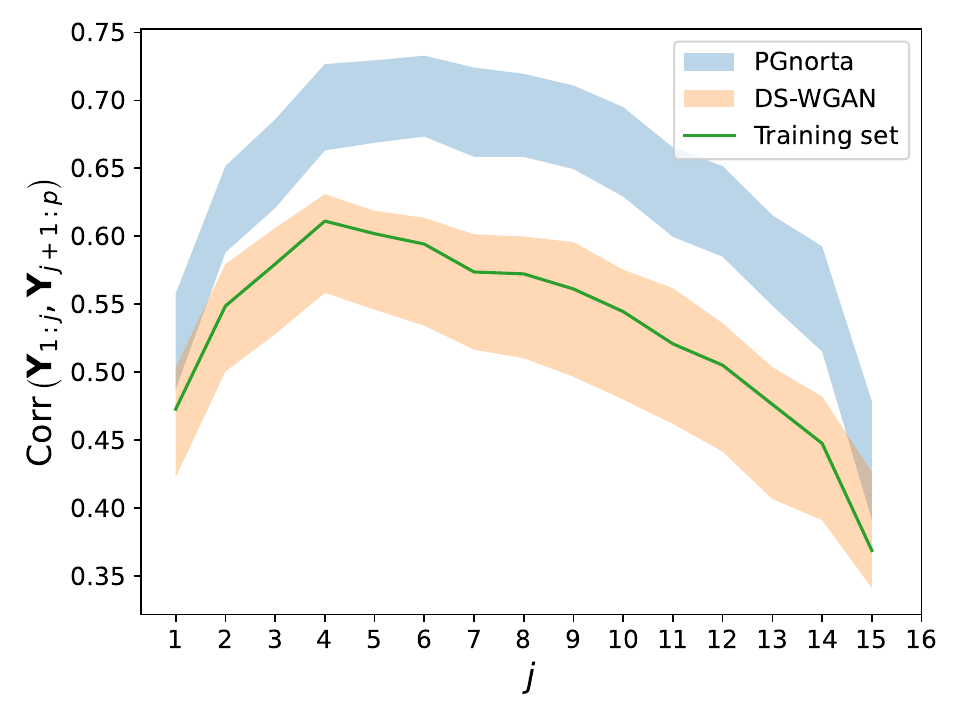}}
    \subfigure[Marginal Wasserstein distance]{
        \label{fig:callcenter.w_dist0.5}
        \includegraphics[width=0.33\textwidth]{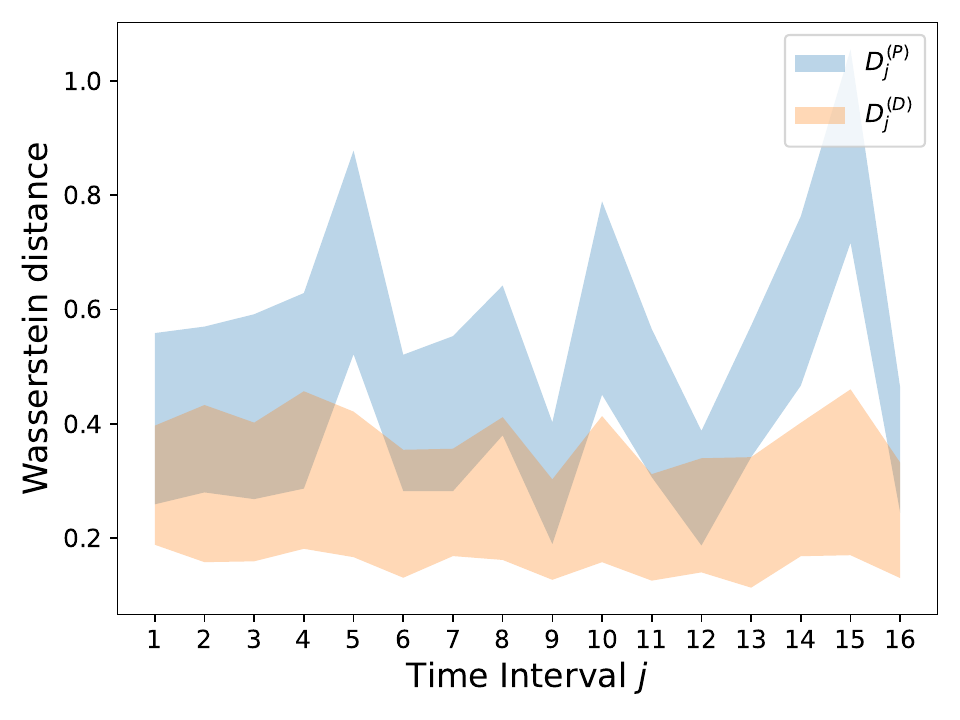}}

    \caption{Experiment results of Section \ref{sec:exper_callcenter_real}: with $0.5\%$ upper and lower quantiles of outliers removed.}
    \label{fig:callcenter_result_0.5}
\end{figure}

Specifically, we examine two cases. In the first case, we remove according to 2\% and 98\% percentiles for the data set as outliers. As a result, about 9\% of days of data are marked as outliers and removed. The experiment results are plotted in Figure \ref{fig:callcenter_result_2}. In the second case, we remove according to 0.5\% and 99.5\% percentiles for the data set as outliers. As a result, about 1\% of days of data are marked as outliers and removed. The experiment results are plotted in Figure \ref{fig:callcenter_result_0.5}. 

The plotted results for these two cases suggest that removing different percentages of outliers do not have significant impact on the reported performances. 



